\documentclass[a4paper,10pt,twoside]{article}

\usepackage{dsfont}
\usepackage{microtype}
\usepackage{times}  
\usepackage{geometry}
\usepackage{macros}
\usepackage{amsthm}
 \newtheorem{theorem}{Theorem}

\newtheorem{remark}{Remark}
\newtheorem{proposition}{Proposition}
\newtheorem{lemma}[theorem]{Lemma}

\usepackage{authblk}

\begin{document}

\title{Concentration bounds for temporal difference learning with linear function approximation: \\[0.5ex] The case of batch data and uniform sampling} 

\author[1]{Prashanth L. A.}
\affil[1]{\small   Indian Institute of Technology Madras}
\author[2]{Nathaniel Korda}
\affil[2]{\small Oxford University}
\author[3]{R\'emi Munos}
\affil[3]{\small Google Deepmind}
\date{}
\maketitle


\begin{abstract} 
	We propose a stochastic approximation (SA) based method with randomization of samples for policy evaluation using the least squares temporal difference (LSTD) algorithm.
		Our proposed scheme is equivalent to running regular temporal difference learning with linear function approximation, albeit with samples picked uniformly from a given dataset.
		Our method results in an $O(d)$ improvement in complexity in comparison to LSTD, where $d$ is the dimension of the data.  
	We provide non-asymptotic bounds for our proposed method, both in high probability and in expectation, under the assumption that the matrix underlying the LSTD solution is positive definite. The latter assumption can be easily satisfied for the pathwise LSTD variant proposed in \cite{lazaric2012finite}. Moreover, we also establish that using our method in place of LSTD does not impact the rate of convergence of the approximate value function to the true value function. These rate results coupled with the low computational complexity of our method make it attractive for implementation in {\em big data} settings, where $d$ is large. A similar low-complexity alternative for least squares regression is well-known as the stochastic gradient descent (SGD) algorithm. We provide finite-time bounds for SGD. 
	We demonstrate the practicality of our method as an efficient alternative for pathwise LSTD empirically by combining it with the least squares policy iteration (LSPI) algorithm in a traffic signal control application. We also conduct another set of experiments that combines the SA based low-complexity variant for least squares regression with the LinUCB  algorithm for contextual bandits, using the large scale news recommendation dataset from Yahoo. 
\end{abstract} 

\section{Introduction}
Several machine learning problems involve solving a linear system of equations from a given set of training data.
In this paper we consider the problem of policy evaluation in reinforcement learning (RL).
The objective here is to estimate the value function $V^\pi$ of a given policy $\pi$. Temporal difference (TD) methods are well-known in this context, and they are known to converge to the fixed point $ V^\pi =  \T^\pi(V^\pi)$, where $\T^\pi$ is the Bellman operator (see Section \ref{sec:td-background} for a precise definition).

The TD algorithm stores an entry representing the value function estimate for each state, making it computationally difficult to implement for problems with large state spaces. A popular approach to alleviate this curse of dimensionality is to parameterize the value function using a linear function approximation architecture. For every $s$ in the state space $\S$, we approximate $V^\pi(s) \approx \theta\tr \phi(s)$, where $\phi(\cdot)$ is a $d$-dimensional feature vector with $d << |\S|$, and $\theta$ is a tunable parameter.
The function approximation variant of TD \cite{tsitsiklis1997analysis} is known to converge to the fixed point of $\Phi \theta = \Pi \T^\pi(\Phi\theta)$, where $\Pi$ is the orthogonal projection onto the space within which we approximate the value function, and $\Phi$ is the feature matrix that characterizes this space. 
For a detailed treatment of this subject matter, the reader is referred to the classic textbooks \cite{bertsekas1996neuro,sutton1998reinforcement}. 

Batch reinforcement learning is a popular paradigm for policy learning. Here, we are provided with a (usually) large set of state transitions $\D\triangleq\{(s_i,r_i,s'_{i}),i=1,\ldots,T)\}$ obtained by simulating the underlying Markov decision process (MDP). For every $i=1,\ldots,T$, the 3-tuple $(s_i,r_i,s'_{i})$ corresponds to a transition from state $s_i$ to $s'_i$ and the resulting reward is denoted by $r_i$. The objective is to learn an {\em approximately optimal} policy from this set.  Least squares policy iteration (LSPI) \cite{lagoudakis2003least} is a well-known batch RL algorithm in this context, and it is based on the idea of policy iteration. A fundamental component of LSPI is least squares temporal difference (LSTD) \cite{Bradtke1996}, which is introduced next.

LSTD estimates the fixed point of $\Pi\T^\pi$, for a given policy $\pi$, using empirical data $\D$. The LSTD estimate is given as the solution to  
	\begin{align}
	\label{eq:lstd}
	&\hat \theta_T = \bar A_T^{-1} \bar b_T,\\
	& \hspace{-3em} \text{ where } \bar A_T \triangleq  \frac{1}{T} \sum_{i=1}^{T} \phi(s_i)(\phi(s_i) - \beta \phi(s'_i))\tr, \text{~~ and ~~}\bar b_T \triangleq  \frac{1}{T} \sum_{i=1}^{T} r_i \phi(s_i).\nonumber
	\end{align}

We consider a special variant of LSTD called {pathwise LSTD}, proposed in \cite{lazaric2012finite}. The idea behind pathwise LSTD is to 
\begin{inparaenum}[(i)]\item have the dataset $\D$ created using a sample path simulated from the underlying MDP for the policy $\pi$ and \item set $s_T'=0$ while computing $\bar A_T$ defined above. \end{inparaenum} The latter setting ensures the existence of the LSTD solution $\hat\theta_T$ under the condition that the family of features on the data set $\D$ are linearly independent.

Our primary focus in this work is to solve the LSTD system in a computationally efficient manner. 
Solving \eqref{eq:lstd} is computationally expensive, especially when $d$ is large.
	For instance, in the case when $\bar A_T^{-1}$ is invertible, the complexity of the approach above is $O(d^2 T)$, where $\bar A_T^{-1}$ is computed iteratively using the Sherman-Morrison lemma. On the other hand, if we employ the Strassen algorithm or the Coppersmith-Winograd algorithm for computing $\bar A_T^{-1}$, the complexity is of the order $O(d^{2.807})$ and $O(d^{2.375})$, respectively, in addition to $O(d^2 T)$ complexity for computing $\bar A_T$. An approach for solving \eqref{eq:lstd} without explicitly inverting $\bar A_T$ is computationally expensive as well.

\begin{figure}
	\centering
	\tikzset{
		>=stealth',
		punkt/.style={
			rectangle,
			rounded corners,
			draw=black, very thick,
			text width=10em,
			minimum height=4.5em,
			text centered},
		pil/.style={
			->,
			thick,
			shorten <=2pt,
			shorten >=2pt,}
	}
	\begin{tikzpicture}[auto, node distance=2cm,>=latex']
	\node (theta) {$\boldsymbol{\theta_n}$};
	\node [punkt, fill=blue!20,right=1.3cm of theta,label=below:{\color{bleu2}\bf Random Sampling},align=center] (sample) {\textbf{Pick $\boldsymbol{i_n}$ uniformly}\\[1ex] \textbf{in $\boldsymbol{\{1,\ldots,T}\}$}}; 
	\node [punkt, fill=green!20,right=1.4cm of sample,label=below:{\color{violet!90}\bf TD Update},align=center] (update) {\textbf{Update} $\boldsymbol{\theta_n}$ \\[1ex]\textbf{using $\boldsymbol{(s_{i_n},r_{i_n},s'_{i_n})}$}};
	\node [right=1.3cm of update] (end) {$\boldsymbol{\mathbf{\theta_{n+1}}}$};
	\draw [pil,->] (theta) --  (sample);
	\draw [pil,->] (sample) -- (update);
	\draw [pil,->] (update) -- (end);
	\end{tikzpicture}
	\caption{Overall flow of the the batchTD algorithm.}
	\label{fig:algorithm-flow}
\end{figure}

From the above discussion, it is evident that LSTD scales poorly with the number of features, making it inapplicable for large datasets with many features. 
We propose the batchTD algorithm to alleviate the high computation cost of LSTD in high dimensions. The batchTD algorithm replaces the inversion of the $\bar A_T$ matrix by the following iterative procedure that performs a fixed point iteration (see Figure \ref{fig:algorithm-flow} for an illustration): Set $\theta_0$ arbitrarily and update
\begin{align}
\theta_n = \theta_{n-1} + \gamma_{n} \left(r_{i_n} + \beta \theta_{n-1}\tr \phi(s'_{i_{n}}) - \theta_{n-1}\tr \phi(s_{i_n})\right)\phi(s_{i_n}),
\label{eq:random-lstd-update-intro}
\end{align}
where each $i_n$ is chosen uniformly at random from the set $\{1,\ldots,T\}$, and $\gamma_n$ are step-sizes that satisfy standard stochastic approximation conditions. 
The random sampling is sufficient to ensure convergence to the LSTD solution. The update iteration \eqref{eq:random-lstd-update-intro} is of order $O(d)$ and our bounds show that after $T$ iterations, the iterate $\theta_T$ is very close to LSTD solution, with high probability. The advantage of the scheme above is that it incurs a computational cost of $O(dT)$, while a traditional LSTD solver based on Sherman-Morrison lemma would require $O(d^2T)$.   

The update rule in \eqref{eq:random-lstd-update-intro} resembles that of TD(0) with linear function approximation, justifying the nomenclature `batchTD'. Note that regular TD(0) with linear function approximation uses a sample path from the Markov chain underlying the policy considered. In contrast, the batchTD algorithm performs the update iteration using a sample picked uniformly at random from a dataset. We establish, through non-asymptotic bounds, that using batchTD in place of LSTD does not  impact the convergence rate of LSTD to the true value function. The advantage with batchTD is the low computational cost in comparison to LSTD. 

From a theoretical standpoint, the scheme \eqref{eq:random-lstd-update-intro} comes under the purview of stochastic approximation (SA). 
Stochastic approximation is a well-known technique that was originally proposed for finding zeroes of a nonlinear function in the seminal work of Robbins and Monro \cite{robbins1951stochastic}. Iterate averaging is a standard approach to accelerate the convergence of SA schemes and was proposed independently in \cite{ruppert1991stochastic} and \cite{polyak1992acceleration}. Non asymptotic bounds for Robbins Monro schemes have been provided in \cite{frikha2012concentration} and extended to incorporate iterate averaging in \cite{fathi2013transport}. The reader is referred to \cite{kushner2003stochastic} for a textbook introduction to SA.

Improving the complexity of TD-like algorithms is a popular line of research in RL. The popular Computer Go setting \cite{silver2007reinforcement}, with dimension $d=10^6$,  and several practical application domains  (e.g. transportation, networks) involve high-feature dimensions. Moreover, considering that linear function approximation is effective with a large number of features, our $O(d)$ improvement in complexity of LSTD by employing a TD-like algorithm on batch data is meaningful. For other algorithms treating this complexity problem, see GTD \cite{sutton2009convergent}, GTD2 \cite{sutton2009fast}, iLSTD \cite{zinkevich2007ilstd} and the references therein. In particular, iLSTD is suitable for settings where the features admit a sparse representation.

In the context of improving the complexity of LSTD, our contributions can be summarized as follows:
First, through finite sample bounds, we show that our batchTD algorithm \eqref{eq:random-lstd-update-intro} converges to the pathwise LSTD solution at the optimal rate of $O(n^{-1/2})$ in expectation (see Theorem \ref{thm:flstd-rate} in Section \ref{sec:results}).
	By projecting the iterate \eqref{eq:random-lstd-update-intro} onto a compact and convex subset of $\R^d$, we are able to establish high probability bounds on the error $\l\theta_n - \hat\theta_T\r$. In particular, we show that, with probability $1-\delta$, the batchTD iterate $\theta_n$ constructs an $\epsilon$-approximation of the corresponding pathwise LSTD solution with $O(d\ln(1/\delta)/\epsilon^2)$ complexity, irrespective of the number of batch samples $T$. 
	The above rate results are for a step-size choice that is inversely proportional to the number of iterations of \eqref{eq:random-lstd-update-intro}, and also require the knowledge of the minimum eigenvalue of the symmetric part of $\bar A_T$. We overcome the latter dependence on the knowledge of the minimum eigenvalue through iterate averaging. As an aside, we note that using completely parallel arguments to those used in arriving at non-asymptotic bounds for batchTD, one could derive bounds for the regular TD algorithm with linear function approximation, albeit for the special case when the underlying samples arrive in an i.i.d. fashion.
	Second, through a performance bound, we establish that using our batchTD algorithm in place of LSTD does not impact the rate of convergence of the approximate value function to the true value function. 
	Third, we investigate the rates when larger step sizes ($\Theta(n^{-\alpha})$ where $\alpha \in $ $(1/2,1)$) are used in conjunction with averaging of the iterates, i.e., the well known Polyak-Ruppert averaging scheme.  
	The rate obtained in high probability for the iterate-averaged variant is of the order $O(n^{-\alpha/2})$, with the added advantage that, unlike non-averaged case, the step-size choice does not require knowledge of the minimum eigenvalue of the symmetric part of $\bar A_T$. Further, with iterate averaging the complexity of the algorithm stays at $O(d)$ per iteration as before.
	Fourth, we consider a traffic control application, and implement a variant of LSPI which uses the batchTD algorithm in place of LSTD. In particular, for the 
	experiments we employ step-sizes that were used to derive the non-asymptotic bounds mentioned above. 
	We demonstrate that running batchTD for a short number of iterations ($\sim 500$) on big-sized problems with feature dimension $\sim 4000$, one gets a performance that is almost as good as regular LSTD at a significantly lower computational cost.

We now turn our attention to solving least squares regression problems via the popular stochastic gradient descent (SGD) method.
Many practical machine learning algorithms require computing the least squares solution at each iteration in order to make a decision. As in the case of LSTD, classic least squares solution schemes such as Sherman-Morrison lemma are of complexity of the order $O(d^2)$. A practical alternative is to use a SA based iterative scheme that is of the order $O(d)$. Such SA based schemes when applied to the least squares parameter estimation context are well known in the ML literature as stochastic gradient descent (SGD) algorithms. 

We also analyze the low-complexity SGD alternative for the classic least squares parameter estimation problem. 
	Using the same template as for the results of batchTD, we derive non-asymptotic bounds, which hold both in high probability as well as in expectation, for the tracking error $\| \theta_n - \hat\theta_T\|_2$. Here $\theta_n$ is the SGD iterate, while $\hat\theta_T$ is the least squares solution. We describe a fast variant of the LinUCB \cite{li2010contextual} algorithm for contextual bandits, where the SGD iterate  is used in place of the least squares solution. We demonstrate the empirical usefulness of the SGD based LinUCB algorithm using the large scale news recommendation dataset from Yahoo \cite{webscope}. We observe that, using the step-size suggested by our bounds, the SGD based LinUCB algorithm exhibits low tracking error, while providing significant computational gains. 

The rate results coupled with the low complexity of our schemes, in the context of LSTD as well as least squares regression, make them more amenable to practical implementation in the canonical {\em big data} settings, where the dimension $d$ is large. 
This is amply demonstrated in our applications in transportation and recommendation systems domains, where we establish that batchTD and SGD perform almost as well as regular LSTD and regression solvers, albeit with much less computation (and with less memory).  Note that the empirical evaluations are for higher level machine learning algorithms  - least squares policy iteration (LSPI) \cite{lagoudakis2003least} and linear bandits \cite{dani2008stochastic,li2010contextual}, which use LSTD and regression in their inner loops. 

The rest of the paper is organized as follows: In Section \ref{sec:related}, we discuss related work.  In Section \ref{sec:flstd} we present the batchTD algorithm, and in Section \ref{sec:results} we provide the non-asymptotic bounds for this algorithm.  In Section \ref{sec:iterate-averaging}, we analyze a variant of our algorithm that incorporates iterate averaging. In Section \ref{sec:relatedworkrecent}, we compare our bounds to those in recent work. In Section \ref{sec:flspi}, we describe a variant of LSPI that uses batchTD in place of LSTD. Next, in Section \ref{sec:analysis}, we provide detailed proofs of convergence and derivation of rates.  We provide experiments on a traffic signal control application in Section \ref{sec:experiments}. In Section \ref{sec:random-batch}, we provide extensions to solve the problem of least squares regression and in Section \ref{sec:flinucb}, we provide a set of experiments that tests a variant of the LinUCB algorithm using a SA based subroutine for least squares regression. Finally, in Section \ref{sec:conclusions} we provide the concluding remarks.

\section{Literature review}
\label{sec:related}
\subsection{Previous work related to LSTD}
In Chapter 6 of \cite{kondathesis}, the authors establish that LSTD has the optimal asymptotic convergence rate, while in \cite{antos2008learning} and \cite{lazaric2012finite}, the authors provide a finite time analysis for LSTD and also LSPI. 
	Recent work in \cite{bruno} provides sample complexity bounds for LSTD($\lambda$). LSPE($\lambda$), which is an algorithm that is closely related to LSTD($\lambda$), is analyzed in \cite{yu2009convergence}. The authors there provide asymptotic rate results for LSPE($\lambda$), and show that it matches that of LSTD($\lambda$). Also related is the work in \cite{pires2012statistical}, where the authors study linear systems in general, and as a special case, provide error bounds for LSTD with improved dependence on the underlying feature dimension. 

A closely related contribution that is geared towards improving the computational complexity of LSTD is iLSTD \cite{zinkevich2007ilstd}. However, the analysis for iLSTD requires that the feature matrix be sparse, while we provide finite-time bounds for our fast LSTD algorithm without imposing sparsity on the features. 
Another line of related previous work is GTD \cite{sutton2009convergent}, and its later enhancement GTD2 \cite{sutton2009fast}. The latter algorithms feature an update iteration that can be viewed as gradient descent and operate in the online setting similar to the regular TD algorithm with function approximation. However, the advantage with GTD/GTD2 is that these algorithms are provably convergent to the TD fixed point even when the policy used for collecting samples differs from the policy being evaluated --- the so-called \textit{off-policy} setting.  
Recent work in \cite{liu2015finite} provides finite time analysis for the GTD algorithm. 
Unlike GTD-like algorithms, we operate in an offline setting with a batch of samples provided beforehand. LSTD is a popular algorithm here, but has a bad dependency in terms of computational complexity on the feature dimension, and we bring this down from $O(d^2)$ to $O(d)$ by running an algorithm that closely resembles TD on the batch of samples. This algorithm is shown to retain the convergence rate of LSTD.  

To the best of our knowledge, efficient SA algorithms that approximate LSTD without impacting its rate of convergence to LSTD solution, have not been proposed before in the literature. The high probability bounds that we derive 
	for batchTD do not directly follow from earlier work on LSTD algorithms. Concentration bounds for stochastic approximation schemes have been derived in \cite{frikha2012concentration}. 
	While we use their technique for proving the high-probability bound on batchTD iterate (see Theorem \ref{thm:flstd-rate}), our analysis is more elementary, and we make all the constants explicit for the problem at hand. Moreover, in order to eliminate a possible exponential dependence of the constants in the resulting bound on the reciprocal of the minimum eigenvalue of the symmetric part of $\bar A_T$, we depart from the argument in \cite{frikha2012concentration}.

Finite sample analysis of TD with linear function approximation has received more attention in recent works (cf. \cite{dalal2018finite,bhandari2018finite,csaba18tdbounds}). A detailed comparison of our bounds to those in the aforementioned references is provided in Section \ref{sec:relatedworkrecent}. 
	
	This paper is an extended version of an earlier work (see \cite{prashanth2014fastLS}). This work corrects the errors in the earlier work by using significant deviations in the proofs, and includes additional simulation experiments.
	Finally, in \cite{narayanan2017finite}, the authors list a few problems with the results and proofs in the conference version \cite{prashanth2014fastLS}, and the corrections incorporated in this work address the comments in \cite{narayanan2017finite}. 

\subsection{Previous work related to SGD}
Finite time analysis of SGD methods have been provided in \cite{bach2011non}.  While the bounds in \cite{bach2011non} are given in expectation, many machine learning applications require high probability bounds, which we provide for our case. 
Regret bounds for online SGD techniques have been given in \cite{zinkevich2003online,hazan2011beyond}: the gradient descent algorithm in \cite{zinkevich2003online} is in the setting of optimising the average of convex loss functions whose gradients are available, while that in \cite{hazan2011beyond} is for strongly convex loss functions. 
In comparison to previous work w.r.t. least squares regression, we highlight the following differences:\\
\begin{inparaenum}[\bfseries(i)]
	\item Earlier works on strongly convex optimization (cf. \cite{hazan2011beyond}) require the knowledge of the strong convexity constant in deciding the step-size. 
	While one can regularize the problem to get rid of the step-size dependence on $\mu$, it is not straightforward to choose the regularization constant. Notice that for SGD type schemes, one requires that the matrix $\bar A_T$ have a minimum positive eigenvalue $\mu$. Equivalently, this implies that the original problem is regularized with $T\mu$. This may turn out to be too high a  regularization and hence it is desirable to have SGD get rid of this dependence without changing the problem itself. 
	This is precisely what iterate-averaged SGD achieves, i.e., optimal rates both in high probability and expectation even for the un-regularized problem. To the best of our knowledge, there is no previous work that provides non-asymptotic bounds, both in high probability and in expectation, for iterate-averaged SGD. 
	\\
	\item Our analysis is for the classic SGD scheme that is anytime, whereas the epoch-GD algorithm in \cite{hazan2011beyond} requires the knowledge of the time horizon. \\
	\item While the algorithm in \cite{bach2013non} is shown to exhibit the optimal rate of convergence without assuming strong convexity, the bounds there are in expectation only. In contrast, for the special case of strongly convex functions,  we derive high-probability bounds in addition to bounds in expectation. Furthermore, the bound in expectation from \cite{bach2011non} is not optimal for a strongly convex function in the sense that the initial error (which depends on where the algorithm started) is not forgotten as fast as the rate that we derive. \\
	\item On a minor note, our analysis is simpler since we work directly with least squares problems, and we make all the constants explicit for the problems considered. 
\end{inparaenum}

\section{TD with uniform sampling on batch data {\em\textmd(batchTD)}}
\label{sec:flstd}
We propose here a stochastic approximation variant of the LSTD algorithm, whose iterates converge to the same fixed point as the regular LSTD algorithm, while incurring much smaller overall computational cost.
The algorithm, which we call batchTD, is a simple stochastic approximation scheme that updates incrementally using samples picked uniformly from batch data.
The results that we present establish that the batchTD algorithm computes an $\epsilon$-approximation to the LSTD solution $\hat\theta_T$ with probability $1-\delta$, while incurring a complexity of the order $O(d\ln(1/\delta)/\epsilon^2)$, irrespective of the number of samples $T$. In turn, this enables us to give a performance bound for the approximate value function computed by the batchTD algorithm. 

In the following section, we provide a brief background on LSTD and pathwise LSTD. In the subsequent section, we present our batchTD algorithm. 

\subsection{Background}
\label{sec:td-background}
Consider an MDP with state space $\S$ and action space $\A$, both assumed to be finite. Let  $p(s,a,s')$, $s,s' \in \S, a\in \A$ denote the probability of transitioning from state $s$ to $s'$ on action $a$. Let $\pi$ be a stationary randomized policy, i.e., $\pi(s, \cdot)$ is a distribution over $\A$, for any $s\in \S$. The value function $V^\pi$ is defined by
\begin{equation}
	V^\pi(s) \triangleq \E\left[\sum_{t=0}^{\infty} \beta^t \sum_{a \in \A} r(s_t, a)\pi(s_t, a)\mid s_0=s\right],
	\label{eq:vf}
	\end{equation}
	where $s_t$ denotes the state of the MDP at time $t$, $\beta \in [0,1)$ the discount factor, and $r(s,a)$ denotes the instantaneous reward obtained in state $s$ under action $a$. The value function $V^\pi$ can be expressed as the fixed point of the Bellman operator $\T^\pi$ defined by
	\begin{equation}
	\T^\pi(V)(s) \triangleq \sum_{a \in \A} \pi(s,a) \left( r(s,a) +  \beta \sum\limits_{s'} p(s,a,s') V(s')\right).
	\label{vf}
	\end{equation}

When the cardinality of $\S$ is huge, a popular approach is to parameterize the value function using a linear function approximation architecture, i.e., for every $s \in \S$, approximate $V^\pi(s) \approx  \phi(s)\tr \theta$, where $\phi(s)$ is a $d$-dimensional feature vector for state $s$ with $d \ll |\S|$, and $\theta$ is a tunable parameter.
With this approach, the idea is to  find the best approximation to the value function $V^\pi$ in $\B=\{\Phi \theta \mid \theta \in \R^d\}$, which is a vector subspace of $\R^{|S|}$. In this setting, it is no longer feasible to find the fixed point $V^\pi =\T^\pi V^\pi$. Instead, one can approximate $V^\pi$ within $\B$ by solving the following projected system of equations:
\begin{align}
\Phi \theta^* = \Pi \T^\pi(\Phi \theta^*).
\label{eq:td-fixed-point}
\end{align}
In the above, $\Phi$ denotes the feature matrix with rows $\phi(s)\tr, \forall s \in \S$, and $\Pi $ is the orthogonal projection onto $\B$.
Assuming that the matrix $\Phi$ has full column rank, it is easy to derive that  $\Pi = \Phi (\Phi\tr \Psi \Phi )^{-1}\Phi\tr\Psi$, where $\Psi$ is the diagonal matrix whose diagonal elements form the stationary distribution (assuming it exists) of the Markov chain associated with the policy $\pi$.

The solution $\theta^*$  of \eqref{eq:td-fixed-point} can be re-written as follows (cf. \cite[Section 6.3]{bertsekas2011approximate}):
\begin{align}
	\label{eq:td-system-of-equations}
	A \theta^* = b, \textrm{ where } A \triangleq \Phi\tr \Psi(I- \beta P)\Phi \textrm{ and } b \triangleq \Phi\tr \Psi {\cal R},
	\end{align}
where $P = [P(s,s')]_{s,s'\in\S}$ is the transition probability matrix with components $P(s,s') = p(s,\pi(s),s')$, ${\cal R}$ is the vector with components $\sum_{a \in \A} \pi(s,a) r(s,a)$, for each $s\in \S$, and $\Psi$ the stationary distribution (assuming it exists) of the Markov chain for the underlying policy $\pi$.

In the absence of knowledge of the transition dynamics $P$ and stationary distribution $\Psi$, LSTD is an approach which can approximate the solution $\theta^*$ using a batch of samples obtained from the underlying MDP. In particular it requires a dataset, $\D = \{(s_i,r_i,s'_i),i=1,\ldots,T)\}$, where each tuple in the dataset $(s_i,r_i,s_i')$ represents a state-reward-next-state triple chosen by the policy. The LSTD solution approximates $A$, $b$, and $\theta^*$ with $\bar{A}_T$, $\bar{b}_T$ using the samples in $\D$ as follows:
\begin{align}
	\label{eq:lstd-solution}
	&\hat \theta_T = \bar A_T^{-1} \bar b_T, \\
	&\text{where } \bar A_T \triangleq  \dfrac1{T}\sum_{i=1}^{T} \phi(s_i)(\phi(s_i) - \beta \phi(s'_i))\tr, \text{~~ and ~~}\bar b_T \triangleq \dfrac1{T} \sum_{i=1}^{T} r_i \phi(s_i).\nonumber
	\end{align}
Denoting the current state feature $(T\times d)$-matrix by $\Phi \triangleq (\phi(s_1)\tr,\dots, \phi(s_T))$, next state feature $(T\times d)$-matrix by $\Phi' \triangleq (\phi(s_1')\tr,\dots, \phi(s_T'))$, and reward $(T\times 1)$-vector by ${\cal R} = (r_1,\dots,r_T)\tr$, we can rewrite $\bar A_T$ and $\bar b_T$ as follows\footnote{By an abuse of notation, we shall use $\Phi$ to denote the feature matrix for TD as well as LSTD and the composition of $\Phi$ should be clear from the context.}:
\begin{align*}
\bar A_T = \frac{1}{T}(\Phi\tr\Phi - \beta\Phi\tr\Phi'),
\text{ and }
\bar b_T = \frac{1}{T}\Phi\tr\cal R.
\end{align*}
It is not clear whether $\bar A_T$ is invertible for an arbitrary dataset $\D$. One way to ensure invertibility is to adopt the approach of pathwise LSTD, proposed in \cite{lazaric2012finite}.
The pathwise LSTD algorithm is an on-policy version of LSTD. It obtains samples, $\D$ by simulating a sample path of the underlying MDP using policy $\pi$, so that $s_i' = s_{i+1}$ for $i = 1,\dots,T-1$. The dataset thus obtained is perturbed slightly by setting the feature of the next state of the last transition, $\phi(s_T')$, to zero. This perturbation, as suggested in \cite{lazaric2012finite}, is crucial to ensure that the system of the equations that we solve as an approximation to \eqref{eq:td-system-of-equations} is well-posed. For the sake of completeness, we make this precise in the following discussion, which is based on Sections 2 and 3 of \cite{lazaric2012finite}.

Define the empirical Bellman operator $\hat T: \R^T \rightarrow \R^T$ as follows: For any $y \in \R^T$,
\begin{align}
	(\hat T y)_i \triangleq 
	\begin{cases}
	r_i + \beta y_{i+1}, & \textrm{ for }1\le i < T, \textrm{ and } \\
	r_T, &  \textrm{ for }i=T.
	\end{cases}
	\end{align}
Let $\hat {\cal R}$ be a $T\times 1$ vector with entries $r_i$, $i=1,\ldots,T$ and $(\hat {\cal V} y)_i = y_{i+1}$ if $i<n$ and $0$ otherwise. Then, it is clear that $\hat T y = \hat {\cal R} + \beta \hat {\cal V} y$. 

Let $\G_T \triangleq \{ (\phi(s_1)\tr\theta,\ldots,\phi(s_T)\tr\theta)\tr \mid \theta \in \R^d\} \subset \R^T$ be the vector sub-space of $\R^T$ within which pathwise LSTD approximates the true values of the value function corresponding to the states $s_1,\dots, s_T$, and it is the empirical analogue of $\B$ defined earlier.
	It is easy to see that $\G_T = \{ \Phi \theta \mid \theta \in \R^d\}$. 
	Let $\hat \Pi$ be the orthogonal projection onto $\G_T$ using the empirical norm, which is defined as follows: $\| f \|^2_T \triangleq T^{-1}\sum_{i=1}^{T} f(s_i)^2$, for any function $f$.
	Notice that $\hat \Pi \hat T$ is a contraction mapping, since
	\begin{align*}
	\lt \hat \Pi \hat T y - \hat \Pi \hat T z \rt \le &\lt  \hat T y - \hat T z \rt
	=  \beta \lt \hat {\cal V} y - \hat {\cal V} z \rt 
	\le  \beta \lt  y -  z \rt.
	\end{align*}
	Hence, by the Banach fixed point theorem, there exists some $v^* \in \G_T$ such that $\hat \Pi \hat T v^* = v^*$.

Suppose that the feature matrix $\Phi$ is full rank -- an assumption that is standard in the analysis of TD-like algorithms and also beneficial in the sense that it ensures that the system of equations we attempt to solve is well-posed. Then, it is easy to see that there exists a unique $\hat\theta_T$ such that $v^* = \Phi \hat\theta_T$. 
Moreover, replacing $\bar A_T$ in \eqref{eq:lstd-solution} with
\begin{align}\label{eq:barA-pathwise}
\bar A_T = \frac{1}{T}\Phi\tr(I-\beta\hat P) \Phi,
\end{align}
where $\hat P$ is a $T\times T$ matrix with $\hat P(i,i+1)=1$ for $i=1,\ldots,T-1$ and $0$ otherwise, it is clear that $\bar A_T$ is invertible and $\hat\theta_T$ is the unique solution to \eqref{eq:lstd-solution}.

\begin{remark}\textit{(Regular vs. Pathwise LSTD)}
	For a large data set, $\D$, generated from a sample path of the underlying MDP for policy $\pi$, the difference in the matrix used as $\bar A_T$ in LSTD and pathwise LSTD is negligible. In particular,
	the difference in $\ell_2$-norm of $\bar A_T$ composed with and without zeroing out the next state in the last transition of $\D$ can be upper bounded by a constant multiple of $\dfrac1{T}$. As mentioned earlier, zeroing out the next state in the last transition of $\D$ together with a full-rank $\Phi$ makes the system of equations in \eqref{eq:lstd-solution} well-posed. 
	As an aside, the batchTD algorithm, which we describe below, would work as a good approximation to LSTD, as long as one ensures that $\bar A_T$ is positive definite. Pathwise LSTD presents one approach to achieve the latter requirement, and it is an interesting future research direction to derive other conditions that ensure $\bar A_T$ is positive definite.
\end{remark}

\subsection{Update rule and pseudocode for the batchTD algorithm}
The idea is to perform an incremental update that is similar to TD, except that the samples are drawn uniformly randomly from the dataset $\D$.  Recall that, in the case of pathwise LSTD, the data set corresponds to those along a sample path simulated from the underlying MDP for a given policy $\pi$, i.e., $s'_i = s_{i+1}$, $i=1,\ldots,T-1$ and $s'_T=0$. 

The full pseudocode for batchTD is given in Algorithm \ref{alg:batchTD}.
Starting with an arbitrary $\theta_0$, we update the parameter $\theta_n$ as follows:
\begin{align}
\theta_n = \Upsilon\left(\theta_{n-1} + \gamma_{n} \left(r_{i_n} + \beta \theta_{n-1}\tr \phi(s'_{i_{n}}) - \theta_{n-1}\tr \phi(s_{i_n})\right)\phi(s_{i_n})\right),
\label{eq:random-lstd-update}
\end{align}
where each $i_n$ is chosen uniformly randomly from the set $\{1,\ldots,T\}$. In other words, we pick a sample with uniform probability $1/T$  from the set $\D = \{(s_i,r_i,s'_i),i=1,\ldots,T)\}$ and use it to perform a fixed point iteration in \eqref{eq:random-lstd-update}. The quantities $\gamma_n$ above are \emph{step sizes} that are chosen in advance and satisfy standard stochastic approximation conditions, i.e., $\sum_{n}\gamma_n = \infty$, and $\sum_n\gamma_n^2 <\infty$. 
The operator $\Upsilon$ projects the iterate $\theta_n$ onto the nearest point in a closed ball $\C \subset \R^d$ with a radius $H$ that is large enough to include $\hat \theta_T$. Note that projection via $\Upsilon$ amounts to scaling down the $\ell_2$-norm of the iterate $\theta_n$ so that it does not exceed $H$, and is a computationally inexpensive operation. 

In the next section, we present non-asymptotic bounds for the error $\l \theta_n - \hat \theta_T\r$ that hold with high probability, and in expectation, for the projected iteration in \eqref{eq:random-lstd-update}. Further, we also provide an error bound  that holds in expectation for a variant of \eqref{eq:random-lstd-update} without involving the projection operation. 
	From the bounds presented below, we can infer that, 
	for a step size choice that is inversely proportional to the number $n$ of iterations, obtaining the optimal $O\left(1/\sqrt{n}\right)$ requires the knowledge of the minimum eigenvalue $\mu$ of $\frac{1}{2}\left(\bar A_T + \bar A_T\tr\right)$, where $\bar A_T$ is a matrix made from the features used in the linear approximation (see assumption (A1) below).
	Subsequently, in Section \ref{sec:iterate-averaging}, we present non-asymptotic bounds for a variant of the batchTD algorithm, which employs iterate averaging. The bounds for iterate-averaged batchTD establish that the knowledge of eigenvalue $\mu$ is not needed to obtain a rate of convergence that can be made arbitrarily close to $O\left(1/\sqrt{n}\right)$.

\begin{algorithm}
	\caption{The batchTD algorithm}
	\label{alg:batchTD}
	\begin{algorithmic} 
		\State {\bfseries Input:} Sample path based dataset $\D\triangleq\{(s_i,r_i,s'_i),i=1,\ldots,T)\}$ such that $s'_i = s_{i+1}$, $i=1,\ldots,T-1$ and $s'_T=0$; a choice of step-size sizes, $\gamma_k$; a time horizon $n$.
		\State {\bfseries Initialization:} Set $\theta_0$.
		\State {\bfseries Run:}
		\For{$k=1 \dots n$}
		\State Get a random sample index: $i_k\sim U(\{1,\dots,T\})$.
		\State Perform update iteration: $\theta_k = \Upsilon\left(\theta_{k-1} + \gamma_{k} \left(r_{i_k} + \beta \theta_{k-1}\tr \phi(s'_{i_k}) - \theta_{k-1}\tr \phi(s_{i_k})\right)\phi(s_{i_k})\right)$. 
		\EndFor
		\State {\bfseries Output:} $\theta_n$
	\end{algorithmic}
\end{algorithm}

\section{Main results for the batchTD algorithm}
\label{sec:results}
\paragraph{Map of the results:}
Theorem \ref{thm:flstd-asym} proves almost sure convergence of batchTD iterate $\theta_n$ to LSTD solution $\hat \theta_T$, with and without projection.
Theorem \ref{thm:flstd-rate} provides finite time bounds both in high probability and in expectation for the error $ \| \theta_n - \hat \theta_T \|_2$, where $\theta_n$ is given by \eqref{eq:random-lstd-update}. We require high probability bounds to qualify the rate of convergence of the approximate value function $\Phi \theta_n$ to the true value function, i.e., a variant of Theorem 1 in \cite{lazaric2012finite} for the case of the batchTD algorithm.
Theorem \ref{thm:perf-bound-lstd} presents a performance bound for the special case when the dataset $\D$ comes from a sample path of the underlying MDP for the given policy $\pi$.  
Note that the first three results above hold irrespective of whether the dataset $\D$ is based on a sample path or not. However, the performance bound is for a sample path dataset only and is used to illustrate that using batchTD in place of regular LSTD does not harm the overall convergence rate of the approximate value function to the true value function.

We state all the results in Sections \ref{sec:asym}--\ref{sec:pb} and provide detailed proofs of all the claims in Section \ref{sec:analysis}. Also, all the results are by default for the projected version of the batchTD algorithm, i.e., $\theta_n$ given by \eqref{eq:random-lstd-update}, while Section \ref{sec:noproj-results} presents the results for the projection-free batchTD variant.
In particular, the latter section provides both asymptotic convergence and a bound in expectation for the error $ \| \theta_n - \hat \theta_T \|_2 $ for the projection-free variant of batchTD. 

\subsection{Assumptions}
We make the following assumptions for the analysis of the batchTD algorithm:
\begin{enumerate}[\bfseries({A}1)]
	\item The matrix $\bar{A}_T$ is positive definite, which implies the smallest eigenvalue $\mu$ of its symmetric part $\frac{1}{2}\left(\bar{A}_T + \bar{A}_T\tr\right)$ is greater than zero\footnote{A real matrix $A$ is positive definite if and only if the symmetric part
		$\frac1{2}(A+A\tr)$ is positive definite.}.
	\item Bounded features: $\l \phi(s_i) \r \le \Phi_{\max}< \infty,$ for $i=1,\ldots,T$.
	\item Bounded rewards: $|r_i|\le R_{\max} < \infty$ for $i=1,\ldots,T$.
	\item The set $\C\triangleq\{\theta \in \R^d \mid \l \theta \r \le H\}$ used for projection through $\Upsilon$ satisfies $H > \frac{\l \bar b_T\r}{\mu}$, where $\mu$ is as defined in (A1).
\end{enumerate}
In the following sections, we present results for the generalized setting, i.e., the dataset $\D$ does not necessarily come from a sample path of the underlying MDP, but we assume (see (A1)) that the matrix $\bar A_T$ is positive definite. For pathwise LSTD, (A1) can be replaced by the following assumption:
\begin{description}
		\item[\bfseries({A}1')]The matrix $\Phi$ is full rank.
\end{description}

Recall that the pathwise LSTD  in \cite{lazaric2012finite} perturbs the data set slightly, as discussed in Section \ref{sec:td-background} above. Thus, from  \eqref{eq:barA-pathwise}, we have
	\begin{align}
	\mu \ge \frac{(1-\beta)}{T}\mu', 
	\text{ where }
	\mu' \triangleq \lambda_{\min}(\Phi\tr\Phi) \label{eq:def-mu-prime}.
	\end{align}
	The inequality above holds because $\l \hat P v\r \le \l v \r$, and $\l \hat P\tr v\r \le \l v \r$, leading to 
	the fact that  $\lambda_{\min}\left(I - \frac{\beta}{2}\left(\hat P + \hat P\tr\right)\right)\ge (1-\beta).$ 
	Thus, it is easy to infer that (A1') implies (A1), using \eqref{eq:def-mu-prime} in conjunction with the fact that a full rank $\Phi$ implies $\mu'>0$.

Note that the dataset is assumed to be fixed for all the results presented below.

\subsection{Asymptotic convergence}
\label{sec:asym}
\begin{theorem}
	\label{thm:flstd-asym}
	Assume (A1)-(A4), and also that the step sizes $\gamma_n \in \R_+$ satisfy $\sum_{n}\gamma_n = \infty$, and $\sum_n\gamma_n^2 <\infty$. Then, for the iterate $\theta_n$ updated according to \eqref{eq:random-lstd-update}, we have  
	\begin{align}
	\theta_n \rightarrow \hat\theta_T \text{ a.s. as } n \rightarrow \infty.
	\label{eq:theta-asymp-conv}
	\end{align}
\end{theorem}
\begin{proof}
	See Section \ref{sec:proof-asymp}.
\end{proof}

\subsection{Non-asymptotic bounds}
\label{sec:ftb}
The main result that bounds the computational error $\l \theta_n - \hat \theta_T \r$ with explicit constants is given below.
\begin{theorem}[\textit{Error bounds for batchTD}]
	\label{thm:flstd-rate}
	\ \\Assume (A1)-(A4). Set $\gamma_n= \frac{c_0c}{(c+n)}$ such that $c_0\in(0,\mu((1+\beta)^2\Phi_{\max}^4)^{-1}]$ and $ c_0 c > \frac{1}{\mu}$.
		Then, for any $\delta >0$, we have
		\begin{align}
		\E \l \theta_n - \hat \theta_T \r &\le \dfrac{K_1(n)}{\sqrt{n+c}}, \label{eq:l2-expect-bound-lstd}
		\textrm{~~ and ~~}\\ \P\left( \l \theta_n - \hat \theta_T \r \right.&\left.\le \dfrac{K_2(n)}{\sqrt{n+c}}\right) \ge 1 - \delta. \label{eq:l2-prob-bound-lstd}
		\end{align}
		In the above, $K_1(n)$ and $K_2(n)$ are functions of order $O(1)$, defined by\footnote{For notational convenience, we have chosen to ignore the dependence of $K_1$ and $K_2$ on the confidence parameter $\delta$.}:
		\begin{align*}
		K_1(n)&\triangleq  \frac{\l \theta_0 - \hat\theta_T \r \sqrt{(c+1)^{c_0 c\mu}}}{\sqrt{(n+c)^{c_0 c\mu-1}}}
		+ \frac{ 2ec_0 c \left(R_{\max} + (1+\beta)H\Phi_{\max}^2\right)}{\sqrt{2c_0 c\mu - 1}}, \textrm{ and }\\					
		K_2(n)&\triangleq 2\sqrt{e} c_0 c \left(R_{\max} +(1+\beta)H\Phi_{\max}^2\right) \sqrt{\frac{\log{\delta^{-1}} }
			{c_0 c \mu - 1} }
		+K_1(n).
		\end{align*}
\end{theorem}
\begin{proof}
	See Section \ref{sec:proof-lstd}.
\end{proof}

A few remarks are in order.
\begin{remark}\label{remark:flstd-error-split}
	\textbf{\textit{(Initial vs. sampling error)}}
	The bound in expectation above can be re-written as
	\begin{align}
	\E \l \theta_n - \hat \theta_T \r \le  \frac{\l \theta_0 - \hat\theta_T \r \sqrt{(c+1)^{c_0 c\mu}}}{(n+c)^{c_0 c\mu/2}}
	+ \frac{ 2ec_0 c \left(R_{\max} + (1+\beta)H\Phi_{\max}^2\right)}{\sqrt{2c_0 c\mu - 1}\sqrt{n+c}}.
	\label{eq:abc}
	\end{align}
	The first term on the RHS above is the initial error, while the second term is the sampling error. The initial error depends on the initial point $\theta_0$ of the algorithm. The sampling error arises out of a martingale difference sequence that depends on the random deviation of the stochastic update from the standard fixed point iteration.
	From \eqref{eq:abc}, it is evident that the initial error is forgotten at the rate $O\left( \dfrac{1}{n^{c_0 c\mu/2}} \right)$. Since $c_0 c \mu >1$, the former rate is faster than the rate $ O(1/\sqrt{n})$ at which the sampling error decays.  
\end{remark}

\begin{remark}\textbf{\textit{(Rate dependence on the minimum eigenvalue $\mu$)}}
		We note that setting $c$ such that  $c_0 c \mu = \eta \in(1,\infty)$ we can rewrite the constants in Theorem \ref{thm:flstd-rate} as:
	\begin{gather*}
	K_1(n)=  \frac{\l \theta_0 - \hat\theta_T \r \sqrt{(c+1)^{\eta}}}
	{\sqrt{ (n+c)^{(\eta - 1)}}}
	+ \frac{2e\eta}{\mu\sqrt{(2\eta -1)}}\left(R_{\max}+(1+\beta)H\Phi_{\max}^2\right), \textrm{ and }\\
	K_2(n)= 2\sqrt{e} \frac{\eta}{\mu} \left(R_{\max}+(1+\beta)H\Phi_{\max}^2\right) \sqrt{ \frac{\log{\delta^{-1}} }
		{\left(\eta - 1\right)} }
	+K_1(n).
	\end{gather*}
	So both the bounds in expectation and high probability have a linear dependence on the reciprocal of $\mu$.
	Note also that the constant $(R_{\max}+(1+\beta)H\Phi_{\max}^2)$ is nothing more than a bound on the size of the random innovations made by the algorithm at each time step.
\end{remark}

\begin{remark}\textbf{\textit{(Eigenvalue dependence on $\beta$)}}\label{rem:dependence-on-beta}
	Notice that the eigenvalue $\mu$ is implicitly dependent on $\beta$:
	\begin{align*}
		\mu
		\triangleq \frac{1}{2}\lambda_{\min}(\bar{A}_T + \bar{A}_T\tr)
		= \frac{1}{2T}\lambda_{\min}\left(2\Phi\tr\Phi - \beta\left({\Phi'}\tr\Phi + \Phi\tr\Phi'\right)\right).
		\end{align*}
		Clearly, as $\beta$ increases, it is harder to satisfy the assumption that $\mu > 0$.
		Moreover, for pathwise LSTD 
		(see Section \ref{sec:td-background}), the inequality in \eqref{eq:def-mu-prime} underlines an \textit{implicit} linear dependence of the rates on the reciprocal of $(1 - \beta)$. However, the bounds' exact sensitivity to this reciprocal is data-dependent.
\end{remark}

\begin{remark}\textbf{\textit{(Regularization)}}
	To obtain the best performance from the batchTD algorithm, we need to know the value of $\mu$.
	However, we can get rid of this dependency easily by explicitly regularizing the problem.  In other words, instead of the LSTD solution \eqref{eq:lstd-solution}, we obtain the following regularized variant:
	\begin{align}
	\label{eq:lstd-solution-reg}
	\hat \theta_T^{reg} = (\bar A_T+\mu I)^{-1} \bar b_T, 
	\end{align}
	where $\mu$ is now a constant set in advance. The update rule for this variant is
	\begin{align}
	\theta_n^{reg} =& (1-\gamma_n\mu)\theta_{n-1} + \gamma_{n} \left(r_{i_n} + \beta \theta_{n-1}\tr \phi(s'_{i_{n}}) - \theta_{n-1}\tr \phi(s_{i_n})\right)\phi(s_{i_n}).
	\label{eq:reg-random-lstd-update}
	\end{align}
	This algorithm retains all the properties of the non-regularized batchTD algorithm, except that it converges to the solution of \eqref{eq:lstd-solution-reg} rather than to that of \eqref{eq:lstd-solution}. In particular, the conclusions of Theorem \ref{thm:flstd-rate} hold without requiring assumption (A1), but measuring $\theta_n-\hat\theta_T^{reg}$, the error to the regularized fixed point $\hat\theta_T^{reg}$.
\end{remark}

\begin{remark}\textbf{\textit{(Computational complexity)}}
		Our theoretical results in Theorem \ref{thm:flstd-rate} show that, with probability $1-\delta$, batchTD constructs an $\epsilon$-approximation of the  pathwise LSTD solution with $O(d\ln(1/\delta)/\epsilon^2)$ complexity. In other words, for the batchTD estimate to be within a distance $\epsilon>0$ of the LSTD solution, the number of iterations of \eqref{eq:random-lstd-update} would be proportional to $\frac{d\ln(1/\delta)}{\epsilon^2}$. This observation coupled with the fact that each iteration of \eqref{eq:random-lstd-update} is of order $O(d)$
		establishes the advantage of batchTD over pathwise LSTD from a time-complexity viewpoint. 
		
		However, batchTD requires storing the entire dataset for the purpose of random sampling. To reduce the storage requirement of batchTD, one could uses mini-batching of the dataset, i.e., store smaller subsets of the dataset and run batchTD updates on these mini-batches. It is an interesting direction for future work to analyze such an approach and recommend appropriate mini-batch sizes based on the parameters of the underlying policy evaluation problem. For the case of regression, such an approach has been recommended in earlier works, cf. \cite{roux2012stochastic}.
	\end{remark}
	
	\begin{remark}\textbf{\textit{(TD with linear function approximation)}}
		\label{remark:td-bounds}
		One could use completely parallel arguments to that in the proof of Theorem \ref{thm:flstd-rate} to obtain rate results for TD(0) with linear function approximation under i.i.d. samples. A similar observation holds for the bounds presented below for the projection-free variant of batchTD in Theorem \ref{thm:flstd-rate-noproj} and for the iterate-averaged variant of batchTD in Theorem \ref{thm:flstd-avg-rate}.
		
		The bounds for TD with linear function approximation under i.i.d. sampling would be a side benefit, while the primary message from our work is that one could run TD(0) on a batch, and obtain a computational advantage, with performance comparable to that of LSTD. We have used pathwise LSTD to drive home this point.  
		
		Finally, note that the regular TD with linear function approximation is under non i.i.d. sampling (or involving a Markov noise component), and deriving non-asymptotic bounds for such a setting is beyond the scope of this paper.  
\end{remark}

\subsection{Projection-free variant of the batchTD algorithm}
\label{sec:noproj-results}
Here we consider a projection-free variant of batchTD that updates according to \eqref{eq:random-lstd-update}, but with $\Upsilon(\theta) = \theta,\ \forall \theta \in \R^d$.
We now present the results for batchTD without a non-trivial projection, under assumptions similar to the projected variant of batchTD, i.e., bounded rewards, features, and a positive lower bound on the minimum eigenvalue $\mu$ of the symmetric part of $\bar A_T$. The results include asymptotic convergence and a bound in expectation on the error $\| \theta_n - \hat \theta_T\|_2$. However, we are unable to derive bounds in high probability without having the iterates explicitly bounded using $\Upsilon$ and it would be a interesting future research direction to get rid of this operator for the bounds in high probability. 

\begin{theorem}
	\label{thm:flstd-asym-noproj}
	Assume (A1)-(A3), and also that the step sizes $\gamma_n \in \R_+$ satisfy $\sum_{n}\gamma_n = \infty$, and $\sum_n\gamma_n^2 <\infty$. Then, for the iterate $\theta_n$ updated according to \eqref{eq:random-lstd-update} without projection (i.e., $\Upsilon$ is the identity map), we have  
	\begin{align}
	\theta_n \rightarrow \hat\theta_T \text{ a.s. as } n \rightarrow \infty.
	\label{eq:theta-asymp-conv-noproj}
	\end{align}
\end{theorem}
\begin{proof}
	See Section \ref{sec:proof-lstd}.
\end{proof}

Using a slightly different proof technique, we are able to give a bound in expectation for the error of the non-projected batchTD. 
\begin{theorem}[\textit{Expectation error bound for batchTD without projection}]
	\label{thm:flstd-rate-noproj}
	\ \\Assume (A2)-(A4). Set $\gamma_n= \frac{c_0c}{(c+n)}$ such that $c_0\in(0,\mu((1+\beta)^2\Phi_{\max}^4)^{-1}]$ and $c_0 c \mu  \in (1,\infty)$. Then, for any $\delta >0$, we have
	\begin{align}
	\E \l \theta_n - \hat \theta_T \r \le \dfrac{K_1(n)}{\sqrt{n+c}},
	\end{align}
	where $K_1(n)$ is a function of order $O(1)$, defined by:
	\begin{gather*}
		K_1(n) \triangleq  \frac{\sqrt{3}\l \theta_0 - \hat\theta_T \r \sqrt{(c+1)^{c_0 c\mu}}}{\sqrt{(n+c)^{c_0 c\mu-1}}}
		+ \frac{ 2\sqrt{3} e c_0 c \left(R_{\max} + (1+\beta)\l \hat\theta_T\r\Phi_{\max}^2\right)}{\sqrt{2c_0 c\mu - 1}}.
		\end{gather*}
\end{theorem}
\begin{proof}
	See Section \ref{sec:expectation-bound-noproj-proof}.
\end{proof}

\subsection{Performance bound} 
\label{sec:pb}
We can combine our error bounds above with the performance bound derived in  \cite{lazaric2012finite} for pathwise LSTD. 
	The theorem below shows that using batchTD in place of pathwise LSTD does not impact the overall convergence rate. 

\begin{theorem}[\textit{Performance bound}]
	\label{thm:perf-bound-lstd}
	Let $\tilde v_n \triangleq \Phi \theta_n$ denote the approximate value function obtained after $n$ steps of batchTD, and let $v$ denote the true value function, evaluated at the states $s_1, \ldots, s_T$ along the sample path. 
		Then, under the assumptions (A1)-(A4), with probability $1-2\delta$ (taken w.r.t. the random path sampled from the MDP, and the randomization in batchTD), we have
		\begin{align}
		\| v - \tilde v_n \|_T 
		\le \underbrace{ \frac{ \| v - \Pi v \|_T }{\sqrt{1-\beta^2}} }_{\textbf{approximation error}}
		+ \underbrace{\frac{\beta R_{\max}\Phi_{\max}}{(1-\beta)}\sqrt{\frac{d}{\mu'}}\left( \sqrt{\dfrac{8\ln{\frac{2d}{\delta}}}{T}} 
			+ \dfrac{1}{T}\right)}_{\textbf{estimation error}}	+ \underbrace{\frac{\Phi_{\max} K_2(n)}{\sqrt{n+c}}}_{\textbf{computational error}}.
		\label{eq:lstd-overall-bounds}
		\end{align}
		where $\| f \|^2_T \triangleq \dfrac1{T}\sum\limits_{i=1}^{T} f(s_i)^2$, for any function $f$ and $\mu'$ is the minimum eigenvalue of $\dfrac1{T}\Phi\tr\Phi$ (see also \eqref{eq:def-mu-prime}).
\end{theorem}
\begin{proof}
	The result follows by combining Theorem \ref{thm:flstd-rate} above with Theorem 1 of \cite{lazaric2012finite} using a triangle inequality.
	\end{proof}

\begin{remark}
	The approximation and estimation errors (first and second terms in the RHS of \eqref{eq:lstd-overall-bounds}) are artifacts of function approximation and least squares methods, respectively.
	The third term is a consequence of using batchTD in place of the LSTD.
	Setting $n=T$ in the above theorem, we observe that using our scheme in place of LSTD does not impact the rate of convergence of the approximate value function $\tilde v_n$ to the true value function $v$.
	Further, the performance bound in Theorem \ref{thm:perf-bound-lstd}, considering only the dimension $d$, minimum eigenvalue $\mu$ and sample size $T$, is of the order $O\left(\frac{\sqrt{d}}{\mu\sqrt{T}}\right)$, which is better than the order $O\left(\frac{d}{\mu T^{1/4}}\right)$ on-policy performance bound for GTD/GTD2 in Proposition 4 of \cite{liu2015finite}.
\end{remark}

\begin{remark}\textit{(Generalization bounds)}\label{rem:generalisation}
	While Theorem \ref{thm:perf-bound-lstd} holds for only states along the sample path $s_1,\ldots, s_T$, it is possible to generalize the result to hold for states outside the sample path. This approach has been adopted in \cite{lazaric2012finite} for regular LSTD and the authors there provide performance bounds over the entire state space assuming a stationary distribution exists for the given policy $\pi$ and the underlying Markov chain is mixing fast (see Lemma 4 in \cite{lazaric2012finite}). In the light of the result in Theorem \ref{thm:perf-bound-lstd} above, it is straightforward to provide generalization bounds similar to Theorems 5 and 6 of \cite{lazaric2012finite} for batchTD as well, and the resulting rates from these generalization bound variants for batchTD are the same as that for regular LSTD. We omit these obvious generalizations, and refer the reader to Section 5 of \cite{lazaric2012finite} for further details.
\end{remark}
\section{Iterate Averaging}
\label{sec:iterate-averaging}
Iterate averaging is a popular approach for which it is not necessary to know the value of the constant $\mu$ (see (A1) in Section \ref{sec:results}) to obtain the (optimal)  approximation error of order $O(n^{-1/2})$. 
Introduced independently by Ruppert \cite{ruppert1991stochastic} and Polyak \cite{polyak1992acceleration}, the idea here is to use a larger step-size $\gamma_n \triangleq c_0\left(c/(c+n)\right)^\alpha$, and then use the averaged iterate, defined as follows:
	\begin{align}
	\bar \theta_{n} \triangleq \frac{1}{n+1} \sum_{i=0}^n \theta_i,
	\end{align}
	where $\theta_n$ is the iterate of the batchTD algorithm, presented earlier.
The following result bounds the the distance of the averaged iterate to the LSTD solution.
\begin{theorem}[\textit{Error Bound for iterate averaged batchTD}]\label{thm:flstd-avg-rate}
	\ \\Assume (A1)-(A4). Set $\gamma_n= c_0\left(\frac{c}{c+n}\right)^\alpha$, with $\alpha \in (1/2,1)$ and $c,c_0>0$. Then, for any $\delta >0$, and any $n>n_0\triangleq \max\{\lfloor\left(\frac{2c_0(1+\beta^2)\Phi_{\max}^4}{\mu})^{1/\alpha} - 1\right)c\rfloor,0\}$, we have
	\begin{align}
	\E \l \bar\theta_n - \hat \theta_T \r &\le \dfrac{K_1^{IA}(n)}{(n+c)^{\alpha/2}},
	\textrm{~~ and } \\
	\P\left( \l \bar\theta_n - \hat \theta_T \r \right.&\left.\le \dfrac{K_2^{IA}(n)}{(n+c)^{\alpha/2}}\right) \ge 1 - \delta \label{eq:l2-prob-bound-lstd-averaging},
	\end{align}
	where
		\begin{align*}
		K_1^{IA}(n)
		\triangleq& C_0\Bigg[C_1 C_2\l \theta_0 - \hat\theta_T\r
		+ \sqrt{e}\left(\frac{2\alpha}{1-\alpha}\right)^{\frac{1}{2(1-\alpha)}}\\
		&\qquad
		+  \underbrace{2c_0 C_1 C_2\left(R_{\max} + (1+\beta)H\Phi_{\max}^2\right)\sqrt{n_0}}_{(E1)}
		\Bigg]\frac{1}{(n+1)(n+c)^{-\frac{\alpha}{2}}}\\
		& +  \underbrace{\left(R_{\max} + (1+\beta)H\Phi_{\max}^2\right) c^\alpha c_0
			\left( 2c_0\mu c^\alpha\right)^{\frac{\alpha}{2(1-\alpha)}}}_{E2},
		\end{align*}
		\begin{gather*}
		C_0 \triangleq \sum_{n=1}^{\infty}\e\left(-c_0\mu c^\alpha(n+c)^{1-\alpha}\right),\ 
		C_1\triangleq\exp\left(2c_0(1+\beta)\Phi_{\max}^2(n_0+1)\right),\\
		C_2\triangleq\e\left(c_0\mu c^\alpha(n_0+c+1)^{1-\alpha}\right), \text{and }
		\end{gather*}
		\begin{align*}
		K_2^{IA}(n)
		\triangleq&\left\{
		\underbrace{ \frac{4\sqrt{\log{\delta^{-1}} }}{\mu^2 c_0^2} 
			\frac{1}{\mu}
			\left[2^\alpha
			+ \left[\frac{2\alpha}{ c_0\mu c^{\alpha}}\right]^{\frac{1}{1-\alpha}}
			+ \frac{2(1 - \alpha)(c_0\mu)^{\alpha}}{\alpha}
			\right]}_{(E3)}
		\right.\\[1ex]
		&\qquad\qquad\quad
		\left. 
		+ \underbrace{\frac{ \sqrt{n_0}e^{(1+\beta)\Phi_{\max}^2c_0 (2n_0+1)}}
			{(1+\beta)\Phi_{\max}^2(n+1)}}_{(E4)}
		\right\}
		\frac{1}{ (n+1)(n+c)^{-\frac{\alpha}{2}} }
		+K_1^{IA}(n).
		\end{align*}
\end{theorem}
\begin{proof}
	The proof of both the high probability bound as well as bound in expectation proceed by splitting the analysis into the error before and after $n_0$. The individual terms in the definition of $K_2^{IA}(n)$
	can be classified based on whether they are bounding the error before or after $n_0$. In particular, the term labelled (E4) in the definition of $K_2^{IA}(n)$ is a bound on the error before $n_0$, while the terms collected under (E3) are a bound on the error after $n_0$. 
	
	While the proof of the bound in expectation involves splitting the analysis before and after $n_0$, the resulting bound via $K_1^{IA}(n)$ does not have a clear split into additive terms that directly correspond to before or after $n_0$. However, from the proof presented later, it is apparent that $C_1$ arises out of a bound on the initial error before $n_0$, the term involving the factor labelled (E1) in the definition of $K_1^{IA}(n)$ arises out of a bound on the sampling error before $n_0$. Further, $C_0$ arises out of a bound on the initial error after $n_0$ and the term labelled (E2) in $K_1^{IA}(n)$ is used to bound the sampling error after $n_0$.
	
	For a detailed proof,  the reader is referred to 
	Section \ref{sec:proof-lstd-avg}.
\end{proof}

A few remarks are in order.
	
	\begin{remark}\textbf{\textit{(Explicit constants)}}
		Unlike \cite{fathi2013transport}, where the authors provide concentration bounds for general stochastic approximation schemes, our results provide an explicit $n_0$, after which the error of iterate averaged batchTD is nearly of the order $O(1/n)$.  
\end{remark}

\begin{remark}\textbf{\textit{(Rate dependence on eigenvalue)}}
	From the bounds in Theorem \ref{thm:flstd-avg-rate}, it is evident that the dependency on the knowledge of $\mu$  for the choice of $c$ can be removed through averaging of the iterates, while obtaining a rate that is close to $1/\sqrt{n}$. In particular, iterate averaging results in a rate that is of the order $O\left(1/n^{(1-\alpha)/2}\right)$, where  the exponent $\alpha$ has to be chosen strictly less than $1$.  Setting $\alpha=1$ causes the constant $C_0$ as well as $K_1^{IA}(n), K_2^{IA}(n)$ to blowup and hence, there is a loss of $\alpha/2$ in the rate, when compared to non-averaged batchTD. However, unlike the latter, iterate averaged batchTD does not need the knowledge of $\mu$ in setting the step size $\gamma_n$. 
\end{remark}

\begin{remark}\textbf{\textit{(Decay rate of initial error)}}
	The bound in expectation in Theorem \ref{thm:flstd-avg-rate} can be re-written as follows:
	\begin{align*}
	\E \l \bar\theta_n - \hat \theta_T \r \le \dfrac{C_0 C_1 C_2 \l \theta_0 - \hat\theta_T\r}{(n+1)} +  \dfrac{const}{(n+c)^{\alpha/2}}.
	\end{align*}
	Thus, the initial error is forgotten at the rate $O(1/n)$ and this is slower than the corresponding rate obtained for the case of non-averaged batchTD (see Remark \ref{remark:flstd-error-split}). 
	Hence, as suggested by earlier works on stochastic approximation (cf. \cite{fathi2013transport}), it is preferred to average after a few iterations since the initial error is not forgotten faster than the sampling error with averaging.
\end{remark}

	\begin{remark}\textbf{\textit{(Computational cost vs. accuracy)}}
		Let $\epsilon, \delta > 0$. Then, the number of iterations $n$ requires to achieve an accuracy $\epsilon$, i.e., $\l \bar\theta_n - \hat \theta_T \r \le \epsilon$ with probability $1-\delta$, is of the order $O\left(\frac{1}{\epsilon^{2/\alpha}} \log \left(\frac{1}{\delta} \right)\right)$. On the other hand, the corresponding number of iterations for the non-averaged case (see Theorem \ref{thm:flstd-rate}) is $O\left(\frac{1}{\epsilon^{2}} \log \left(\frac{1}{\delta} \right)\right)$.
\end{remark}

	\section{Recent works: A comparison}
	\label{sec:relatedworkrecent}
	Non-asymptotic bounds for TD(0) with linear function approximation are derived in three recent works -- see \cite{dalal2018finite,bhandari2018finite,csaba18tdbounds}. In \cite{dalal2018finite,csaba18tdbounds}, the authors consider the i.i.d. sampling case, while the authors in \cite{bhandari2018finite} provide bounds in the i.i.d. as well as the more general Markov noise settings. As noted earlier in Remark \ref{remark:td-bounds}, our analysis could be re-used to derive bounds for TD with linear function approximation in the i.i.d. sampling scenario, while the case of Markov noise is not handled by us. This observation justifies a comparison of the bounds that we derive for batchTD to those in the aforementioned references for TD under i.i.d. sampling, and we provide this comparison below.
	\begin{itemize}
		\item In comparison to the references \citep{bhandari2018finite} and \citep{csaba18tdbounds} listed above, we would like to point out that we derive non-asymptotic bounds that hold with high probability, in addition to bounds that hold in expectation. The aforementioned references provide bounds that hold in expectation only. 
		\item In \citep{bhandari2018finite}, the bound in expectation that we derived in Theorem \ref{thm:flstd-rate} matches the bound derived in \cite{bhandari2018finite}, up to constants. Note that our result in Theorem \ref{thm:flstd-rate}, as well as those in \citep{bhandari2018finite} are for the projected variant of TD(0). In addition, we also provide a bound in expectation in Theorem \ref{thm:flstd-rate-noproj} for the projection-free variant of TD(0).
		\item Continuing the comparison with \cite{bhandari2018finite}, the bounds in their work require the knowlege of the minimum eigenvalue $\mu$, which is unknown in a typical RL setting. We get rid of this problematic eigenvalue dependence through iterate averaging, while obtaining a nearly optimal rate of the order $O\left( n^{\alpha/2}\right)$, where $\frac{1}{2} < \alpha <1$. 
		\item The bounds in \citep{dalal2018finite} are for TD(0) with linear function approximation under the i.i.d. sampling case, allowing a comparison of bounds for batchTD with their results. The bound in expectation on the error $\l \theta_n - \theta^*\r$ in Theorem 3.1 of \citep{dalal2018finite} is $O(\frac{1}{n^\sigma})$, where $0<\sigma<\frac{1}{2}$. Here $\theta_n$ is the TD(0) iterate, and $\theta^{*}$ is the TD fixed point. In contrast, the bound we obtain in Theorem \ref{thm:flstd-asym-noproj} is $O(\frac{1}{\sqrt{n}})$. Both results are for the projection-free variant. However, our bound involves a stepsize that require the knowledge of $\mu$ (see (A1)), while their stepsize is $\Theta(\frac{1}{n^{2\sigma}})$. Our results for the iterate-averaged variant in Theorem \ref{thm:flstd-avg-rate} get rid of this stepsize dependence, and the rate we obtain for this variant are comparable to that in Theorem 3.1 of \citep{dalal2018finite}. An advantage with our bounds is that, unlike \cite{dalal2018finite}, we make all the constants explicit. 
		\item Continuing the comparison with  \citep{dalal2018finite}, we first note that the high-probability bound in \ref{thm:flstd-rate} in our work, which is for the case when $\mu$ is known, has a rate of order $O\left(\frac{1}{\sqrt{n}}\right)$, while the iterate averaged variant in Theorem \ref{thm:flstd-avg-rate} exhibits a rate $O\left(\frac{1}{n^{\alpha/2}}\right)$, where $0 < \alpha < \frac{1}{2}$.  On the other hand, we find it difficult to infer the rate from the bounds in Theorem 3.6 of \citep{dalal2018finite}, as the same depends on a parameter $\lambda$ that is below the minimum eigenvalue (which is $\mu$ in our notation). Further, our high probability bound in Theorem \ref{thm:flstd-rate} applies for all $n$, while that in Theorem \ref{thm:flstd-avg-rate} is for all $n\ge n_0$, with $n_0$ explicitly specified (as a function of the underlying parameters). In contrast, the bound in Theorem 3.6 of \citep{dalal2018finite} applies to sufficiently large $n$, where the threshold beyond which the bound applies is not explicitly specified. Finally, we project the iterates to keep it bounded, while the bounds in \cite{dalal2018finite} do not involve a projection operator. Note that we require projection for the high-probability bounds, while we derive a bound in expectation for the projection-free variant (see Theorem \ref{thm:flstd-rate-noproj}).
		\item In \citep{csaba18tdbounds}, the authors derive non-asymptotic bounds in expectation, which could be applied for TD(0) with linear function approximation, or even to our batchTD algorithm. The authors in \citep{csaba18tdbounds} derive lower bounds, while we focus on Theorem 1, which contains the upper bound. Our bound in expectation in Theorem \ref{thm:flstd-rate} is comparable to that in Theorem 1 there, since the overall rate is $O(\frac{1}{\sqrt{n}})$ in either case, and both results assume knowledge about underlying dynamics (through the minimum eigenvalue $\mu$ in our case, while through a certain distribution constant for setting the stepsize there). Further, unlike \cite{csaba18tdbounds}, we derive bounds for the iterate-averaged variant, which gets rid of the problematic stepsize dependence, at a compromise in the rate, which turns out to be $O(\frac{1}{n^{\alpha}})$, with $\alpha < \frac{1}{2}$.
\end{itemize}

\section{Fast LSPI using batchTD {\em\textmd(fLSPI)}}
\label{sec:flspi}
LSPI \cite{lagoudakis2003least} is a well-known algorithm for control based on the policy iteration procedure for MDPs. We propose a computationally efficient variant of LSPI, which we shall henceforth refer to as fLSPI. The latter algorithm works by substituting the regular LSTDQ with batchTDQ --- an algorithm that is quite similar to batchTD described earlier. We first briefly describe the LSPI algorithm and later provide a detailed description of fLSPI.

\subsection{Background for LSPI}
We are given a set of samples $\D\triangleq\{(s_i,a_i,r_i,s'_{i}),i=1,\ldots,T)\}$, where each sample $i$ denotes a one-step transition of the MDP  from state $s_i$ to $s'_i$ under action $a_i$, while resulting in a reward $r_i$. The objective is to find an {\em approximately optimal} policy using this set. This is in contrast with the goal of LSTD, which aims to approximate the state-value function of a particular policy (see Section \ref{sec:td-background}).

For a given stationary policy $\pi$, the Q-value function $Q^\pi(s,a)$ for any state $s\in \S$ and action $a\in\A(\S)$ is defined as follows:
 \begin{equation}
	Q^\pi(s,a) \triangleq \E\left[\sum_{t=0}^{\infty} \beta^t r(s_t,\pi(s_t))\mid s_0=s,a_0=a\right].
	\label{eq:qf}
	\end{equation}
In the above, the initial state $s$ and the action $a$ in $s$ are fixed, and thereafter the actions taken are governed by the policy $\pi$. 
This function can be thought of as the value function for a policy $\pi$ in state $s$, given that the first action taken is the action $a$.
As before, we parameterize the Q-value function using a linear function approximation architecture,
\begin{align}
Q^\pi(s,a) \approx \theta\tr \phi(s,a),
\end{align}
where $\phi(s,a)$ is a $d$-dimensional feature vector corresponding to the tuple $(s,a)$ and $\theta$ is a tunable policy parameter.

LSPI is built in the spirit of policy iteration algorithms. These perform policy evaluation and policy improvement in tandem.
For the purpose of policy evaluation, LSPI uses a LSTD-like algorithm called LSTDQ, which learns an approximation to the Q- (state-action value) function. It does this for any policy $\pi$, by solving the linear system 
\begin{align}
\label{eq:lstdq}
&\hat\theta_T = \bar A_T^{-1} \bar b_T, \text{ where}\\
&\bar A_T = \dfrac{1}{T}\sum_{i=1}^{T} \phi(s_i,a_i)(\phi(s_i,a_i) - \beta \phi(s'_i,\pi(s'_i)))\tr, \text{ and }\bar b_T =  T^{-1} \sum_{i=1}^{T} r_i \phi(s_i,a_i).\nonumber
\end{align}  
As in the case of LSTD, the above can be seen as approximately solving a system of equations similar to \eqref{eq:td-system-of-equations}, but in this case for the Q-value function. The pathwise LSTDQ variant is obtained by forming the dataset $\D$ from a sample path of the underlying MDP for a given policy $\pi$ and also zeroing out the feature vector of the next state-action tuple in the last sample of the dataset.

The policy improvement step uses the approximate Q-value function to derive a greedily updated policy as follows:
$$\pi'(s) =\argmax_{a\in\A} {\theta}\tr \phi(s,a).$$    
Since this policy is provably better than $\pi$, iterating this procedure allows LSPI to find an approximately optimal policy.

\subsection{fLSPI Algorithm}
The fLSPI algorithm works by substituting the regular LSTDQ with its computationally efficient variant batchTDQ. The overall structure of fLSPI is given in Algorithm \ref{alg:lspiSalstda}.

For a given policy $\pi$, batchTDQ approximates LSTDQ solution \eqref{eq:lstdq} by an iterative update scheme as follows (starting with an arbitrary $\theta_0$):
\begin{align}
\theta_k = \theta_{k-1} + \gamma_{k} \left(r_{i_k} + \beta \theta_{k-1}\tr \phi(s'_{i_{k}},\pi(s'_{i_{k}})) - \theta_{k-1}\tr \phi(s_{i_k},a_{i_k})\right)\phi(s_{i_k},a_{i_k})
\label{eq:flstdq}
\end{align}
From Section \ref{sec:flstd}, it is evident that the claims in Proposition \ref{prop:main} and Theorem \ref{thm:flstd-rate} hold for the above scheme as well.

\algblock{PEval}{EndPEval}
\algnewcommand\algorithmicPEval{\textbf{\em Policy Evaluation}}
\algnewcommand\algorithmicendPEval{}
\algrenewtext{PEval}[1]{\algorithmicPEval\ #1}
\algrenewtext{EndPEval}{\algorithmicendPEval}

\algblock{PImp}{EndPImp}
\algnewcommand\algorithmicPImp{\textbf{\em Policy Improvement}}
\algnewcommand\algorithmicendPImp{}
\algrenewtext{PImp}[1]{\algorithmicPImp\ #1}
\algrenewtext{EndPImp}{\algorithmicendPImp}

\algtext*{EndPEval}
\algtext*{EndPImp}

\begin{algorithm}[h]
	\caption{fLSPI}
	\label{alg:lspiSalstda}
	\begin{algorithmic}
		\State {\bfseries Input:} Sample set $D\triangleq\{s_i,a_i,r_i, s'_i\}_{i=1}^{T}$, obtained from an initial (arbitrary) policy.
		\State {\bfseries Initialization:} $\epsilon$, $\tau$, step-sizes $\{\gamma_k\}_{k=1}^{\tau}$, initial policy $\pi_0$ (given as $\theta_0$). 
		\State $\pi \leftarrow \pi_0$, $\theta \leftarrow \theta_0$.
		\Repeat
		\PEval
		\State Approximate LSTDQ$(D,\pi)$ using batchTDQ$(D,\pi)$ as follows:      
		\For{$k=1 \dots \tau$}
		\State Get random sample index: $i_k\sim U(\{1,\dots,T\})$.
		\State Update batchTDQ iterate $\theta_k$ using \eqref{eq:flstdq}.	      
		\EndFor
		\EndPEval    
		\State $\theta' \leftarrow \theta_\tau$, $\Delta = \l\theta - \theta'\r$.
		\PImp 	  
		\State Obtain a greedy policy $\pi'$ as follows:
		$\pi'(s) =\argmax_{a\in\A} {\theta'}\tr \phi(s,a)$.        
		\EndPImp   
		\State $\theta \leftarrow \theta'$, $\pi \leftarrow \pi'$.
		\Until{$\Delta < \epsilon$}
	\end{algorithmic}
\end{algorithm}

\begin{remark}
	Error bounds for fLSPI can be derived along the lines of those for regular on-policy LSPI in \cite{lazaric2012finite}, and we omit the details.
\end{remark}

\section{Convergence proofs}
\label{sec:analysis}

	Let $\F_n$ denotes the $\sigma$-field generated by $\theta_0,\ldots,\theta_n$, $n\ge 0$. Let
	\begin{align}
	f_n(\theta)\triangleq \left(r_{i_n} + \beta \theta\tr \phi(s'_{i_{n}}) - \theta\tr \phi(s_{i_n})\right)\phi(s_{i_n}).\label{eq:fn}
	\end{align}
	Recall that we denote the current state feature $(T\times d)$-matrix by $\Phi \triangleq (\phi(s_1)\tr,\dots, \phi(s_T))$, the next state feature $(T\times d)$-matrix by $\Phi' \triangleq (\phi(s_1')\tr,\dots, \phi(s_T'))$, and the reward $(T\times 1)$-vector by ${\cal R} = (r_1,\dots,r_T)\tr$. Recall also that the LSTD solution is given by
	\begin{align*}
	\hat \theta_T = \bar A_T^{-1} \bar b_T,
	\text{ where }
	\bar A_T = \frac{1}{T}(\Phi\tr\Phi - \beta\Phi\tr\Phi')
	\text{ and }
	\bar b_T = \frac{1}{T}\Phi\tr\cal R.
	\end{align*}
	Finally we note also that the pathwise LSTD solution has the same form as above, except that $\Phi' \triangleq \hat P \Phi = (\phi(s_1')\tr,\dots, \phi(s_{T-1}')\tr,\bf{0}\tr)$, where $\bf{0}$ is the $d\times 1$ zero-vector.

\subsection{Proof of asymptotic convergence}
\label{sec:proof-asymp}
\paragraph{\textbf{Proof of Theorem \ref{thm:flstd-asym-noproj} (batchTD without projection):}}
\begin{proof}
	We first rewrite \eqref{eq:random-lstd-update} as follows:
	\begin{align}
	\label{eq:f1}
	\theta_n = \theta_{n-1} + \gamma_{n} \left( -\bar A_T\theta_{n-1} + \bar b_T + \Delta M_n\right) ,
	\end{align}
	where $\Delta M_n = f_n(\theta_{n-1}) - \E(f_n(\theta_{n-1})\mid \F_{n-1})$ is a martingale difference sequence, with $f_n(\cdot)$ as defined in \eqref{eq:fn}.
	
	The ODE associated with \eqref{eq:f1} is
		\begin{align}
		\dot \theta(t) =  q(\theta(t)), t\ge 0.  \label{eq:todep}
		\end{align} 
		In the above, $q(\theta(t)) \triangleq -\bar A_T \theta(t) + \bar b_T$.
	
	To show that $\theta_n$ converges a.s. to $\hat \theta_T$, one requires that the iterate $\theta_n$ remains bounded a.s. Both boundedness and convergence can be inferred from Theorems 2.1-2.2(i) of \cite{borkar2000ode}, provided we verify assumptions (A1)-(A2) there. These assumptions are as follows:
	
	\noindent \textbf{(a1)} The function $q$ is Lipschitz. For any $\eta \in \R$, define $q_{\eta}(\theta) = q(\eta \theta)/\eta$. Then, there exists a continuous function $q_\infty$ such that $q_{\eta} \rightarrow q_{\infty}$ as $\eta \rightarrow \infty$ uniformly on compact sets. Furthermore, the origin is a globally asymptotically stable equilibrium for the ODE
	\begin{align}\label{eq:asym-q}
	\dot \theta(t) =  -q_\infty(\theta(t)).
	\end{align}
	
	\noindent \textbf{(a2)} The martingale difference $\{\Delta M_{n}, n\ge 1\}$ is square-integrable with 
	$$\E [\l \Delta M_{n+1} \r^2 \mid \F_n] \le C_0 (1 + \l \theta_n \r^2), \ n\ge 0,$$
	for some $C_0 < \infty$.
	
	We now verify (a1) and (a2) in our context. Notice that
	$q_\eta(\theta)\triangleq-\bar A_T \theta + \bar b_T/\eta$ converges to $q_{\infty}(\theta(t))=-\bar A_T \theta(t)$ as $\eta\rightarrow \infty$.
	Since the matrix $\bar A_T$ is positive definite by (A1), the aforementioned ODE has the origin as its globally asymptotically stable equilibrium. This verifies (a1).
	
	For verifying (a2), notice that 
		\begin{align*}
		\E [\l \Delta M_{n+1} \r^2 \mid \F_n]  \le &\E [\l f_{n+1}(\theta_2) \r^2 \mid \F_n]\\
		\le &  (R_{\max}\Phi_{\max}+(1+\beta)\Phi^2_{\max} \l \theta_n \r)^2
		\end{align*}
	The first inequality follows from the fact that for any scalar random variable $Y$, \\$\E\left(Y -  E\left[Y\mid\F_n\right]\right)^2 \le \E Y^2$, while the second inequality follows from (A2) and (A3).
	The claim follows.
	\end{proof}

\paragraph{\textbf{Proof of Theorem \ref{thm:flstd-asym} (batchTD with projection):}}
\begin{proof}
	We first rewrite \eqref{eq:random-lstd-update} as follows:
	\begin{align}
	\label{eq:f}
	\theta_n = \Upsilon\left(\theta_{n-1} + \gamma_{n} \left( -\bar A_T\theta_{n-1} + \bar b_T + \Delta M_n\right) \right),
	\end{align}
	where $\Delta M_n$, $\F_n$ 
	and $f_n(\theta)$ are as defined in \eqref{eq:fn}.
	
	From (A3) and the fact that the iterate $\theta_n$ is projected onto a compact and convex set $\C$, it is easy to see that the norm of the martingale difference $\Delta M_n$ is upper bounded by $2\left(R_{\max}\Phi_{\max}+(1+\beta)H\Phi_{\max}^2\right)$.  Thus, \eqref{eq:f} can be seen as a discretization of the ODE
	\begin{align}
	\dot \theta(t) = \check\Upsilon( - \bar A_T \theta(t) + \bar b_T), \ t\ge 0, \label{eq:tode}
	\end{align} 
	where $\check\Upsilon(\theta)  = \lim_{\tau \rightarrow 0}
	\left[\left(\Upsilon\left(\theta + \tau f(\theta)\right) - \theta\right)/\tau\right]$
	, for any bounded continuous $f$. The operator $\check\Upsilon$ ensures that $\theta$ governed by \eqref{eq:tode} evolves within the set $\C$ that contains $\hat \theta_T$.  As in the proof of Lemma 4.1 in \cite{yu2015convergence}, we have
	$$ 0 = \langle \hat\theta_T, -\bar A_T \hat\theta_T + \bar b_T \rangle 
	\le -\mu \l \hat\theta_T \r^2 + \l \bar b_T \r \l \hat\theta_T \r,$$
	where the inequality follows from (A1). From the foregoing, we have that $\l \hat \theta_T \r \le \frac{\l \bar b_T\r}{\mu} < H \Rightarrow \hat \theta_T \in \C$. Following similar arguments as before, it can be inferred that at any boundary point $\theta$ of $\C$, $\langle \theta, -\bar A_t \theta + \bar b_T \rangle < 0$ and hence the ODE \eqref{eq:tode} has the origin as its globally asymptotically stable equilibrium. The claim now follows from Theorem 2 in Chapter 2 of \cite{borkar2008stochastic} (or even Theorem 5.3.1 on pp. 191-196 of \cite{kushner-clark}).
	\end{proof}

\subsection{Proofs finite-time error bounds for batchTD}
\label{sec:proof-lstd}
To obtain high probability bounds on the computational error $\| \theta_n - \hat \theta_T \|_2$, we consider separately the deviation of this error from its mean (see \eqref{eq:lstd-high-prob-bound} below), and the size of its mean itself (see \eqref{eq:lstd-expectation-bound} below). In this way the first quantity can be directly decomposed as a sum of martingale differences, and then a standard martingale concentration argument applied, while the second quantity can be analyzed by unrolling iteration \eqref{eq:random-lstd-update}.

Proposition \ref{prop:main} below gives these results for general step sequences. The proof involves two martingale analyses, which also form the template for the proofs for the least squares regression extension (see Section \ref{sec:random-batch}), and the iterate averaged variant of batchTD (see Theorem \ref{thm:flstd-avg-rate}).

After proving the results for general step sequences, we give the proof of Theorem \ref{thm:flstd-rate}, which gives explicit rates of convergence of the computational error in high probability for a specific choice of step sizes.

\begin{proposition}
	\label{prop:main}
	Let $z_n = \theta_n - \hat \theta_T$, where $\theta_n$ is given by \eqref{eq:random-lstd-update}.
	Under (A1)-(A4),  we have $\forall \epsilon > 0$,
	\begin{itemize}
		\item[(1)] a bound in \textbf{\textit{high probability}} for the \textbf{\textit{centered error}}:
		\begin{align}
		\!\!\P &\left( \l z_n \r - \E \l z_n \r \ge \epsilon \right)\le  \e\left(- \dfrac{\epsilon^2}{4 \left(R_{\max}+(1+\beta)H\Phi_{\max}^2\right)^2\sum\limits_{k=1}^{n} L^2_k}\right),
		\label{eq:lstd-high-prob-bound}
		\end{align}
		where $L_k \triangleq \gamma_k \prod_{j=k+1}^{n} (1 -  \gamma_{j} (2\mu- \gamma_{j}(1+\beta)^2\Phi_{\max}^4))^{1/2}$,\\[1ex]
		\item[(2)]and a bound in \textbf{\textit{expectation}} for the \textbf{\textit{non-centered error}}:
			\begin{align}
			\label{eq:lstd-expectation-bound}
			&\E\left(\l z_n\r\right)^2
			\le \underbrace{\left[\prod_{k = 1}^n
				\left(1 - \gamma_k(2\mu - \gamma_k(1+\beta)^2\Phi_{\max}^4)\right)
				\l z_0\r\right]^2}_{\textbf{initial error}}\\
			&+  \underbrace{4\sum_{k=1}^{n}\gamma_k^2
				\left[\prod_{j = k}^{n-1}
				(1 - \gamma_j(2\mu  - \gamma_j(1+\beta)^2\Phi_{\max}^4) \right]^2
				\left(R_{\max}+(1+\beta) H \Phi_{\max}^2\right)^2}_{\textbf{sampling error}}\nonumber.
			\end{align}
	\end{itemize}
\end{proposition}
As mentioned earlier, the initial error relates to the starting point $\theta_0$ of the algorithm, while the sampling error arises out of a martingale difference sequence (see Step 1 in Section \ref{sec:expectation-prop-proof} below for a precise definition).

We establish later, in Section \ref{sec:flstd-rate-derive}, that under a suitable choice of step sizes, the initial error is forgotten faster than the sampling error.

	We claim that the terms of the form $1 - \gamma_j(2\mu - \gamma_j \Phi_{\max}^4(1 + \beta)^2)$, which go into a product in the Lipschitz constant $L_i$ as well as in the initial/sampling error terms of the expectation bound, are positive. This claim can be seen as follows: 
	\begin{align}
	1 - \gamma_j\left(2\mu - \gamma_j \Phi_{\max}^4(1 + \beta)^2\right)& \ge 1 - 2 \gamma_j(1+\beta)\Phi_{\max}^2 + \gamma_j^2 \Phi_{\max}^4(1 + \beta)^2\nonumber\\
	& = \left( 1 - \gamma_j(1+\beta)\Phi_{\max}^2 \right)^2 \ge 0, \label{eq:liinc}
	\end{align}
	where the inequality above follows from the fact that $\mu \le (1+\beta)\Phi_{\max}^2$. 
	
	In Section \ref{sec:flstd-rate-derive}, to establish the rates of Theorem \ref{thm:flstd-rate}, we first prove that $\sum_{i=1}^n L_i$ is an order $1/n$ term and the claim of positivity of $L_i$ is necessary for the aforementioned proof.

\subsubsection{\textbf{Proof of Proposition \ref{prop:main} part (1)}}
\begin{proof}
	The proof gives a martingale analysis of the centered computational error. It proceeds in three steps:\ \\[1ex]
	\noindent{\bf Step 1: (Decomposition of error into a sum of martingale differences)}\ \\[1ex]
	Recall that $z_n\triangleq \theta_n - \hat\theta_T$.
	We rewrite {$\l z_n \r - \E \l z_n \r$} as follows:
	\begin{align}
	\l z_n \r - \E \l z_n \r
	= \sum\limits_{k=1}^{n} \left(g_k -  g_{k-1} \right)
	= \sum\limits_{k=1}^{n} D_k, 
	\label{eq:prob-equivalence}
	\end{align}
	where $g_k \triangleq \E [\l z_n\r \left| \F_k \right.]$, $D_k \triangleq g_k - \E [ g_k \left| \F_{k-1} \right.]$, and $\F_k$ denotes the $\sigma$-field generated by the random variables $\{\theta_i,i\le k\}$ for $k \ge 0$.
		
		Recall that $f_k(\theta) \triangleq (\theta\tr \phi(s_{i_k}) - (r_{i_k} + \beta \theta \tr \phi(s_{i_{k}}')  ))\phi(s_{i_k})$ denotes the random innovation at time $k$ given that $\theta_{k-1} = \theta$. 
	\ \\[1ex]
	\noindent{\bf Step 2: (Showing that $g_k$ is a Lipschitz function of the random innovation $f_k$)\footnote{For notational convenience, we have not chosen to make the dependence of $g_k$ on the random innovation $f_k$ explicit. The Lipschitzness of $g_k$ as a function of $f_k$ is clear from equation \eqref{eq:gk-lip} presented below.}}\\[1ex]
	The next step is to show that the functions $g_k$ are Lipschitz continuous in the random innovation at time $k$, with Lipschitz constants $L_k$.
	It then follows immediately that the martingale difference $D_k$ is a Lipschitz function of the $k^{th}$ random innovation with the same Lipschitz constant, which is the property leveraged in Step 3 below.
	In order to obtain Lipschitz constants with no exponential dependence on the inverse of $(1-\beta)\mu$ we depart from the general scheme of \cite{frikha2012concentration}, and use our knowledge of the form of the random innovation $f_k$ to eliminate the noise due to the rewards between time $k$ and time $n$:
	
	Let $\Theta_j^k(\theta)$ denote the value of the random iterate at instant $j$ evolving according to \eqref{eq:random-lstd-update} and beginning from the value $\theta$ at time $k$.
	
	First we note that as the projection, $\Upsilon$, is non-expansive,
	\begin{align*}
	\E&\left(\l\Theta_{j}^k(\theta) - \Theta_{j}^k(\theta')\r\mid \F_{j-1}\right)\\
	&\le\E\left(  \l \Theta_{j-1}^k(\theta) - \Theta_{j-1}^k(\theta')
	- \gamma_{j} [ f_{j}(\Theta_{j-1}^k(\theta))
	- f_{j}(\Theta_{j-1}^k(\theta')) ]\r\mid \F_{j-1}\right).
	\end{align*} 
	Expanding the random innovation terms, we have
	\begin{align}
	&\Theta_{j-1}^k(\theta) - \Theta_{j-1}^k(\theta')
	- \gamma_{j} [ f_{j}(\Theta_{j-1}^k(\theta)) - f_{j}(\Theta_{j-1}^k(\theta')) ]\nonumber\\
	&= \Theta_{j-1}^k(\theta) - \Theta_{j-1}^k(\theta') 	- \gamma_{j} [\phi(s_{i_{j}})\phi(s_{i_{j}})\tr
	- \beta\phi(s_{i_{j}})\phi(s_{i_{j}}')\tr]
	(\Theta_{j-1}^k(\theta) - \Theta_{j-1}^k(\theta'))\nonumber \\
	&=  [I - \gamma_{j} a_j ] (\Theta_{j-1}^k(\theta) - \Theta_{j-1}^k(\theta')), \label{eq:L-equality} 
	\end{align}
	where $a_{j}\triangleq[\phi(s_{i_{j}})\phi(s_{i_{j}})\tr - \beta\phi(s_{i_{j}})\phi(s_{i_{j}}')\tr]$. Note that
	\begin{align*}
	a_{j}\tr a_{j}
	&=\phi(s_{i_j}) \phi(s_{i_j})\tr \phi(s_{i_j}) \phi(s_{i_j})\tr \\
	& - \beta \left(\phi(s_{i_j}) \phi(s_{i_j})\tr \phi(s_{i_j}) \phi(s_{i_j}')\tr
	+ \phi(s_{i_j}') \phi(s_{i_j})\tr \phi(s_{i_j}) \phi(s_{i_j})\tr\right)\\
	& +\beta^2 \phi(s_{i_j}') \phi(s_{i_j}) \tr\phi(s_{i_j}) \phi(s_{i_j}')\tr\\
	&= \l \phi(s_{i_j}) \r^2
	\left[ \phi(s_{i_{j}})\phi(s_{i_{j}})\tr\right.\\
	&\qquad\qquad\left.
	- \beta(\phi(s_{i_{j}})\phi(s_{i_{j}}')\tr+ \phi(s_{i_{j}}')\phi(s_{i_{j}})\tr)
	+ \beta^2 \phi(s_{i_{j}}')\phi(s_{i_{j}}')\tr\right].
	\end{align*}
	Recall that $\Phi\tr \triangleq (\phi(s_1),\dots,\phi(s_T))$ and ${\Phi'}\tr \triangleq (\phi(s_1)',\dots,\phi(s_T)')$. 
		Let
		$\Delta \triangleq \rm{diag}( \l \phi(s_1) \r^2, \dots,$ $ \l \phi(s_{T}) \r^2  )$.
	Then, for any vector $\theta$, we have 
	\begin{align*}
	\E&\left(\theta\tr
	\left(I - \gamma_j a_{j}\right)\tr\left(I - \gamma_j a_{j} \right)
	\theta
	\mid \F_{j-1}\right)\\
	&=\theta\tr
	\E(I - \gamma_{j}[a_{j}\tr + a_{j} -\gamma_{j}a_{j}\tr a_{j}])
	\theta
	\mid \F_{j-1})\\
	&= \l\theta\r^2
	- \gamma_{j}\theta\tr\frac{1}{T}\left[2\Phi\tr\Phi - \beta\left(\Phi\tr\Phi' + {\Phi'}\tr\Phi\right)\right.\\
	&\qquad\qquad\qquad\qquad\left.
	-\gamma_j \left(\Phi\tr\Delta\Phi-\beta\left({\Phi'}\tr\Delta\Phi+ \Phi\tr\Delta\Phi'\right)+\beta^2{\Phi'}\tr\Delta\Phi'\right)\right]\theta\stepcounter{equation}\tag{\theequation}\label{eq:badlabel0}\\
	& \le \l\theta\r^2
	- \gamma_{j} 2\mu \l\theta\r^2 + \gamma_j^2\theta\tr\frac{1}{T}\left(\Phi\tr\Delta\Phi-\beta\left({\Phi'}\tr\Delta\Phi+ \Phi\tr\Delta\Phi'\right)\right)\theta + \beta^2 \l\theta\r^2 \Phi_{\max}^4\stepcounter{equation}\tag{\theequation}\label{eq:badlabel1}\\  
	&\le(1 - \gamma_j(2\mu - \gamma_j \Phi_{\max}^4 (1+\beta)^2))\l\theta\r^2.\stepcounter{equation}\tag{\theequation}\label{eq:badlabel2}
	\end{align*}
	For the equality in \eqref{eq:badlabel0}, we have used that $\sum_{k = 1}^T\phi(s_k)\phi(s_k)\tr = \Phi\tr\Phi$ and similar identities.
	Further, the inequality in \eqref{eq:badlabel1} can be inferred using the following fact: 
	\begin{align*}
	\lambda_{\min}\left(2\Phi\tr\Phi - \beta\left({\Phi'}\tr\Phi+ \Phi\tr\Phi'\right)\right)
	&= \lambda_{\min}\left((\Phi\tr\Phi - \beta{\Phi'}\tr\Phi )
	+ (\Phi\tr\Phi - \beta{\Phi'}\tr\Phi )\tr\right)
	\\ & =\lambda_{\min}\left( T\left(\bar A_T + \bar A_T\tr\right)\right)	\ge 2T\mu,
	\end{align*}
	where we have used assumption (A1) for the last inequality above.
	The last term in \eqref{eq:badlabel1} follows from $|\theta\tr {\Phi'}\tr\Delta\Phi' \theta | \le \l\theta\r^2\Phi_{\max}^4$, where we have used assumption (A2) that ensures features are bounded.
	The inequality in \eqref{eq:badlabel2} can be inferred as follows:
	\begin{align*}
		&|\theta\left( {\Phi'}\tr\Delta\Phi+ \Phi\tr\Delta\Phi'\right)\theta| \le 2\l \theta \r^2 \Phi_{\max}^4 \\
		\Rightarrow &  -2\l \theta \r^2 \Phi_{\max}^4 \le \theta\tr \left( {\Phi'}\tr\Delta\Phi+ \Phi\tr\Delta\Phi'\right) \theta \\
		\Rightarrow & \,\, \theta\tr (\Phi\tr\Delta\Phi  -  \beta \left( {\Phi'}\tr\Delta\Phi+ \Phi\tr\Delta\Phi'\right)  +  \beta^2{\Phi'}\tr\Delta\Phi' )\theta \\
		&\le \l \theta \r^2 (1 + 2\beta + \beta^2) \Phi_{\max}^4 =  (1+\beta)^2 \Phi_{\max}^4 \l \theta \r^2. 
		\end{align*}
	In the above, we have used the boundedness of features to infer  $|\theta\tr \Phi\tr\Delta\Phi \theta | \le \l\theta\r^2\Phi_{\max}^4$ and $|\theta\tr {\Phi'}\tr\Delta\Phi' \theta | \le \l\theta\r^2\Phi_{\max}^4$. 
	
	Hence, from the tower property of conditional expectations, it follows that:
	\begin{align}\label{eq:Theta}
	&\E\left[\l \Theta_{n}^k(\theta) - \Theta_{n}^k(\theta')  \r^2 \right]
	=\E\left[\E\left(\l \Theta_{n}^k(\theta) - \Theta_{n}^k(\theta')  \r^2
	\mid \F_{n-1} \right)\right]\nonumber\\
	&\le  \left(1 - \gamma_{n}\left(2\mu- \gamma_{n}\Phi_{\max}^4 (1+\beta)^2\right)\right)
	\E\left[\l\Theta_{n-1}^k(\theta) - \Theta_{n-1}^k(\theta')\r^2\right]\nonumber\\
	&\le  \left[\prod_{j = k+1}^n
	\left(1 - \gamma_{j} \left(2\mu -
	\gamma_{j}\Phi_{\max}^4 (1 + \beta)^2 \right) \right) \right]
	\l\theta- \theta'\r^2
	\end{align}
	
	Finally, writing $f$ and $f'$ for two possible values of the random innovation at time $k$, and writing $\theta = \theta_{k-1} +\gamma_k f$ and $\theta' = \theta_{k-1} +\gamma_k f'$ and using Jensen's inequality, we have that
	\begin{align}
	&\left| \E\left[ \l \theta_n - \hat\theta_T \r \left| \theta_{k} = \theta\right.\right]\right.
	\left.- \E\left[ \l \theta_n - \hat\theta_T \r \left| \theta_{k} = \theta' \right.\right] \right|\nonumber\\
	&\le  \E\left[ \l \Theta^k_n\left(\theta\right) - \Theta^k_n\left(\theta'\right) \r\right]
	\le   L_k \l f - f'\r,\label{eq:gk-lip}
	\end{align}
	which proves that the functions $g_k$ are $L_k$-Lipschitz in the random innovations at time $k$.
	Recall that  $D_k=g_k - g_{k-1}$, and hence, the Lipschitz constant of $D_k$ is $\max\left(L_k,L_{k-1}\right)$. However, from \eqref{eq:liinc}, we have $L_k > L_{k-1}$, leading to a Lipschitz constant of $L_k$ for $D_k$.
	\ \\[1ex]
	\noindent{\bf Step 3: (Applying a sub-Gaussian concentration inequality)}\\[1ex]
	Now we derive a standard martingale concentration bound in the lemma below. Note that, for any $\lambda>0$,
	\begin{align*}
	\P(   \l z_n \r - \E \l z_n \r \ge \epsilon )
	& =  \P\left(  \sum\limits_{k=1}^{n} D_k  \ge \epsilon \right)
	\le \e(-\lambda \epsilon)
	\E\left(\e\bigg(\lambda \sum\limits_{k=1}^{n} D_k\bigg)\right) \\
	&=  \e(-\lambda \epsilon)
	\E\left(\e\bigg(\lambda \sum\limits_{k=1}^{n-1} D_k\bigg)
	\E\bigg(\e(\lambda D_n) \left| \F_{n-1}\right.\bigg)\right).
	\end{align*}
	The last equality above follows from \eqref{eq:prob-equivalence}, while the first inequality follows from Markov's inequality.
	
		Let $Z$ be a zero-mean random variable (r.v) satisfying $|Z|\le B$ w.p. $1$, and $g$ be a $L$-Lipschitz function $g$. Letting $Z'$ denote an independent copy of $Z$ and $\varepsilon$ a Rademacher r.v., we have 
		\begin{align}
		\E \left( \e\left(\lambda g(Z)\right)\right) &= \E \left( \e\left(\lambda \left( g(Z) - \E(g(Z') \right)\right)\right) \nonumber\\
		& \le \E \left( \e\left(\lambda \left( g(Z) - g(Z') \right)\right)\right) \label{eq:line1}\\
		& = \E \left( \e\left(\lambda \varepsilon \left( g(Z) - g(Z') \right)\right)\right) \label{eq:line2}\\
		& \le \E \left( \e\left(\lambda^2  \left( g(Z) - g(Z') \right)^2/2\right)\right) \label{eq:line3}\\
		& \le \E \left( \e\left(\lambda^2  L^2\left( Z - Z' \right)^2/2\right)\right) \label{eq:line4}\\
		& \le \e\left( \lambda^2 B^2 L^2 / 2 \right).\label{eq:line5}
		\end{align}
		In the above, we have used Jensen's inequality in \eqref{eq:line1}, the fact that distribution of $g(Z)-g(Z')$ is the same as $\varepsilon(g(Z)-g(Z'))$ in \eqref{eq:line2}, a result from Example 2.2 in \cite{wainwright2019high} in \eqref{eq:line3}, the fact that $g$ is $L$-Lipschitz in \eqref{eq:line4}, and the boundedness of $Z$ in \eqref{eq:line5}.
	
	Note that by (A3), and the projection step of the algorithm, we have that $|f_k(\theta_{k-1})|<(R_{\max}+(1+\beta)H\Phi_{\max}^2 )$ is a bounded random variable, and, conditioned on $\F_{k-1}$, $D_k$ is Lipschitz in $f_k(\theta_{k-1})$ with constant $L_k$. So we obtain
	\begin{align*}
	\E\left(\e(\lambda D_n) \left| \F_{n-1}\right.\right)
	\le \e\left(\frac{\lambda^2\left(R_{\max}+(1+\beta)H \Phi_{\max}^2 \right)^2 L_n^2}{2}\right),
	\end{align*}
	and so
	\begin{align}
	\P(   \l z_n \r - \E \l z_n \r \ge \epsilon )
	\le \e(-\lambda \epsilon)
	\e\bigg(\frac{\lambda^2\left(R_{\max}+(1+\beta)H \Phi_{\max}^2 \right)^2}{2}
	\sum\limits_{k=1}^{n} L^2_k \bigg).
	\label{eq:thm1-concbound}
	\end{align}
	The proof of Proposition \ref{prop:main} part (1) follows by optimizing over $\lambda$ in \eqref{eq:thm1-concbound}.
	\end{proof}

\subsubsection{\textbf{Proof of Proposition \ref{prop:main} part (2)}}
\label{sec:expectation-prop-proof}
\begin{proof}
	The proof of this result also follows a martingale analysis. In contrast to the high probability bound, here we work directly with the error, rather than the centered error, and split it into predictable and martingale parts. Bounding the predictable part then bounds the influence of the initial error, and bounding the martingale part bounds the error due to sampling.\ \\[1ex]
	\noindent{\bf Step 1: (Extract a martingale difference from the update)}\\[1ex]
	First, by using that $\bar{A}_T = \E((\phi(s_{i_n}) - \beta \phi(s_{i_n}'))\phi(s_{i_n})\tr\mid \F_{n-1})$ and that $\E(f_n(\hat\theta_T)\mid \F_{n-1}) = 0$, we can rearrange the update rule \eqref{eq:random-lstd-update} to get
	\begin{align*}
	\theta_{n-1} - \hat\theta_T - \gamma_n f_n(\theta_{n-1})
	&= \theta_{n-1} - \hat\theta_T - \gamma_n (\E( f_n(\theta_{n-1}) + \Delta M_n )\\
	&= \left( I - \gamma_n \bar{A}_T\right)z_{n-1}
	- \gamma_n \Delta M_n,
	\end{align*}
	where $\Delta M_n : = f_n(\theta_{n-1}) - \E( f_n(\theta_{n-1}) \mid\F_{n-1})$ is a martingale difference.\\[1ex]
	\noindent{\bf Step 2: (Apply Jensen's inequality to the square of the norm)}\\[1ex]
	From Jensen's inequality, and the fact that the projection in the update rule \eqref{eq:random-lstd-update} is non-expansive, we obtain
	\begin{align}
	\E&\left(\l z_n\r\mid\F_{n-1}\right)^2
	\le \E (\langle z_n, z_n \rangle \mid\F_{n-1})\nonumber\\
	&\le \E (\langle \theta_{n-1} - \hat\theta_T - \gamma_n f_n(\theta_{n-1}),
	\theta_{n-1} - \hat\theta_T - \gamma_n f_n(\theta_{n-1}) \rangle \mid\F_{n-1} )\nonumber\\
	&=  \E (\langle \left( I - \gamma_n \bar{A}_T\right)z_{n-1} - \gamma_n \Delta M_n,
	\left( I - \gamma_n \bar{A}_T\right)z_{n-1} - \gamma_n \Delta M_n \rangle \mid\F_{n-1} )\nonumber\\
	&=  z_{n-1}\tr \left( I - \gamma_n \bar{A}_T\right)\tr\left( I - \gamma_n \bar{A}_T\right) z_{n-1}
	+  \gamma_n^2 \E \left(\langle \Delta M_n, \Delta M_n \rangle\mid\F_{n-1}\right)\label{eq:thm2-it-exp}\\
	&\le  \l z_{n-1}\r^2 \l \left( I - \gamma_n \bar{A}_T\right)\tr\left( I - \gamma_n \bar{A}_T\right)\r
	+  \gamma_n^2 \E \left(\l \Delta M_n\r^2 \mid\F_{n-1}\right).\nonumber
	\end{align}
	Note that the cross-terms have vanished in \eqref{eq:thm2-it-exp} since $\Delta M_n$ is martingale difference, independent of the other terms, given $\F_{n-1}$.\\[1ex]
	\noindent{\bf Step 3: (Unroll the iteration)}\\[1ex] 
	Using assumptions (A1) and (A2)
	\begin{align}
	\l (I - \gamma_n \bar{A}_T)\tr(I - \gamma_n \bar{A}_T)\r 
	&= \l (I - \gamma_n ( (\bar{A}_T\tr + \bar{A}_T) - \gamma_n \bar{A}_T\tr\bar{A}_T)\r \\
	&\le 1 - \gamma_n(2\mu - \gamma_n(1+\beta)^2\Phi_{\max}^4)
	\label{eq:thm2-it-exp-1}
	\end{align}
	Furthermore, by assumption (A3), and the projection step, the martingale differences $\Delta M_n$ are bounded in norm by $2(R_{\max}+(1+\beta)H\Phi_{\max}^2)$. 
	By applying the tower property of conditional expectations repeatedly together with \eqref{eq:thm2-it-exp-1} we arrive at the following bound:
		\begin{align*}
		&\E\left(\l z_n\r\right)^2
		\le \left[\prod_{k = 1}^n
		\left(1 - \gamma_k(2\mu - \gamma_k(1+\beta)^2\Phi_{\max}^4)\right)
		\l z_0\r\right]^2\\
		&+ 4 \sum_{k=1}^{n}\gamma_k^2
		\left[\prod_{j = k}^{n-1}
		(1 - \gamma_j(2\mu  - \gamma_j(1+\beta)^2\Phi_{\max}^4) \right]^2
		\left(R_{\max}+(1+\beta) H \Phi_{\max}^2\right)^2
		\end{align*}
	\end{proof}

\subsubsection{\textbf{Derivation of rates given in Theorem \ref{thm:flstd-rate}}}
\label{sec:flstd-rate-derive}
\begin{proof}\ \\[1ex]
	To obtain the rates specified in the bound in expectation in Theorem \ref{thm:flstd-rate}, we simplify the bound in expectation in Proposition \ref{prop:main} using the choice $\gamma_n = \frac{c_0 c}{(c+n)}$, with $c_0\in(0,\mu((1 + \beta)^2\Phi_{\max}^4)^{-1}]$ and $2c_0 c\mu \in (1,\infty)$. 
	Consider the sampling error term in \eqref{eq:lstd-expectation-bound} under the aforementioned choice for the step size.
	\begin{align}
	&\sum_{k=1}^{n}\gamma_k^2
	\left[\prod_{j = k+1}^{n}
	(1 - \gamma_j(2\mu  - \gamma_j(1+\beta)^2\Phi_{\max}^4) \right]^2\nonumber\\
	&=\sum_{k=1}^{n}\gamma_k^2 \exp \left(2 \sum_{j = k+1}^{n}\ln \left(1 - \gamma_j(2\mu  - \gamma_j(1+\beta)^2\Phi_{\max}^4)\right)\right)\label{eq:exploginrateproof} \\
	&=\sum_{k=1}^{n}\frac{c_0^2 c^2}{(c+k)^2} \exp \left(2 \sum_{j = k+1}^{n}\ln \left(1 - \frac{c_0 c}{c+j}\left(2\mu  - \frac{c_0 c}{c+j}(1+\beta)^2\Phi_{\max}^4\right)\right)\right) \nonumber\\
	&\le\sum_{k=1}^{n}\frac{c_0^2 c^2}{(c+k)^2} \exp \left(2 \sum_{j = k+1}^{n}\ln \left(1 - \frac{c_0 c\mu}{c+j}\right) \right)\label{eq:coplugin}\\
	&\le\sum_{k=1}^{n} \frac{c_0^2c^2}{(c+k)^2}
	\e\left(-2c_0 c \mu\left(\sum_{j=k+1}^{n} \frac{1}{c+j} \right)\right) \label{eq:lnulessthanu}\\
	&\le c_0^2c^2 (c+n+1)^{-2c_0c\mu} \sum_{k=1}^{n} (c+k+1)^{2c_0c\mu} (c+k)^{-2}\label{eq:sumprodintegralbound}\\
	&\le \frac{c_0^2c^2 e^{2} }{(2c_0c\mu - 1)(n+c+1)}\label{eq:compar}
	\end{align}
	In the above, the inequality in \eqref{eq:exploginrateproof} uses the fact that $1 - \gamma_j\left(2\mu  - \gamma_j(1+\beta)^2\Phi_{\max}^4\right) > 0$, a claim that was established earlier in \eqref{eq:liinc}.
	The inequality in \eqref{eq:coplugin} uses $c_0\in(0,\mu((1 + \beta)^2\Phi_{\max}^4)^{-1}]$.
	The inequality in \eqref{eq:lnulessthanu} follows by using $\ln(1+u)\le u$. 
	To infer the inequality in \eqref{eq:sumprodintegralbound}, we  compare to an integral, using that
	$\sum_{j=k+1}^{n} (c+j)^{-1} \ge \int_{x=c+k+1}^{n+c+1} x^{-1} dx$ because the LHS is the upper Riemann sum of RHS.
	Now, evaluating the integral of $x^{-1}$, the exponential term inside the summand of \eqref{eq:lnulessthanu} becomes:
	\begin{align*}
	\exp\left(-2c_0c\mu \sum_{j=k+1}^n (c+j)^{-1}\right) &\le \exp\left(-2c_0c\mu[\ln(c+n+1) - \ln(c+k+1)]\right)\\
	&= (c+n+1)^{-2c_0c\mu} (c+k+1)^{2c_0c\mu}, 
	\end{align*}
	and the inequality in \eqref{eq:sumprodintegralbound} follows by substituting the bound on the RHS above.
	We obtain the final inequality, \eqref{eq:compar}, by upper bounding the term $\sum_{k=1}^n (k + c + 1)^{2 c_0 c \mu} (k+c)^{-2}$ on the RHS of \eqref{eq:sumprodintegralbound} as follows:  
	\begin{align*}
	\sum_{k=1}^n (k + c + 1)^{2 c_0 c \mu} (k+c)^{-2}
	& = \sum_{k=1}^n ( ( (k + c)(1 + 1/(k+c)) ) ^{2 c_0 c \mu} (k+c)^{-2} \\
	& \le \sum_{k=1}^n (1 + 1/c)^{2c} (k + c) ^{2 c_0 c \mu} (k+c)^{-2} \stepcounter{equation}\tag{\theequation}\label{eq:ab0}\\
	&\le e^2 \sum_{k=1}^n (k + c)^{2 (c_0 c \mu - 1)} \stepcounter{equation}\tag{\theequation}\label{eq:ab1}\\
	&\le e^2 \int_{x=0}^{n+1} (x+c)^{2 (c_0 c \mu - 1)} dx \stepcounter{equation}\tag{\theequation}\label{eq:ab2}\\
	& = \frac{e^2(n+c+1)^{-(1 - 2c_0c\mu)}}{(2c_0c\mu - 1)},
	\end{align*}
	where the inequality in \eqref{eq:ab1} holds because
	\[c_0 \mu \le \dfrac{\mu^2}{\Phi_{\max}^4(1+\beta)^2}\le \left(\frac{\mu}{\Phi_{\max}^2}\right)^2 \le 1.\]
	Further, the inequality in \eqref{eq:ab1} follows from the fact that
	$(1 + 1/c)^{2c} \le e^2$ for all $c>0$ and the inequality in \eqref{eq:ab2} follows by comparison of a sum with an integral together with the assumption that $c_0 c \mu > 1$.
	
	Similarly, the initial error term in \eqref{eq:lstd-expectation-bound} can be simplified from the hypothesis that $c_0 c\mu \in (1,\infty)$ and $c_0\in(0,\mu((1 + \beta)^2\Phi_{\max}^4)^{-1}]$ as follows:
	\begin{align}
	\prod_{k = 1}^{n}&
	(1 - \gamma_k(2\mu  - \gamma_k(1+\beta)^2\Phi_{\max}^4)\nonumber\\
	&\le \e\left(-c_0 c \mu\sum_{j = 1}^n\frac{1}{c+j}\right) 
	\le \left(\frac{c+1}{n+c}\right)^{c_0 c \mu }
	\end{align}
	The last inequality above follows again from a comparison with an integral: $\sum_{j = 1}^n\frac{1}{c+j} \ge \int_{c+1}^{c+n} x^{-1}dx = \ln \frac{n+c}{c+1}$. 
	So we have, 
	\begin{align}
	\E \l \theta_n - \hat\theta_T \r
	&\le  \left( \frac{\l \theta_0 - \hat\theta_T \r \sqrt{(c+1)^{c_0 c\mu}}}{\sqrt{(n+c)^{c_0 c\mu-1}}}
	+ \frac{ 2ec_0 c (R_{\max} + (1+\beta)H\Phi_{\max}^2)}{\sqrt{2c_0 c\mu - 1}}\right)\nonumber\\
	&\qquad\qquad\qquad\qquad\times \sqrt{\frac{1}{n+c}},\label{eq:expectation-bound-1byn}
	\end{align}
	and the result concerning the bound in expectation in Theorem \ref{thm:flstd-rate} now follows.
	
	We now derive the rates for the high-probability bound in Theorem \ref{thm:flstd-rate}. 
	With $\gamma_n = \frac{c_0 c}{(c+n)}$, and $c_0\in(0,\mu((1 + \beta)^2\Phi_{\max}^4)^{-1}]$, we have
	\begin{align}
	&\sum_{i=1}^{n} L_i^2
	= \sum_{i=1}^{n}\frac{c_0^2c^2}{(c+i)^2}
	\prod_{j=i+1}^{n}\left( 1 -  \frac{c_0 c}{(c+j)}\left(2\mu - (1 + \beta)^2\Phi_{\max}^4\frac{c_0 c}{(c+j)}\right) \right)\nonumber\\
	&\ \le \sum_{i=1+1}^{n}\frac{c_0^2c^2}{(c+i)^2}
	\prod_{j=i+1}^{n}\left( 1 -  \frac{c_0 c\mu}{(c+j)} \right)\label{eq:rate-ineq-0}\\
	&\ \le \sum_{i=1}^{n}\frac{c_0^2 c^2}{(c+i)^2}
	\e\left(-c_0 c\mu \sum\limits_{j=i+1}^{n}\frac{1}{(c+j)}\right)\label{eq:rate-ineq-1}\\
	&\ \le \frac{c_0^2c^2} {(n+c)^{c_0 c \mu }}
	\sum_{i=1}^{n} (i+c+1)^{c_0 c \mu}(i+c)^{-2}.\label{eq:rate-ineq-xy}\\
	&\ \le \frac{c_0^2c^2 e} {(n+c)^{c_0 c \mu }}
	\sum_{i=1}^{n} (i+c)^{-(2-c_0 c \mu)}.\label{eq:rate-ineq-2}
	\end{align}
	Inequality \eqref{eq:rate-ineq-0} follows from the assumption on $c_0$. To obtain the inequality \eqref{eq:rate-ineq-1}, as in the rates for the bound in expectation, we have taken the exponential of the logarithm of the product, brought the product outside the logarithm as a sum, and applied the inequality $\ln(1 - x) \le x $ which holds for $x\in [0,1)$. The inequality in \eqref{eq:rate-ineq-xy} can be inferred in a manner analogous to that in \eqref{eq:sumprodintegralbound}, while that in \eqref{eq:rate-ineq-2} follows in a similar manner as \eqref{eq:ab1}.
	
	We now find three regimes for the rate of convergence, based on the choice of $c$. Each case is again derived from a comparison of the sum in \eqref{eq:rate-ineq-2} with an appropriate integral:\\
	\begin{inparaenum}[\bfseries(i)]
		\item $\sum_{i=1}^{n} L_i^2 = O\left((n+c)^{c_0 c\mu}\right)$ when $c_0 c \mu \in(0,1)$, \\
		\item $\sum_{i=1}^{n} L_i^2 = O\left(n^{-1}\ln n\right)$ when $c_0 c \mu =1$, and\\
		\item $\sum_{i=1}^{n} L_i^2 \le \frac{c_0^2 c^2 e } {(c_0 c \mu - 1)}(n+c)^{-1}$ when $c_0 c \mu \in(1,\infty)$.\\
	\end{inparaenum}
	Thus, setting $c \in (1/(c_0\mu),\infty)$, the high probability bound from Proposition \ref{prop:main} gives 
	\begin{align}
	\label{eq:prob-bound-1byn}
	\P\left(   \l \theta_n - \hat\theta_T \r - \E \l \theta_n - \hat\theta_T\r \ge \epsilon \right)
	\le \e\left(- \dfrac{\epsilon^2 (n+c)}{4 K_{\mu, c, c_0, \beta}}\right),
	\end{align}
	where $K_{\mu,c,c_0,\beta}\triangleq\dfrac{c_0^2c^2 e\left(R_{\max} + (1+\beta)H\Phi_{\max}^2\right)^2} {(c_0 c \mu - 1)}$.
	The high probability bound in Theorem \ref{thm:flstd-rate} now follows.
	\end{proof}

\subsection{Proof of expectation bound for batchTD without projection}
\label{sec:expectation-bound-noproj-proof}
The proof of the theorem follows just as the proof of Theorem \ref{thm:flstd-rate} but using the following proposition in place of Proposition \ref{prop:main} part 2. The proof of the following proposition differs from that of Proposition \ref{prop:main} part 2 in that the decomposition of the computational error extracts a noise term dependent only on $\hat\theta_T$ rather than on $\theta_n$, and so projection is not needed.
\begin{proposition}
	\label{prop:no-proj}
	Let $z_n = \theta_n - \hat \theta_T$, where $\theta_n$ is given by \eqref{eq:random-lstd-update} with $\Upsilon(\theta)=\theta, \ \forall \theta \in \R^d$.
	Under (A1)-(A4),  we have $\forall \epsilon > 0$,
	\begin{align}
		\label{eq:lstd-expectation-bound-no-proj}
		&\E\left(\l z_n\r\right)^2
		\le \underbrace{3\left[\prod_{k = 1}^n
			\left(1 - \gamma_k(2\mu - \gamma_k(1+\beta)^2\Phi_{\max}^4)\right)
			\l z_0\r\right]^2}_{\textbf{initial error}}\\
		&+  \underbrace{3\sum_{k=1}^{n}\gamma_k^2
			\left[\prod_{j = k}^{n-1}
			(1 - \gamma_j(2\mu  - \gamma_j(1+\beta)^2\Phi_{\max}^4) \right]^2
			\left(R_{\max}+(1+\beta) \l \hat\theta_T\r \Phi_{\max}^2\right)^2}_{\textbf{sampling error}}\nonumber
		\end{align}
\end{proposition}
\begin{proof}\ \\
	\noindent{\bf Step 1: (Unrolling the error recursion)}\\[1ex]
	First, by rearranging the update rule \eqref{eq:random-lstd-update} we obtain an iteration for the computational error $z_n = \theta_n - \hat\theta_T$, and subsequently unroll this iteration:
	\begin{align*}
		z_n
		&= \theta_n - \hat\theta_T = \theta_{n-1} - \hat\theta_T - \gamma_n f_n(\theta_{n-1})\\
		&= \left( I - \gamma_n(\phi(s_{i_n}) - \beta \phi(s_{i_n}'))\phi(s_{i_n})\tr\right)z_{n-1}
		- \gamma_n f_n(\hat\theta_T)\\
		&= \tpi_1^n z_0 - \sum_{k = 1}^n \gamma_k \tpi_{k+1}^{n} f_k(\hat\theta_T).
		\end{align*}
		where $\tpi_k^n \triangleq \prod_{j = k}^n \left( I - \gamma_j(\phi(s_{i_j}) - \beta \phi(s_{i_j}'))\phi(s_{i_j})\tr\right)$ for $1\le k \le n$, and $\tpi_k^n=I$  for $k>n$\footnote{One usually sees terms of the form $\phi(s_{i_j}) (\phi(s_{i_j}) - \beta \phi(s_{i_j}'))$, whereas we use a transposed form to simplify handling the products that get written through the $\tpi_j^n$ matrices.}. In the above, we have used that the random increment at time $n$ has the form $f_n(\theta) = (\theta\tr \phi(s_{i_n}) - (r_{i_n} + \beta \theta\tr \phi(s'_{i_{n}})  ))\phi(s_{i_n})$. Notice that by the definition of the LSTD solution, we have that $\E(f_n(\hat\theta_T)\mid \F_{n-1}) = 0$, and so $f_n(\hat\theta_T)$ is a zero mean random variable.
	\\[3ex]
	\noindent{\bf Step 2: (Taking the expectation of the norm)}\\[1ex] From Jensen's inequality, we obtain
	\begin{align}
		\E\left(\l z_n\r\right)^2
		&\le  3 z_0\tr \E \left({\tpi_1^n}\tr\tpi_1^n\right) z_0
		+  3\sum_{k=1}^{n}\gamma_k^2 \E \left(f_k(\hat\theta_T) \tr {\tpi_{k+1}^{n}}\tr \tpi_{k+1}^{n}f_k(\hat\theta_T) \right),\label{eq:thm2-it-exp-noproj}
		\end{align}
		where we have used the identity $\l x - y \r^2 \le 3 \l x \r^2 + 3 \l y \r^2$
		for any two vectors $x,y$. 
	
	Using assumptions (A1) and (A2), we have
	\begin{align}
	&\l \E ( \left( I - \gamma_n(\phi(s_{i_n}) - \beta \phi(s_{i_n}'))\phi(s_{i_n})\tr\right)\tr
	\left( I - \gamma_n(\phi(s_{i_n}) - \beta \phi(s_{i_n}'))\phi(s_{i_n}))\tr \right)\r \nonumber\\
	&\quad = \l \E \left( I - \gamma_n( (\phi(s_{i_n}) - \beta \phi(s_{i_n}'))\phi(s_{i_n}) \tr
	- \gamma_n\phi(s_{i_n}) (\phi(s_{i_n}) - \beta \phi(s_{i_n}'))\tr\right.\right.\nonumber\\
	&\qquad\left.\left.
	+ \gamma_n^2\left( \l\phi(s_{i_n})\r^2  - 2\beta\langle \phi(s_{i_n}'), \phi(s_{i_n})\rangle
	+ \beta^2\l\phi(s_{i_n}')\r^2\right)
	\phi(s_{i_n}))\phi(s_{i_n}))\tr
	\right)\r\nonumber\\
	&\quad \le 1 - \gamma_n(2\mu - \gamma_n(1+\beta)^2\Phi_{\max}^4)
	\label{eq:thm2-it-exp-1-noproj}
	\end{align}
	Furthermore, by assumption (A3), the random variables $f_n(\hat\theta_T)$ are bounded in norm by $R_{\max}+(1+\beta)\l \hat\theta_T\r\Phi_{\max}^2$. So, by applying the tower property of conditional expectations repeatedly together with \eqref{eq:thm2-it-exp-1-noproj} we arrive at the bound:
	\begin{align*}
		\E&\left(\l z_n\r\right)
		\le \left(3\left[\prod_{k = 1}^n
		(1 - \gamma_k(2\mu - \gamma_k(1+\beta)^2\Phi_{\max}^4) \l z_0\r
		\right]^2
		\right.\\
		&\left.+  3\sum_{k=1}^{n}\gamma_k^2
		\left[\prod_{j = k}^{n-1} (1 - \gamma_j(2\mu - \gamma_j(1+\beta)^2\Phi_{\max}^4)
		\right]^2
		\left(R_{\max}+(1+\beta) \l \hat\theta_T\r \Phi_{\max}^2\right)^2\right)^\frac{1}{2}.
		\end{align*}
	\end{proof}

\paragraph{\textbf{Proof of Theorem \ref{thm:flstd-rate-noproj}}}
\begin{proof}
	We need to prove that  $\E \l \theta_n - \hat \theta_T \r \le \dfrac{K_1(n)}{\sqrt{n+c}},$ where $\theta_n$ is the batchTD iterate that is not projected and $K_1(n)$ is as defined in Theorem \ref{thm:flstd-rate-noproj}. 
	Once we have Proposition \ref{prop:no-proj} in place, the bound mentioned before follows using a completely parallel argument to that used in Section \ref{sec:flstd-rate-derive} to prove the bound in expectation in Theorem \ref{thm:flstd-rate} for projected batchTD. 
	\end{proof}
\subsection{Proofs of finite time bounds for iterate averaged batchTD}
\label{sec:proof-lstd-avg}

For establishing the bounds in expectation and high probability, we follow the technique from \cite{fathi2013transport}, where the authors provide concentration bounds for general stochastic approximation schemes. However, unlike them, we make all the constants explicit and more importantly, we provide an explicit iteration index $n_0$ after which the distance between averaged iterate $\bar \theta_n$ and LSTD solution $\hat\theta_T$ is nearly of the order $O(1/n)$. For providing such a $n_0$, we have to deviate from \cite{fathi2013transport} in several steps of the proof.

\vspace{1ex}

\noindent{\bf\em Proof of the bound in expectation in Theorem \ref{thm:flstd-avg-rate}:}
\begin{proof}
	We bound the expected error by directly averaging the errors of the non-averaged iterates, i.e.,
	\begin{align}
	\E\l \bar\theta_{n} - \hat\theta_T\r \le \frac{1}{n+1}\sum_{k = 0}^n\E\l\theta_k - \hat\theta_T \r.
	\label{eq:avg-ia}
	\end{align}
	For simplifying the RHS above, we apply the bounds in expectation given in Proposition \ref{prop:main}. Recall that the rates in Theorem \ref{thm:flstd-rate} are for step sizes of the form $\gamma_n = \frac{c_0 c}{c+n}$, while iterate averaged batchTD uses a different step size sequence. 
	In the following, we specialize the bound in expectation in Proposition \ref{prop:main} for the new choice of step-size sequence and subsequently, average the resulting bound using \eqref{eq:avg-ia} to obtain the final rate in expectation in Theorem \ref{thm:flstd-avg-rate}. 
	Let $\gamma_n \triangleq c_0\left(c/(c+n)\right)^{\alpha}$.
		We assume $n>n_0$, i.e., 
		\begin{align}
		\frac{c_0 c^{\alpha}}{(c+n)^{\alpha}} (1+\beta)^2\Phi_{\max}^4 < \mu.\label{eq:ap_ia_lem2_assump}
		\end{align}
Using Proposition \ref{prop:main} followed by a split of the individual terms into those before and after $n_0$, we have
		\begin{align}
		&\E\left(\l \theta_n - \hat\theta_T\r\right)^2
		\le \left[\prod_{k = 1}^n
		\left(1 - \gamma_k(2\mu - \gamma_k(1+\beta)^2\Phi_{\max}^4)\right)
		\l z_0\r\right]^2\nonumber\\
		&\quad+  4\sum_{k=1}^{n}\gamma_k^2
		\left[\prod_{j = k}^{n-1}
		(1 - \gamma_j(2\mu  - \gamma_j(1+\beta)^2\Phi_{\max}^4) \right]^2
		\left(R_{\max}+(1+\beta) H \Phi_{\max}^2\right)^2\nonumber\\
		&
		= \left[ \prod_{k = 1}^{n_0}
		\left(1 - \gamma_k(2\mu - \gamma_k(1+\beta)^2\Phi_{\max}^4)\right) \right. \nonumber\\
		& \qquad \left. \times \prod_{k = n_0 + 1}^n
		\left(1 - \gamma_k(2\mu - \gamma_k(1+\beta)^2\Phi_{\max}^4)\right)
		\l z_0\r\right]^2\nonumber\\
		&\quad+  4\sum_{k=1}^{n_0}\gamma_k^2
		\left[\prod_{j = k}^{n-1}
		(1 - \gamma_j(2\mu  - \gamma_j(1+\beta)^2\Phi_{\max}^4) \right]^2
		\left(R_{\max}+(1+\beta) H \Phi_{\max}^2\right)^2\nonumber\\
		&+  4\sum_{k=n_0+1}^{n}\gamma_k^2
		\left[\prod_{j = k}^{n-1}
		(1 - \gamma_j(2\mu  - \gamma_j(1+\beta)^2\Phi_{\max}^4) \right]^2 \!\!
		\left(R_{\max}+(1+\beta) H \Phi_{\max}^2\right)^2\nonumber\\
		&
		\le \left[ \prod_{k = 1}^{n_0}
		\left(1 + (1+\beta)\Phi_{\max}^2c_0\right)^2 \prod_{k = n_0 + 1}^n
		\left(1 - \gamma_k(2\mu - \gamma_k(1+\beta)^2\Phi_{\max}^4)\right)
		\l z_0\r\right]^2\nonumber\\
		&\quad+  4\sum_{k=1}^{n_0}c_0^2
		\left[\prod_{j = k}^{n_0}
		\left(1 + (1+\beta)\Phi_{\max}^2c_0\right)^2 \right]^2\left[\prod_{j = n_0+1}^{n-1}
		(1 - \gamma_j(2\mu  - \gamma_j(1+\beta)^2\Phi_{\max}^4) \right]^2\nonumber\\
		&\qquad \quad 
		\times\left(R_{\max}+(1+\beta) H \Phi_{\max}^2\right)^2\nonumber\\
		&\quad+  4\sum_{k=n_0+1}^{n}\gamma_k^2
		\left[\prod_{j = k}^{n-1}
		(1 - \gamma_j(2\mu  - \gamma_j(1+\beta)^2\Phi_{\max}^4) \right]^2
		\left(R_{\max}+(1+\beta) H \Phi_{\max}^2\right)^2\label{eq:aza1}\\
		&\le \left[ \left(1 + c_0(1+\beta)\Phi_{\max}^2\right)^{2n_0}
		\prod_{k = n_0+1}^n \left(1 - \frac{\mu c_0 c^{\alpha}}{(c+k)^{\alpha}}\right)
		\l z_0\r\right]^2\nonumber\\
		&\quad+  4n_0 c_0^2 \left(1 + c_0(1+\beta)\Phi_{\max}^2\right)^{4n_0} \left[\prod_{j = n_0+1}^{n-1} \left(1 - \frac{\mu c_0 c^{\alpha}}{(c+j)^{\alpha}}\right)\right]
		\left(R_{\max}+(1+\beta) H \Phi_{\max}^2\right)^2\nonumber\\
		&\quad+  4\sum_{k=n_0+1}^{n}\frac{c_0^2 c^{2\alpha}}{(c+k)^{2\alpha}}
		\left[\prod_{j = k}^{n-1}
		\left(1 - \frac{\mu c_0 c^{\alpha}}{(c+j)^{\alpha}}\right) \right]^2
		\left(R_{\max}+(1+\beta) H \Phi_{\max}^2\right)^2\label{eq:ap_ia_lem2_3}\\
		&\le \left[ \e\left(2c_0(1+\beta)\Phi_{\max}^2n_0\right)
		\e\left(-\mu c_0 \sum_{k = n_0+1}^n \frac{c^{\alpha}}{(c+k)^{\alpha}}\right)
		\l z_0\r\right]^2\nonumber\\
		&\quad+  4n_0 c_0^2 \e\left( 4c_0(1+\beta)\Phi_{\max}^2n_0\right)
		\e\left(-2\mu c_0\sum_{j = n_0+1}^{n-1}
		\frac{ c^{\alpha}}{(c+j)^{\alpha}}\right) \nonumber\\
		& \qquad\qquad \times 
		\left(R_{\max}+(1+\beta) H \Phi_{\max}^2\right)^2\nonumber\\
		&\quad+  4\sum_{k=n_0+1}^{n}\frac{c_0^2 c^{2\alpha}}{(c+k)^{2\alpha}}
		\e\left(-2\mu c_0\sum_{j = k}^{n-1}
		\frac{ c^{\alpha}}{(c+j)^{\alpha}}\right)
		\left(R_{\max}+(1+\beta) H \Phi_{\max}^2\right)^2\nonumber.
		\end{align}
	In the above, the inequality in \eqref{eq:aza1} can be inferred from the following:
	\begin{align*}
	\left(1 - \gamma_k(2\mu - \gamma_k(1+\beta)^2\Phi_{\max}^4)\right) 
	& \le \left(1 + 2(1+\beta)\Phi_{\max}^2\gamma_{k} + (1+\beta)^2\Phi_{\max}^4\gamma_{k}^2\right)\\
	&\le \left(1 + (1+\beta)\Phi_{\max}^2c_0\right)^2,\stepcounter{equation}\tag{\theequation}\label{eq:squarestuff}
	\end{align*}
	where we have used the fact that $\mu>0$ and $\gamma_k < c_0$. 
	To obtain the inequality in \eqref{eq:ap_ia_lem2_3}, we have split the product at $n_0$, and, when $k\le n_0$, we have used $(1+x)^{n_0} = e^{n_0\ln(1+x)}\le e^{x n_0}$ and when $k>n_0$, we have applied \eqref{eq:ap_ia_lem2_assump}.
	For the final inequality above, we have exponentiated the logarithm of the products, and used the inequality $\ln(1+x) < x$ in several places.
	
	With $C_1$ and $C_2$ as defined in the statement of Theorem \ref{thm:flstd-avg-rate}, we have that
		\begin{align}
		\E&\l \theta_n - \hat\theta_T \r
		\le C_1\e\left(-c_0\mu c^\alpha\left((n+c)^{1-\alpha} - (n_0+c+1)^{1-\alpha}\right)\right)\l \theta_0 - \hat\theta_T\r \nonumber\\
		& +\left(R_{\max} + (1+\beta)H\Phi_{\max}^2\right)
		.\Bigg(4 n_0c_0^2 C_1^2 \e\left(-2c_0\mu c^\alpha((n+c)^{1-\alpha} - (n_0+c+1)^{1-\alpha}\right)\nonumber\\
		&\quad +\sum_{k = n_0 + 1}^{n}c_0^2\left(\frac{c}{k+c}\right)^{2\alpha}
		\e\left(-2c_0\mu c^\alpha((n+c)^{1-\alpha} - (k+c)^{1-\alpha}\right)
		\Bigg)^{\frac{1}{2}}\label{eq:lem2-eq-0}\\
		=  & \e\left(-c_0\mu c^\alpha(n+c)^{1-\alpha}\right)\nonumber\\
		& \qquad \times \Bigg[C_1 C_2\l \theta_0 - \hat\theta_T\r + \left(R_{\max} + (1+\beta)H\Phi_{\max}^2\right) \nonumber\\
		& \qquad\qquad
		\times\left\{4n_0c_0^2 C_1^2 C_2^2 + \sum_{k = n_0 + 1}^{n}c_0^2\left(\frac{c}{k+c}\right)^{2\alpha}
		\e\left(2c_0\mu c^\alpha((k+c)^{1-\alpha}\right)
		\right\}^{\frac{1}{2}}
		\Bigg]\nonumber\\	
		\le& \e\left(-c_0\mu c^\alpha(n+c)^{1-\alpha}\right)\nonumber\\
		& \qquad \times \Bigg[C_1 C_2\l \theta_0 - \hat\theta_T\r
		+ \left(R_{\max} + (1+\beta)H\Phi_{\max}^2\right) \nonumber\\
		& \qquad\qquad
		\times \left\{4n_0c_0^2 C_1^2 C_2^2
		+ c^{2\alpha} c_0^2\int_{1}^{n+c}x^{-2\alpha}\e \left(2c_0\mu c^\alpha x^{1-\alpha}\right)dx
		\right\}^{\frac{1}{2}}
		\Bigg]\label{eq:lem2-eq-2}\\	
		\le&  \e\left(-c_0\mu c^\alpha (n+c)^{1-\alpha}\right)\nonumber\\
		&\qquad\qquad
		\times\Bigg[C_1 C_2 \l \theta_0 - \hat\theta_T\r
		+ \left(R_{\max} + (1+\beta)H\Phi_{\max}^2\right) \nonumber\\
		&\qquad\qquad\qquad\qquad
		\times\Bigg\{ 4n_0c_0^2 C_1^2 C_2^2
		+ c^{2\alpha} c_0^2\left(2c_0\mu c^\alpha \right)^{\frac{2\alpha}{1-\alpha}}\nonumber\\
		&\qquad\qquad\qquad\qquad\qquad\qquad\times \int_{\left(2c_0\mu c^\alpha\right)^{1/(1-\alpha)}}^{(n+c)\left(2c_0\mu c^\alpha\right)^{1/(1-\alpha)}}
		y^{-2\alpha}\e (y^{1-\alpha})dy
		\Bigg\}^{\frac{1}{2}}.
		\Bigg]\label{eq:lem2-eq-3}
		\end{align}
	In the above, the inequality in \eqref{eq:lem2-eq-0} follows by an application of Jensen's Inequality together with the fact that $\sum_{j=k}^{n-1}(c+j)^{-\alpha}\ge \int_{j=k}^n(c+j)^{-\alpha}dj=(c+n)^{1-\alpha} - (c+k)^{1-\alpha}$. To obtain the inequality in \eqref{eq:lem2-eq-2}, we have upper bounded the sum with an integral, the validity of which follows from the observation that $x\mapsto x^{-2\alpha}e^{x^{1-\alpha}}$ is convex for $x\ge 1$. Finally, for arriving at the inequality in \eqref{eq:lem2-eq-3}, we have applied the change of variables $y = (2c_0\mu c^\alpha)^{1/(1-\alpha)}x$.
	
	Now, since $y^{-2\alpha} \le \frac{2}{1-\alpha} ((1-\alpha)y^{-2\alpha} - \alpha y^{-(1+\alpha)})$ when $y\ge\left( \frac{2\alpha}{1-\alpha}\right)^{\frac{1}{1-\alpha}}$, we have
	\begin{align*}
	&\int_{\left( \frac{2\alpha}{1-\alpha}\right)^{\frac{1}{1-\alpha}}}
	^{(n+c)\left(2c_0\mu c^\alpha\right)^{1/(1-\alpha)}}
	y^{-2\alpha}\e (y^{1-\alpha})dy\\
	&\quad
	\le \frac{2}{1-\alpha} \int_{\left( \frac{2\alpha}{1-\alpha}\right)^{\frac{1}{1-\alpha}}}
	^{(n+c)\left(2c_0\mu c^\alpha\right)^{1/(1-\alpha)}}
	((1-\alpha)y^{-2\alpha} - \alpha y^{-(1+\alpha)})
	\e (y^{1-\alpha})dy\\
	&\quad
	\le \frac{2}{1-\alpha} \e \left(2c_0\mu c^\alpha (n+c)^{1-\alpha}\right)
	(n+c)^{-\alpha}\left(2c_0\mu c^\alpha\right)^{-\alpha/(1-\alpha)}.
	\end{align*}
	and furthermore, since $y\mapsto y^{-2\alpha}\e(y^{1-\alpha})$ is non-decreasing for $y\le\left( \frac{2\alpha}{1-\alpha}\right)^{\frac{1}{1-\alpha}}$, we have
	\begin{align*}
	\int_{1}^{\left( \frac{2\alpha}{1-\alpha}\right)^{\frac{1}{1-\alpha}}}
	y^{-2\alpha}\e (y^{1-\alpha})dy
	\le e \left( \frac{2\alpha}{1-\alpha}\right)^{\frac{1}{1-\alpha}}.
	\end{align*}
	
		Plugging these into \eqref{eq:lem2-eq-3}, we obtain
		\begin{align*}
		\E&\l \theta_n - \hat\theta_T \r\\
		\le& \e\left(-c_0\mu c^\alpha(n+c)^{1-\alpha}\right)\\
		&\quad
		.\left(C_1 C_2\l \theta_0 - \hat\theta_T\r
		+ \sqrt{e}\left(\frac{2\alpha}{1-\alpha}\right)^{\frac{1}{2(1-\alpha)}}c^\alpha c_0
		\left( 2c_0\mu c^\alpha\right)^{\frac{\alpha}{(1-\alpha)}}\right.\\
		& \qquad\quad\left.
		+  2c_0 C_1 C_2\left(R_{\max} + (1+\beta)H\Phi_{\max}^2\right) \sqrt{n_0}
		\right)\\
		&\qquad
		+  \sqrt{\frac{2}{1-\alpha}}\left(R_{\max} + (1+\beta)H\Phi_{\max}^2\right) c^\alpha c_0
		\left( 2c_0\mu c^\alpha\right)^{\frac{\alpha}{2(1-\alpha)}}.
		(n+c)^{-\frac{\alpha}{2}}\stepcounter{equation}\tag{\theequation}\label{eq:t122}
		\end{align*}
		The bound in expectation in the theorem statement can be inferred by using the inequality above in 
		\[\E\l \bar\theta_{n+1} - \hat\theta_T\r \le \frac{1}{n+1}\sum_{k = 0}^n\E\l\theta_k - \hat\theta_T \r,\]
		followed by a straightforward bound on the sum of the first exponential term on the RHS of \eqref{eq:t122}, using the constant $C_0$.
	\end{proof}

\newpage

\noindent{\bf\em Proof of the high probability bound in Theorem \ref{thm:flstd-avg-rate}:}\ \\[0.5ex]

The proof of the high probability bound is considerably more involved than the proof of the bound in expectation in Theorem \ref{thm:flstd-avg-rate}.
We first state and prove a bound on the error in high probability for the averaged iterates in Proposition \ref{prop:flstd-avg-hpb} below. This result is for general step-size sequences and can be seen as the iterate average counterpart to Proposition \ref{prop:main}.

\begin{proposition}
	\label{prop:flstd-avg-hpb}
	Let $z_n = \bar\theta_n - \hat\theta_T$. Under (A1)-(A3) we have, for all $\epsilon \ge 0$ and $\forall n\geq 1$,
	\begin{align*}
	&\P(   \l z_n \r - \E \l z_n \r \ge \epsilon )
	\le \e\left(- \dfrac{\epsilon^2}
	{2(R_{\max}+(1+\beta)H\Phi_{\max}^2)^2\sum\limits_{m=1}^{n} L_m^2}
	\right),
	\end{align*}
	where
		$L_i \triangleq \frac{\gamma_i}{n+1}
		\left( \sum_{l=i+1}^{n-1}\prod\limits_{j=i}^{l}
		\left(1- \gamma_{j+1}( 2\mu - (1+\beta)^2\Phi_{\max}^4\gamma_{j+1}))
		\right)^{1/2}\right)$.
\end{proposition}
\begin{proof}
	Recall that $z_n$ denotes the error of the algorithm at time $n$, which in this case is $z_n = \bar{\theta}_n - \theta$.
	The proof follows the scheme of the proof of Proposition \ref{prop:main}, part (1), given in Section \ref{sec:proof-lstd}:
	
	\noindent{\bf Step 1:} As before, we decompose the centered error $\l z_n \r - \E \l z_n \r$ as follows:
	\begin{align}
	\l z_n \r - \E \l z_n \r
	= \sum\limits_{k=1}^{n} D_k, \label{eq:prob-equivalence-averaged}
	\end{align}
	where $D_k \triangleq g_k - \E [ g_k \left| \F_{k-1} \right.]$ and $g_k \triangleq  \E [\l z_n\r \left| \F_k \right.]$.
	
	\noindent{\bf Step 2:}
	We need to prove that the functions $g_k$ are Lipschitz continuous in the random innovation at time $k$ with the new constants $L_k$.
	Recall from Step 2 of the proof of the high probability bound in Theorem \ref{prop:main} in Section \ref{sec:proof-lstd} that the random variable $\Theta_n^k(\theta)$ is defined to be the value of the iterate at time $n$ that evolves according to \eqref{eq:random-lstd-update}, and beginning from $\theta$ at time $k$. Now we define
	$$\bar\Theta_n^k(\bar\theta,\theta) = \frac{k\bar\theta}{n+1} + \frac{1}{n+1}\sum_{j=k}^n \Theta_j^k(\theta).$$
	Then, letting $f$ and $f'$ denote two possible values for the random innovation at time $k$, and setting $\theta = \theta_{k-1} + \gamma_k f$ and $\theta' = \theta_{k-1} + \gamma_k f'$, we have
	\begin{align}
	&\E \l \bar\Theta_{n}^k\left(\bar\theta_{k-1},\theta\right) - \bar\Theta_{n}^k\left(\bar\theta_{k-1},\theta'\right) \r
	= \E\l \frac{1}{n+1} \sum_{l = k}^n\left(\Theta^k_l\left(\theta\right) - \Theta^k_l\left(\theta'\right)\right) \r
	\nonumber \\
	&\le  \dfrac{1}{n+1} \sum\limits_{l=k}^{n}
	\prod\limits_{j=k+1}^{l} \left(1 - \gamma_{j}
	\left(2\mu- \gamma_{j}(1+\beta)^2\Phi_{\max}^4\right)
	\right)^{1/2}
	\l f - f'\r \label{eq:av1}
	\end{align}
	where we have used \eqref{eq:Theta} derived in Step 2 of the proof the high probability bound in Proposition \ref{prop:main}.
	Hence, similarly to Step 2 of the proof of Proposition \ref{prop:main}, part (1), we find that $g_k$ is $L_k$-Lipschitz in the random inovation at time $k$, and so $D_k$ is also.
	
	\noindent{\bf Step 3} follows in a similar manner to the proof of Proposition \ref{prop:main}, part (1).
\end{proof}

We now bound the sum of squares of the Lipschitz constants $L_m$ when the iterates are averaged and the step-sizes are chosen to be $\gamma_n = c_0\left(\frac{c}{c+n}\right)^{\alpha}$ for some $\alpha \in \left(1/2,1\right)$.  
This is a crucial step that helps in establishing the order $O(n^{-\alpha/2})$ rate for the high-probability bound in Theorem \ref{thm:flstd-rate}, independent of the choice of $c$. Recall that in order to obtain this rate for the algorithm without averaging one had to choose $c_0 \mu c \in (1,\infty)$.  

\begin{lemma}
		\label{lemma:avg-hpb}
		Under conditions of Theorem \ref{thm:flstd-avg-rate}, we have 
		\begin{align}\label{eq:avhighprob-Lsum-lemma}
		\sum_{i=1}^n L_i^2
		&\le \,\, \frac{n_0}{(n+1)^2}\left[\frac{e^{(1+\beta)\Phi_{\max}^2c_0 (2n_0+1)}}{(1+\beta)\Phi_{\max}^2}\right]^2\\
		&\quad	+
		\frac{1}{\mu^2}
		\left\{2^\alpha
		+ \left[ \left[\frac{2\alpha}{ c_0\mu c^{\alpha}}\right]^{\frac{1}{1-\alpha}}
		+ \frac{2(1 - \alpha)(c_0\mu)^{\alpha}}{\alpha} \right]
		\right\} ^2\frac{1}{n+1}.
		\end{align}
\end{lemma}
\begin{proof}
	Recall from the statement of Theorem \ref{thm:flstd-avg-rate} that  $n$ satisfies,
		\begin{align}
		\frac{c_0 c^{\alpha}}{(c+n)^{\alpha}} (1+\beta)^2\Phi_{\max}^4 < \mu.
		\label{eq:largen}
		\end{align}
		Recall also from the formula in Proposition \ref{prop:flstd-avg-hpb}, that:
		\begin{align*}
		L_i
		= \frac{\gamma_i}{n+1}
		\left( \sum_{l=i+1}^{n-1}\prod\limits_{j=i}^{l}
		\left(1- \gamma_{j+1}( 2\mu - (1+\beta)^2\Phi_{\max}^4\gamma_{j+1}))
		\right)^{1/2}\right).
		\end{align*}
		We split the bound on the sum into two terms as follows:
		\begin{align}
		\sum_{i = 1}^n L_i^2 = \sum_{i = 1}^{n_0-1} L_i^2 + \sum_{i = n_0}^{n} L_i^2.
		\label{eq:split-hpb}
		\end{align}
		
		The first term in \eqref{eq:split-hpb} is simplified as follows:
		\begin{align}
		\sum_{i = 1}^{n_0 - 1} L_i^2
		&= \sum_{i = 1}^{n_0 - 1}\left[\frac{\gamma_i}{n+1}
		\left( \sum_{l=i+1}^{n_0}\prod\limits_{j=i}^{l}
		\left(1- \gamma_{j+1}( 2\mu - (1+\beta)^2\Phi_{\max}^4\gamma_{j+1}))
		\right)^{1/2}\right)
		\right]^2\nonumber\\
		&\le \frac{1}{(n+1)^2}\sum_{i = 1}^{n_0 - 1}\left[c_0
		\left( \sum_{l=i+1}^{n_0}\prod\limits_{j=i}^{l}
		\left(1 + (1+\beta)\Phi_{\max}^2c_0)\right)
		\right)
		\right]^2\label{eq:n0-l3}\\
		&\le \frac{1}{(n+1)^2}\sum_{i = 1}^{n_0 - 1}
		\left[c_0 (1 + (1+\beta)\Phi_{\max}^2c_0)^{2n_0}
		\sum_{l=1}^{n_0} \left(1 + (1+\beta)\Phi_{\max}^2c_0\right)^{-l}\right]^2\label{eq:n0-l4}\\
		&\le \frac{1}{(n+1)^2}c_0^2n_0 \left[\frac{(1 + (1+\beta)\Phi_{\max}^2c_0)^{2n_0+1}}{(1+\beta)\Phi_{\max}^2c_0}\right]^2\label{eq:n0-l5}\\
		&\le \frac{n_0}{(n+1)^2}\left[\frac{e^{(1+\beta)\Phi_{\max}^2c_0 (2n_0+1)}}{(1+\beta)\Phi_{\max}^2}\right]^2.
		\label{eq:n0-l6}										
		\end{align}
	In the above, the inequality in \eqref{eq:n0-l3} follows from \eqref{eq:squarestuff}.
	, while the inequality in \eqref{eq:n0-l3} applies the form of the step sizes. 
	In obtaining the inequality in \eqref{eq:n0-l4}, we have replaced $i$ with $1$. For the inequality in \eqref{eq:n0-l5}, we have used the formula for the sum of a geometric series, and for the final inequality we have used that $(1+x)^{n_0} = e^{n_0\ln(1+x)}\le e^{x n_0}$.
	
	We now analyze the second term in \eqref{eq:split-hpb}. Notice that 
	\begin{align}
	&\sum_{i = n_0}^n L_i^2
	= \sum_{i = n_0}^n\left[\frac{\gamma_i}{n+1}
	\left( \sum_{l=i+1}^{n-1}\prod\limits_{j=i}^{l}
	\left(1- \gamma_{j+1}( 2\mu - (1+\beta)^2\Phi_{\max}^4\gamma_{j+1}))
	\right)^{1/2}\right)
	\right]^2\nonumber\\
	&\quad\le \frac{1}{(n+1)^2}\sum_{i = n_0}^n\left[ \gamma_i
	\left( \sum_{l = i+1}^{n-1}
	\e\left(- \sum_{j=i}^l \gamma_{j+1}(2\mu - (1+\beta)^2\Phi_{\max}^4\gamma_{j+1}))
	\right)\right)
	\right]^2\label{eq:s12}\\
	& \quad< \frac{1}{(n+1)^2}\sum_{i = n_0}^n
	{\underbrace{
			\left[
			c_0\left(\frac{c}{c+i}\right)^\alpha
			\left( \sum_{l = i+1}^{n-1}
			\e\left(- c_0\mu\sum_{j=i}^l \left(\frac{c}{c+j}\right)^\alpha
			\right)\right)
			\right]}_{\triangleq(A)}}^2.\label{eq:it-av-1}
	\end{align}
	To produce the final bound, we bound the summand (A) highlighted in line \eqref{eq:it-av-1} by a constant, uniformly over all values of $i$ and $n$, as follows:
	\begin{align}
	&\sum_{l = i+1}^{n-1}
	\e\left(- c_0\mu
	\sum_{j=1}^l \left(\frac{c}{c+i}\right)^\alpha
	\right)\nonumber\\
	&= \sum_{l = i+1}^{n-1}
	\left[\left(\frac{c}{c+l}\right)^\alpha
	\e\left(-c_0\mu
	\sum_{j=1}^l \left(\frac{c}{c+i}\right)^\alpha
	\right)\right]
	\left(\frac{c+l}{c}\right)^{\alpha}\nonumber\\
	&\le \sum_{l = i+1}^{n-1}
	\left[\frac{1}{c_0\mu}
	\left(\e\left( -c_0\mu\sum_{j=1}^{l-1} \left(\frac{c}{c+i}\right)^\alpha\right) \right.\right.\nonumber\\
	&\qquad\qquad\qquad\qquad\qquad\qquad \left.\left.-\e\left( -c_0\mu\sum_{j=1}^l \left(\frac{c}{c+i}\right)^\alpha\right)
	\right)
	\right]
	\left(\frac{c+l}{c}\right)^{\alpha}\label{eq:beyondn0-l1}\\
	&= \frac{1}{c_0\mu}
	\Bigg\{
	- \left(\frac{c}{c+n}\right)^{-\alpha}\e\left( -c_0\mu\sum_{j=1}^{n} \left(\frac{c}{c+i}\right)^\alpha\right)\nonumber\\
	&\qquad\qquad
	+ \left(\frac{c}{c+i+1}\right)^{-\alpha}\e\left( -c_0\mu\sum_{j=1}^{i+1} \left(\frac{c}{c+i}\right)^\alpha\right)\nonumber\\
	&\qquad\qquad
	+ \sum_{l = i+1}^{n-1}
	\e\left( -c_0\mu\sum_{j=1}^{l} \left(\frac{c}{c+i}\right)^\alpha\right)
	\left[\left(\frac{c}{c+l+1}\right)^{-\alpha} - \left(\frac{c}{c+l}\right)^{-\alpha}\right]
	\Bigg\},\label{eq:beyondn0-l2}
	\end{align}
	where the inequality in \eqref{eq:beyondn0-l1} follows from the convexity of $e^{-\frac{c_0 \mu}{2}x}$, while that in \eqref{eq:beyondn0-l2} follows by applying an Abel transform.
	
	From the foregoing, the summand term (A) highlighted in \eqref{eq:it-av-1} can be bounded by
	\begin{align*}
	&(A) \le \frac{1}{\mu}\Bigg(\left(\frac{c+i+1}{c+i}\right)^\alpha+\\
	&\frac{1}{(c+i)^\alpha}\sum_{l=i+1}^{n-1}
	\e\left(- c_0\mu c^\alpha\frac{ ((c+l)^{1-\alpha}- (c+i)^{1-\alpha})}{1-\alpha}\right)			
	((c+l+1)^\alpha-(c+l)^\alpha)\Bigg)
	\end{align*}
	Now, using convexity of $x^\alpha$ followed by comparison with an integral, and then a change of variable, we have
	\begin{align}
	&\sum_{l=i+1}^{n-1}
	\e\left(-c_0\mu\frac{c^{\alpha}((c+l)^{1-\alpha} - (c+i)^{1-\alpha})}
	{(1- \alpha)}\right)\left((c+l+1)^\alpha - (c+l)^\alpha \right)\label{eq:it-av-2}\\
	&\le\sum_{l=i+1}^{n-1}
	\e\left(-c_0\mu\frac{c^{\alpha}((c+l)^{1-\alpha} - (c+i)^{1-\alpha})}
	{(1- \alpha)}\right)
	\alpha \left(c+l\right)^{-(1-\alpha)}\nonumber\\
	&\le \alpha
	\e\left(c_0\mu\frac{c^{\alpha} (c+i)^{1-\alpha}}
	{(1-\alpha)}
	\right)
	\Bigg[
	\int_{i}^{n-1}
	\e\left(-c_0\mu\frac{ c^{\alpha} (c+l)^{1-\alpha}}
	{(1-\alpha)}
	\right)
	(c+l)^{-(1-\alpha)}dl
	\Bigg]\nonumber\\
	&= \alpha
	\e\left(c_0\mu\frac{c^{\alpha} (c+i)^{1-\alpha}}
	{(1-\alpha)}
	\right)
	\Bigg[
	\int_{c_0\mu(c+i)^{1-\alpha}}^{c_0\mu(c+n-1)^{1-\alpha}}
	\e\left(-\frac{c^{\alpha} l}
	{(1-\alpha)}
	\right)
	l^{\frac{2\alpha-1}{1-\alpha}}dl
	\Bigg].\label{eq:it-av-3}
	\end{align}
	For the second inequality we have used that the mapping $x\rightarrow e^{-d(c+x)^{1-\alpha}}(c+x)^{-(1-\alpha)}$ is decreasing in $x$ for all $x>1$.
	
	By taking the derivative and setting it to zero, we find that  $l\mapsto \e\left(-\frac{ c^\alpha l}{(1-\alpha)}\right) l^{\frac{2\alpha}{1-\alpha}}$ is decreasing on $[2\alpha/ c^{\alpha},\infty)$, and so we deduce that when $c_0\mu(c+i+1)^{1-\alpha}\ge2\alpha/ c^{\alpha}$,
	\begin{align*}
	\e&\left(\frac{c^{\alpha} (c+i)^{1-\alpha}}
	{(1-\alpha)}
	\right)
	\int_{c_0\mu(c+i+1)^{1-\alpha}}^{c_0\mu(c+n)^{1-\alpha}}
	\e\left(-\frac{c^{\alpha} l}
	{(1-\alpha)}
	\right)
	l^{\frac{2\alpha-1}{1-\alpha}}dl\\
	&\le (c_0\mu)^{\frac{2\alpha}{1-\alpha}}(c+i+1)^{2\alpha}
	\int_{c_0\mu(c+i+1)^{1-\alpha}}^{c_0\mu(c+n)^{1-\alpha}}
	l^{\frac{-1}{1-\alpha}}dl
	< \frac{1-\alpha}{\alpha}((c_0\mu(c+i+1))^{\alpha}.
	\end{align*}
	When $c_0\mu(c+i+1)^{1-\alpha}<2\alpha/ c^{\alpha}$ we can bound the summand of \eqref{eq:it-av-2} by 1, and
	\begin{gather*}
	c_0\mu(c+i+1)^{1-\alpha}<\frac{2\alpha}{ c^{\alpha}}
	\implies
	(c+i+1)^{1-\alpha}< \frac{2\alpha}{ c_0\mu c^{\alpha}}\\
	\implies
	i < \left[\frac{2\alpha}{ c_0\mu c^{\alpha}}\right]^{\frac{1}{1-\alpha}} - c - 1.
	\end{gather*}
	Hence, we conclude that
	\begin{align*}
	\sum_{i=n_0}^n L_i^2
	\le \frac{1}{\mu^2}
	\left\{2^\alpha
	+ \left[ \left[\frac{2\alpha}{ c_0\mu c^{\alpha}}\right]^{\frac{1}{1-\alpha}}
	+ \frac{2(1 - \alpha)(c_0\mu)^{\alpha}}{\alpha} \right]
	\right\} ^2\frac{1}{n+1}.
	\end{align*}
	\end{proof}

\begin{proof}\textbf{\textit{(High probability bound in Theorem \ref{thm:flstd-avg-rate})}}
	Once we have established the bound in expectation for batchTD with iterate averaging and the bound on sum of squares of Lipschitz constants in the lemma above, the proof of the high probability bound is straightforward and follows by arguments similar to that used in establishing the corresponding claim for non-averaged batchTD (see Section \ref{sec:flstd-rate-derive}).\end{proof}

\section{Traffic Control Application}
\label{sec:experiments}
\subsection{Simulation Setup}
The idea behind the experimental setup is to study both LSPI and the variant of LSPI, fLSPI, where we use batchTDQ as a subroutine to approximate the LSTDQ solution. Algorithm \ref{alg:lspiSalstda} provides the pseudo-code for the latter algorithm.

We consider a traffic signal control application for conducting the experiments. The problem here is to adaptively choose the sign configurations for the signalized intersections in the road network considered, in order to maximize the traffic flow in the long run. 
Let $L$ be the total number of lanes in the road network considered. Further, let $q_i(t), i = 1,\ldots,L$ denote the queue lengths and $t_i(t), i = 1,\ldots,L$ the elapsed time (since signal turned to red) on the individual lanes of the road network. Following \cite{prashanth2011reinforcement}, the traffic signal control MDP is formulated as follows: 
\begin{description}
	\item[\textbf{State}] $s_t = \big(q_1(t),\ldots, q_L(t), t_1(t), \ldots, t_L(t)\big)$,
	\item[\textbf{Action}] $a_t$ belongs to the set of feasible sign configurations,
	\item[\textbf{Single-stage cost}] 
	\begin{small}
		$
		h(s_t)  =   u_1\;\Big[ \sum_{i \in I_p} u_2\cdot q_i(t) + \sum_{i \notin I_p} w_2\cdot q_i(t)\Big] 
		+ w_1\;\Big[\sum_{i \in I_p} u_2\cdot t_i(t) + \sum_{i \notin I_p} w_2\cdot t_i(t) \Big],  
		$
	\end{small}
	where $u_i,w_i \ge 0$ such that $u_i + w_i =1$ for $i=1,2$ and $u_2 > w_2$. Here, the set $I_p$ is the set of prioritized lanes.
\end{description}
Function approximation is a standard technique employed to handle high-dimensional state spaces (as is the case with the traffic signal control MDP on large road networks). We employ the feature selection scheme from \cite{prashanth2012threshold}, which is briefly described in the following:
the features $\phi(s,a)$ corresponding to any state-action tuple $(s,a)$ is an $L$-dimensional vector, with one bit for each line in the road network. The feature value $\phi_{i}(s,a), i=1,\ldots,L$ corresponding to lane $i$ is chosen as described in Table \ref{tab:qtlcfanfs}, with $q_i$ and $t_i$ denoting the queue length and elapsed times for lane $i$. Thus, as the size of the network increases, the feature dimension scales in a linear fashion. 

\begin{table}
	\caption{Features for the traffic control application}
	\centering
	\begin{tabular}[b]{|C{5cm}|C{3cm}|C{3cm}|}
		\toprule
		\multirow{2}{*}{\textbf{State}} & \multirow{2}{*}{\textbf{Action}} & \multirow{2}{*}{\textbf{Feature $\phi_{i}(s,a)$}}\\
		&&\\\midrule
		\multirow{2}{*}{$q_i < {\cal L}_1$  and   $t_i < {\cal T}_1$} & RED & 0.01\\
		& GREEN & 0.06\\\hline
		\multirow{2}{*}{$q_i < {\cal L}_1$ and   $t_i \ge {\cal T}_1$} & RED & 0.02\\
		& GREEN & 0.05\\\hline
		\multirow{2}{*}{${\cal L}_1 \le q_i < {\cal L}_2$ and  $t_i < {\cal T}_1$} & RED & 0.03\\
		& GREEN & 0.04\\\hline
		\multirow{2}{*}{${\cal L}_1 \le q_i < {\cal L}_2$ and  $t_i \ge {\cal T}_1$} & RED & 0.04\\
		& GREEN & 0.03\\\hline
		\multirow{2}{*}{$q_i \ge {\cal L}_2$  and  $t_i < {\cal T}_1$} & RED & 0.05\\
		& GREEN & 0.02\\\hline
		\multirow{2}{*}{$q_i \ge {\cal L}_2$  and  $t_i \ge {\cal T}_1$} & RED & 0.06\\
		& GREEN & 0.01\\\bottomrule
	\end{tabular}
	\label{tab:qtlcfanfs}
\end{table}

Note that the feature selection scheme depends on certain thresholds ${\cal L}_1$ and ${\cal L}_2$ on the queue length and ${\cal T}_1$ on the elapsed times. The motivation for using such graded thresholds is owing to the fact that queue lengths are difficult to measure precisely in practice. We set $({\cal L}_1,{\cal L}_2,{\cal T}_1) = (6,14,130)$ in all our experiments and this choice has been used, for instance, in \cite{prashanth2012threshold}.

\begin{figure}
	\begin{tabular}{cc}
		\begin{subfigure}[b]{0.45\textwidth}
			\centering
			\tabl{c}{\scalebox{0.7}{\begin{tikzpicture}
					\begin{axis}[xlabel={step $k$ of batchTD},ylabel={$\l\theta_{k} - \hat\theta_T\r$}, width=8cm,height=7.25cm,ytick pos=left,xtick pos=left,grid,grid style={gray!30}]
					\addplot table[x index=0,y index=1,col sep=space,each nth point={10}] {l2diff_7x9.txt};
					\end{axis}
					\end{tikzpicture}}\\[1ex]}
			\caption{Tracking error on 7x9-grid network}
			\label{fig:normdiff7x9}
		\end{subfigure}
		&
		\begin{subfigure}[b]{0.45\textwidth}
			\centering
			\tabl{c}{\scalebox{0.7}{\begin{tikzpicture}
					\begin{axis}[xlabel={step $k$ of batchTD},ylabel={$\l\theta_{k} - \hat\theta_T\r$}, width=8cm,height=7.25cm,ytick pos=left,xtick pos=left,grid,grid style={gray!30}]
					\addplot table[x index=0,y index=1,col sep=space,each nth point={10}] {l2diff_14x9.txt};
					\end{axis}
					\end{tikzpicture}}\\[1ex]}
			\caption{Tracking error on 14x9-grid network}
			\label{fig:normdiff14x9}
		\end{subfigure}
	\end{tabular}
	\caption{Tracking error of batchTDQ in iteration $1$ of fLSPI on two grid networks.}
	\label{fig:normdiff-perf}
\end{figure}
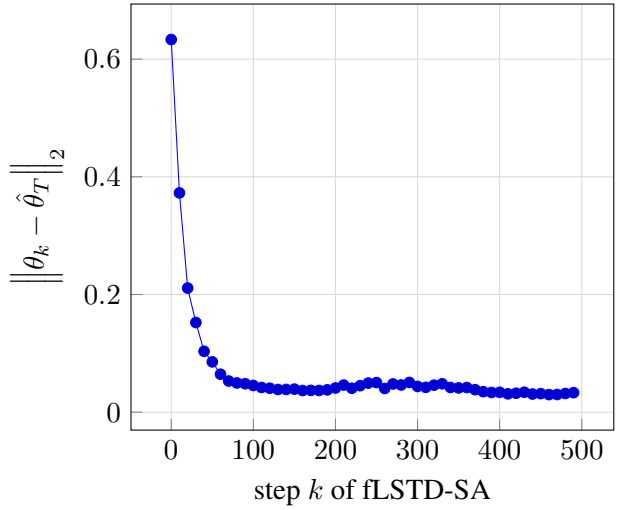
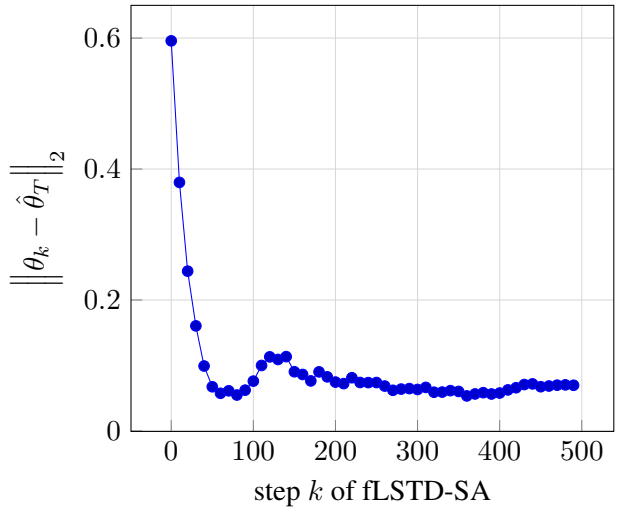

We implement both LSPI as well as fLSPI for the above problem. The experiments involve two stages - an initial training stage where LSPI/fLSPI is run to find an approximately optimal policy and a test stage where ten independent simulations are run using the policy that LSPI/fLSPI converged to in the training stage.  
In the training stage, for both LSPI and fLSPI, we collect $T=10000$ samples from an exploratory policy that picks the actions in a uniformly random manner. For both LSPI and fLSPI, we set $\beta=0.9$ and $\epsilon=0.1$.  We set $\tau$, the number of batchTDQ iterations in fLSPI, to $500$. 
This choice is motivated by an experiment where we observed that at $500$ steps, batchTD is already very close to LSTDQ and taking more steps did not result in any significant improvements for fLSPI.
We implement the regularized variant of LSTDQ, with regularization constant $\mu$ set to $1$.
The step-size $\gamma_k$ used in the update iteration of batchTDQ is set as recommended by Theorem \ref{thm:flstd-rate}. 

\begin{figure}
	\begin{tabular}{cc}
		\begin{subfigure}[b]{0.45\textwidth}
			\centering
			\tabl{c}{\scalebox{0.7}{\begin{tikzpicture}
					\begin{axis}[xlabel={time steps},ylabel={TAR}, smooth,legend pos=south east,width=8cm,height=7.25cm,ytick pos=left,xtick pos=left,grid,grid style={gray!30}]
					\addplot+[thick] table[x index=0,y index=1,col sep=space,each nth point={25}] {LSPI_TAR_7x9_8.txt};
					\addplot+[thick] table[x index=0,y index=1,col sep=space,each nth point={25}] {fLSPISA_TAR_7x9_8.txt};
					\addplot+[thick] table[x index=0,y index=1,col sep=space,each nth point={25}] {FixedExt20_TAR_7x9_1.txt};	
					\addplot+[thick] table[x index=0,y index=1,col sep=space,each nth point={25}] {FixedExt30_TAR_7x9_1.txt};	
					\legend{LSPI,fLSPI,Fixed20,Fixed30}
					\end{axis}
					\end{tikzpicture}}\\[1ex]}
			\caption{Throughput (TAR) on 7x9-grid network}
			\label{fig:tar7x9}
		\end{subfigure}
		&
		\begin{subfigure}[b]{0.45\textwidth}
			\centering
			\tabl{c}{\scalebox{0.7}{\begin{tikzpicture}
					\begin{axis}[xlabel={time steps},ylabel={TAR}, smooth,legend pos=south east,width=8cm,height=7.25cm,ytick pos=left,xtick pos=left,grid,grid style={gray!30}]
					\addplot+[thick] table[x index=0,y index=1,col sep=space,each nth point={25}] {LSPI_TAR_14x9_3.txt};
					\addplot+[thick] table[x index=0,y index=1,col sep=space,each nth point={25}] {fLSPISA_TAR_14x9_1.txt};
					\addplot+[thick] table[x index=0,y index=1,col sep=space,each nth point={25}] {FixedExt20_TAR_14x9_1.txt};	
					\addplot+[thick] table[x index=0,y index=1,col sep=space,each nth point={25}] {FixedExt30_TAR_14x9_1.txt};	
					\legend{LSPI,fLSPI,Fixed20,Fixed30}
					\end{axis}
					\end{tikzpicture}}\\[1ex]}
			\caption{Throughput (TAR) on 14x9-grid network}
			\label{fig:tar14x9}
		\end{subfigure}
	\end{tabular}
	\caption{Performance comparison of LSPI and fLSPI using throughput (TAR) on two grid networks.}
	\label{fig:tar-perf}
\end{figure}
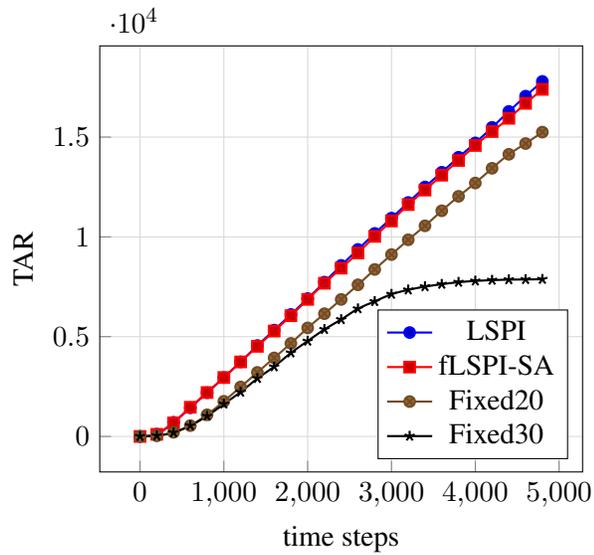
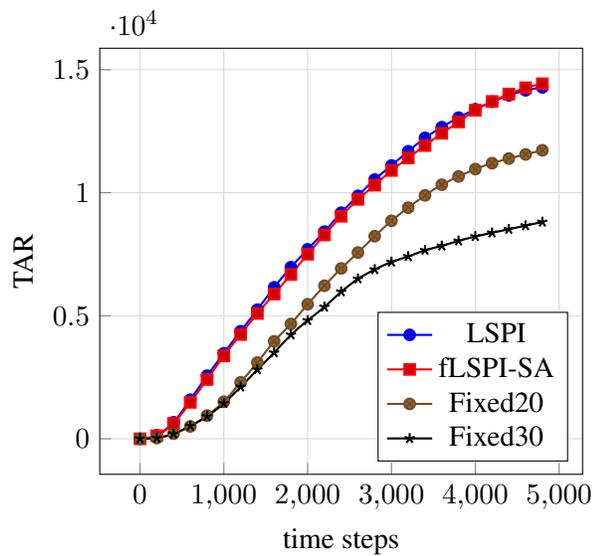

\subsection{Results}
We use total arrived road users (TAR) and runtimes as performance metrics for comparing the algorithms implemented. TAR is a throughput metric that denotes the total number of road users who have reached their destination, while runtimes are measured for the policy evaluation step in LSPI/fLSPI. For batchTDQ, which is the policy evaluation algorithm in fLSPI, we also report the tracking error, which measures the distance in ${\ell}^2$ norm between the batchTD iterate $\theta_k$, $k=1,\ldots,\tau$ and LSTDQ solution $\hat\theta_T$.   

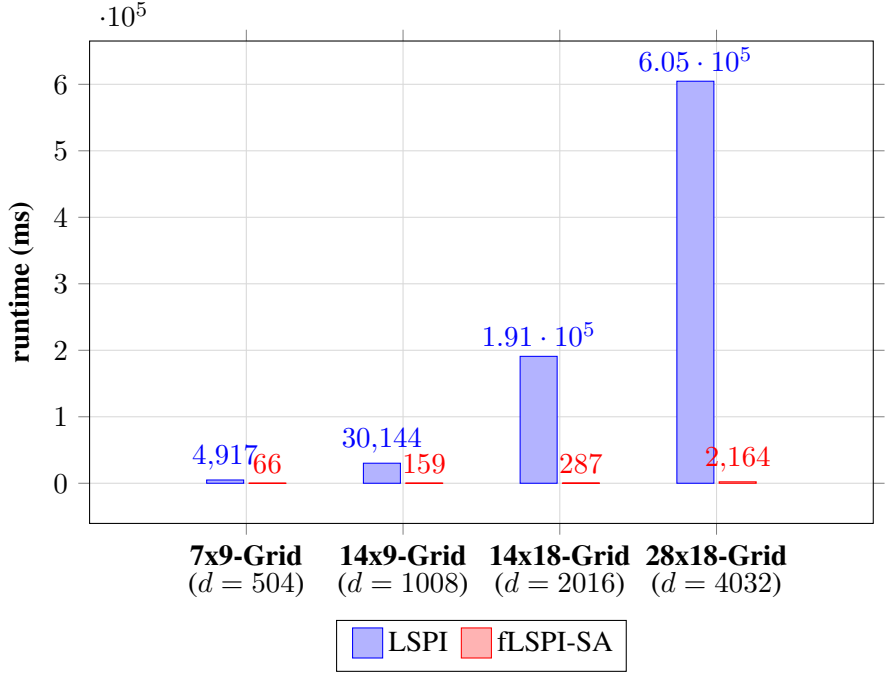
\begin{figure}
	\centering
	\tabl{c}{\scalebox{0.85}{\begin{tikzpicture}
			\begin{axis}[
			ybar={2pt},
			legend style={at={(0.5,-0.2)},anchor=north,legend columns=-1},
			legend image code/.code={\path[fill=white,white] (-2mm,-2mm) rectangle
				(-3mm,2mm); \path[fill=white,white] (-2mm,-2mm) rectangle (2mm,-3mm); \draw
				(-2mm,-2mm) rectangle (2mm,2mm);},
			ylabel={\bf runtime (ms)},
			xlabel={},
			symbolic x coords={0, 1, 2, 3, 4, 5},
			xmin={0},
			xmax={5},
			xtick=data,
			ytick align=outside,
			xticklabels={{\bf7x9-Grid\\[0.5ex]($d=504$),\bf 14x9-Grid\\[0.5ex]($d=1008$),\bf 14x18-Grid\\[0.5ex]($d=2016$),\bf 28x18-Grid\\[0.5ex]($d=4032$)}},
			xticklabel style={align=center},
			bar width=14pt,
			nodes near coords,
			grid,
			grid style={gray!30},
			width=11cm,
			height=7.25cm,
			]
			\addplot   coordinates {  (1,4917) (2,30144) (3,190865) (4,604647)}; 
			\addlegendentry{LSPI}
			\addplot coordinates {  (1,66) (2,159) (3,287) (4,2164)}; 
			\addlegendentry{fLSPI}
			\end{axis}
			\end{tikzpicture}}\\[1ex]}
	\caption{Run-times of LSPI and fLSPI on four road networks}
	\label{fig:runtimes}
\end{figure}

We report the tracking error and total arrived road users (TAR) in Fig. \ref{fig:normdiff-perf} and Fig. \ref{fig:tar-perf}, respectively. The run-times obtained from our experimental runs for LSPI and fLSPI is presented in Fig. \ref{fig:runtimes}. 
Iteration $1$ for fLSPI is used for reporting the tracking error and we observed similar behavior across iterations, i.e., we observed that batchTD iterate $\theta_{\tau}$ is close to the corresponding LSTDQ solution in each iteration of fLSPI.
The experiments are performed for four different grid networks of increasing size and hence, increasing feature dimension. 

From Fig. \ref{fig:normdiff7x9}--\ref{fig:normdiff14x9}, we observe that batchTD algorithm converges rapidly to the corresponding LSTDQ solution.
Further, from the runtime plots (see Fig. \ref{fig:runtimes}), we notice that fLSPI is several orders of magnitude faster than regular LSPI. 
From a traffic application standpoint, we observe in Figs. \ref{fig:tar7x9}--\ref{fig:tar14x9} that fLSPI results in a throughput (TAR) performance that is on par with LSPI. Moreover, the throughput observed for LSPI/fLSPI is higher than that for a traffic light control (TLC) algorithm that cycles through the sign configurations in a round-robin fashion, with a fixed green time period for each sign configuration. We report the TAR results in Figs. \ref{fig:tar7x9}--\ref{fig:tar14x9} for two such fixed timing TLCs with periods $10$ and $20$, respectively denoted Fixed10 and Fixed20. The rationale behind this comparison is that fixed timing TLCs are the de facto standard. Moreover, the results establish that LSPI outperforms fixed timing TLCs that we implemented and fLSPI gives performance comparable to that of LSPI, but at a lower computational cost.   

\section{Extension to Least Squares Regression}
\label{sec:random-batch}
In this section, we describe the classic parameter estimation problem using the method of least squares, the standard approach to solve this problem and the low-complexity SGD alternative.
Subsequently, we outline the fast LinUCB algorithm that uses a SGD iterate in place of least squares solutions and present the numerical experiments for this algorithm on a news recommendation application.

\subsection{Least squares regression and SGD}
In this setting, we are given a set of samples  $\D\triangleq\{(x_i,y_i),i=1,\ldots,T\}$ with the underlying observation model $y_i = x_i\tr \theta^* + \xi_i$ ($\xi_i$ is a bounded, zero-mean random variable, and $\theta^*$ is an unknown parameter).
The least squares estimate $\hat\theta_T$ minimizes $\sum_{i=1}^{T} (y_i - \theta\tr x_i)^2$.
It can be shown that $\hat \theta_T =\bar  A^{-1}_T b_T$, where
$\bar A_T = T^{-1}\sum_{i=1}^{T} x_i x_i\tr$ and  $\bar b_T =  T^{-1} \sum_{i=1}^{T} x_i y_i$.

Notice that, unlike the RL setting, $\hat \theta_T$ here is the minimizer of an empirical loss function.
However, as in the case of LSTD, the computational cost of a Sherman-Morrison lemma based approach for solving the above would be of the order $O(d^2 T)$. 
Similarly to the case of the batchTD algorithm, we update the SGD iterate $\theta_n$ using a SA scheme as follows (starting with an arbitrary $\theta_0$),
\begin{align}
\theta_n = \theta_{n-1} + \gamma_{n}  (y_{i_n} - \theta_{n-1}\tr x_{i_n}) x_{i_n},
\label{eq:random-batch-update}
\end{align}
where, as before, each $i_n$ is chosen uniformly randomly from $\{1,\dots,T\}$, and $\gamma_n$ are step-sizes chosen in advance.  

Unlike batchTD which is a fixed point iteration, the above is a stochastic gradient descent procedure.
Nevertheless, using the same proof template as for batchTD earlier, we can derive bounds on the computational error, i.e., the distance between $\theta_n$ and the least squares solution $\hat\theta_T$, both in high probability as well as expectation. 

\subsection{Main results}
\subsubsection{Assumptions}
As in the case of batchTD, we make some assumptions on the step sizes, features, noise and the matrix $\bar A_T$:\\
\begin{inparaenum}[\bfseries({A}1)]
	\item The step sizes $\gamma_n$ satisfy $\sum_{n}\gamma_n = \infty$, and $\sum_n\gamma_n^2 <\infty$.\\[0.5ex]
	\item Boundedness of $x_i$, i.e., $\l x_i \r \le \Phi_{\max}$,  for $i=1,\ldots,T$.\\[0.5ex]
	\item The noise $\{\xi_i\}$ is i.i.d., zero mean and $|\xi_i|\le \sigma$,  for $i=1,\ldots,T$.\\[0.5ex]
	\item The matrix $\bar A_T$ is positive definite, and its smallest eigenvalue is at least $\mu>0$.\\
\end{inparaenum}
Assumptions (A2) and (A3) are standard in the context of least squares minimization. As for batchTD, in cases when the fourth assumption is not satisfied we can employ either explicit regularization or iterate averaging to produce similar results.

\subsubsection{Asymptotic convergence}
An analogue of Theorem \ref{thm:flstd-asym} holds as follows:
\begin{theorem}
	\label{thm:fls-asym}
	Under (A1)-(A4), the iterate $\theta_n \rightarrow \hat\theta_T$ a.s. as $n \rightarrow \infty$, where $\theta_n$ is given by \eqref{eq:random-batch-update} and $\hat\theta_T = \bar A_T^{-1} \bar b_T$.
\end{theorem}
\begin{proof}
	Follows in exactly the same manner as the proof of Theorem \ref{thm:flstd-asym}.\end{proof}

\subsubsection{Finite time bounds}
An analogue of Theorem \ref{thm:flstd-rate} for this setting holds as follows:
\begin{theorem}[\textit{Error Bound for iterates of SGD}]
	\label{cor:random-batch}
	\ \\Assume (A1)-(A4). Choosing $\gamma_n= \frac{c_0 c}{(c+n)}$ and $c$ such that $c_0\Phi_{\max}^2 \in(0,1)$ and $\mu c_0 c \in (1,\infty)$, for any $\delta >0$,
	\begin{align*}
	\E \l \theta_n - \hat \theta_T \r& \le \dfrac{K_1^{LS}}{\sqrt{n+c}}
	\text{ and } \P\left( \l \theta_n - \hat \theta_T \r \le \dfrac{K_2^{LS}}{\sqrt{n+c}}\right) \ge 1 - \delta,
	\end{align*}
	where
	\begin{gather*}
		K_1^{LS}(n)\triangleq  \frac{\sqrt{c^{c_0c\mu}}\l \theta_0 -\hat \theta_T \r}{(n+c)^{\mu c_0 c - \frac{1}{2}}}
		+ \frac{2ec_0 c h(n)}{2c_0 c \mu - 1},\\ 
		K_2^{LS}(n)\triangleq 2\sqrt{e}c_0 ch(n)\sqrt{\frac{\log{\delta^{-1}}} {\mu c_0 c -1 }} + K_1(n).
		\end{gather*}
		In the above, $h(n)\triangleq \left(\l\theta^*\r + \l\theta_0\r + \sigma\Phi_{\max}\Gamma_n\right)\Phi_{\max}^2 + \sigma\Phi_{\max}.$
\end{theorem}
\begin{proof}
	See Section \ref{sec:proof-ls}. 
	\end{proof}

With step-sizes specified in Theorem \ref{cor:random-batch}, we see that the initial error is forgotten faster than the sampling error, which vanishes at the rate $\tilde O\left(n^{-1/2}\right)$, where $\tilde O(\cdot)$ is like $O(\cdot)$ with the log factors discarded. Thus, the rate derived in Theorem \ref{cor:random-batch} matches the asymptotically optimal convergence rate for SGD type schemes (cf. \cite{nemirovsky1983problem}).

\subsection{Iterate Averaging}
The expectation and high-probability bounds in Theorem \ref{cor:random-batch} as well as earlier works on SGD (cf. \cite{hazan2011beyond})  require the knowledge of the strong convexity constant $\mu$. 
Iterate averaged SGD gets rid of this dependence while exhibiting the optimal convergence rates both in high probability and expectation and this claim is made precise in the following theorem.
\begin{theorem}[\textit{Error Bound for iterate averaged SGD}]\label{cor:ls-iter-av}
	\ \\Under (A2)-(A3), choosing $\gamma_n= c_0\left(\frac{ c}{(c+n)}\right)^\alpha$, with $\alpha \in (1/2,1)$, and $c_0\Phi_{\max}^2 \in(0,1)$, we have, for any $\delta >0$,
	\begin{align}
	\E \l \bar\theta_n - \hat \theta_T \r \le \dfrac{K_1^{IA}(n)}{(n+c)^{\alpha/2}}
	\text{ and } \P\left( \l \bar\theta_n - \hat \theta_T \r \le \dfrac{K_2^{IA}(n)}{(n+c)^{\alpha/2}}\right) \ge 1 - \delta,
	\end{align}
	where, writing $C_0 \triangleq \sum_{n=1}^{\infty}\e(-\mu c_0 c^\alpha n^{1-\alpha}$ and $C_1\triangleq(\e\left(c_0\mu c^\alpha(1+c)^{1-\alpha}\right)$,
		\begin{gather*}
		K_1^{IALS}(n)
		\triangleq C_0\left(C_1\l \theta_0 - \theta_T\r
		+ 2h(n)c^{\alpha}c_0 \left(2 c_0 \mu c^{\alpha}\right)^{\frac{\alpha}{(1-\alpha)}} \sqrt{e}\left(\frac{2\alpha}{1-\alpha}\right)^{\frac{1}{2(1-\alpha)}}\right)\\
		\qquad\qquad\qquad\qquad\qquad
		+ 2 h(n) c^\alpha c_0
		\left( 2c_0\mu c^\alpha\right)^{\frac{\alpha}{2(1-\alpha)}}
		(n+c)^{1-\frac{\alpha}{2}},
		\end{gather*}
		and
		\begin{gather*}
		K_2^{IALS}(n)
		\triangleq\frac{4\sqrt{\log{\delta^{-1}} }}{\mu^2 c_0^2}
		\frac{	
			\frac{1}{\mu}
			\left\{2^\alpha
			+ \left[ \left[\frac{2\alpha}{ c_0\mu c^{\alpha}}\right]^{\frac{1}{1-\alpha}}
			+ \frac{2(1 - \alpha)(c_0\mu)^{\alpha}}{\alpha} \right]
			\right\}
		}
		{
			(n+c)^{(1-\alpha)/2}
		}
		+K_1^{IALS}(n).
		\end{gather*}
\end{theorem}
\begin{proof}
	The proof is similar to that of Theorem \ref{thm:flstd-avg-rate} and is provided in Appendix \ref{sec:appendix-ls}.
	\end{proof}
\begin{remark}
	Note that, unlike in the case of Theorem \ref{thm:flstd-avg-rate}, there is no dependence on a quantity $n_0$ which defines a time when the step sizes have become sufficiently small. This is because for the regression setting here, the assumption that $c_0\Phi_{\max}^2\in (0,1)$ already ensures that the step sizes are sufficiently small. If it was not possible to set $c_0$ in this way, then a similar bound including a dependence on the smallest $n$ such that $\gamma_n\Phi_{\max}^2 < 1$ would be derivable.
\end{remark}

\subsection{Proofs for least squares regression extension}
\label{sec:proof-ls}
The overall schema of the proof here is the same as that used to prove Theorem \ref{thm:flstd-rate}. Proposition \ref{prop:main-ls} below is an analogue of Proposition \ref{prop:main} for the least squares setting. From this proposition the derivation of the rates in Theorem \ref{cor:random-batch} is essentially the same as for Theorem \ref{thm:flstd-rate}. (Recall that in this section $\hat\theta_T=\bar  A^{-1}_T b_T$ is the least squares solution):
\begin{proposition}
	\label{prop:main-ls}
	Let $z_n = \theta_n - \hat \theta_T$, where $\theta_n$ is given by \eqref{eq:random-batch-update}.
	Under (A1)-(A4), and assuming that $\gamma_n\Phi^2_{\max}\le 1$ for all $n$, we have $\forall \epsilon > 0$,
	\begin{itemize}
		\item[(1)] a bound in \textbf{\textit{high probability}} for the \textbf{\textit{centered error}}:
		\begin{align}
		\P &\left( \l z_n \r - \E \l z_n \r \ge \epsilon \right)
		\le  \e\left(- \frac{\epsilon^2}
		{4 h(n)^2\sum_{i=1}^{n} L^2_i}
		\right),\label{eq:ls-high-prob-bound}
		\end{align}
		where
			\begin{gather*}
			L_i \triangleq \gamma_i \prod_{j=i}^{n-1} (1 - \gamma_{j+1} \mu (2-\Phi_{\max}^2\gamma_{j+1}))^{1/2},\\
			h(n)\triangleq\left(\l\theta^*\r + \l\theta_0\r + \sigma\Phi_{\max}\Gamma_n\right)\Phi_{\max}^2 + \sigma\Phi_{\max},
			\end{gather*}
			and $\Gamma_n \triangleq \sum_{i=1}^n \gamma_i$.
		\item[(2)] and a bound in \textbf{\textit{expectation}} for the \textbf{\textit{non-centered error}}:
		\begin{align}
		\E\left(\l z_n\r\right)^2
		\le&\underbrace{\prod_{j=1}^n\left(1 - \mu\gamma_j\right) \l \theta_0 - \hat\theta_T \r
		}_{\textbf{initial error}}\nonumber\\
		&+\underbrace{\left(\sum_{k=1}^{n-1}
			4 h(k)^2\gamma^2_{k+1}
			\left[\prod_{j=k+1}^n\left( 1 - \mu\gamma_j \right)\right]^2
			\right)^{\frac{1}{2}}
		}_{\textbf{sampling error}}\label{eq:ls-expectation-bound}.
		\end{align}
	\end{itemize}
\end{proposition}

The proof of the Proposition \ref{prop:main-ls} has the same scheme as the proof of Proposition \ref{prop:main}. The major difference is that the update rule is no longer the update rule of a fixed point iteration, but of a gradient descent scheme. 
In the following proofs, we give only the major differences with the proof of Proposition \ref{prop:main}:
\begin{description}
	\item[\textbf{High-probability bound.}]  There are two alterations to the proof of the high probability bound in Proposition \ref{prop:main}: slightly different Lipschitz constants are derived according to the different form of the random innovation (Step 2 of the proof of Proposition \ref{prop:main}); the constant by which the the size of the random innovations is bounded is different, and projection is not necessary to achieve this bound (Step 3 of the proof of Proposition \ref{prop:main}). 
	\item[\textbf{Bound in expectation.}] The overall scheme of this proof is similar to that used in proving the expectation bound in Proposition \ref{prop:no-proj}. However, we see differences in the proof wherever the update rule is unrolled and bounds on the various quantities in the resulting expansion need to be obtained.
\end{description}

\paragraph{\textbf{Proof of Proposition \ref{prop:main-ls} part (1):}}
\begin{proof}
	First we derive the Lipschitz dependency of the $i^{th}$ iterate on the random innovation at time $j<i$, as in Step 2 of Proposition \ref{prop:main}.
	
	Let $\Theta_j^i(\theta)$ denote the mapping that returns the value of the iterate updated according to \eqref{eq:random-batch-update} at instant $j$, given that $\theta_i = \theta$. Now we note that
	\begin{align*}
	\Theta_n^i(\theta) - \Theta_n^i(\theta')
	= \left(I - \gamma_n x_{i_n}x_{i_n}^T\right)
	\left[\Theta_{n-1}^i(\theta) - \Theta_{n-1}^i(\theta')\right]
	\end{align*}
	and
	\begin{align*}
	\left(I - \gamma_n x_{i_n}x_{i_n}^T\right)^T
	\left(I - \gamma_n x_{i_n}x_{i_n}^T\right)
	= \left(I - \gamma_n (2-\|x_{i_n}\|_2^2\gamma_n)x_{i_n}x_{i_n}^T\right).
	\end{align*}
	So using Jensen's inequality, the tower property of conditional expectations, and Cauchy-Schwarz, we can deduce that
	\begin{align}
	\E&\left[ \| \Theta_n^i(\theta) - \Theta_n^i(\theta')\|_2 \mid
	\Theta_{n-1}^i(\theta), \Theta_{n-1}^i(\theta')\right]\nonumber\\
	&\le\left[  \| I- \gamma_n(2-\Phi_{\max}^2\gamma_n)\bar{A}_T\|_2^2
	\| \Theta_{n-1}^i(\theta) - \Theta_{n-1}^i(\theta')  \|_2^2
	\right]^{1/2} \label{eq:for-unrolling-ls}
	\end{align}
		Notice that since $ \gamma_n\Phi_{\max}^2 \in (0, 1)$, the largest eigenvalue of $\gamma_n \bar{A}_T$ must be less than $1$. Hence, a repeated application of \eqref{eq:for-unrolling-ls}, together with (A1) yields the following
	\begin{align*}
	\E \left[ \l \Theta_{n}^i(\theta) - \Theta_{n}^i(\theta') \r^2\right]  \le \l \theta - \theta' \r^2 \prod\limits_{j=i}^{n-1} (1- \mu\gamma_{j+1}(2- \Phi_{\max}^2\gamma_{j+1})).
	\end{align*}
	Finally putting all this together, if $f$ and $f'$ denote two possible values for the random innovation at time $i$, and letting $\theta = \theta_{i-1} + \gamma_i f$ and $\theta' = \theta_{i-1} + \gamma_i f'$, then we have
	\begin{align*}
	&\l \E\left[ \l \theta_n - \hat\theta_T \r \left| \theta_{i} = \theta\right.\right]- \E\left[ \l \theta_n - \hat\theta_T \r \left| \theta_{i} = \theta' \right.\right] \r \\
	& \le  \E\left[ \l \Theta^m_n\left(\theta\right) - \Theta^m_n\left(\theta'\right) \r\right]
	\le  \left( \prod\limits_{j=i}^{n-1} (1- \mu \gamma_{j+1}(2 - \Phi_{\max}^2\gamma_{j+1})) \right)^{\frac{1}{2}} 
	\gamma_i \l f - f'\r\\
	&=  L_i \l f - f'\r.
	\end{align*}
	
	Finally we need to bound the size of the random innovations. Recall that in Proposition \ref{prop:main}, the bound on the size of the iterates followed from the projection step in the algorithm. In this case, we can derive a bound for the iterates directly:
	\begin{align}\label{eq:ls-it-bound}
		\l \theta_n\r
		=& \left\| \left[\prod_{k=1}^n(I - \gamma_{k}x_{i_k}x_{i_k}\tr)\right]
		\theta_0  + \sum_{k = 1}^n\gamma_k
		\left[\prod_{j=k}^n(I - \gamma_{j}x_{i_j}x_{i_j}\tr)\right]\xi_k x_k
		\right\|_2\nonumber\\
		\le& \l \theta_0 \r + \sigma\Phi_{\max}\sum_{j = 1}^n\gamma_j,
		\end{align}
		where we have used that $\gamma_{j}x_{i_j}x_j\tr$ is a positive semi-definite matrix. Now we can bound the random innovation by
	\begin{align*}
	\l(y_{i_n} - \theta_{n-1}\tr x_{i_n})x_{i_n} \r
	&= \l(x_{i_n}\tr\theta^* + \xi_{i_n} - \theta_{n-1}\tr x_{i_n})x_{i_n} \r\\
	&\le \left(\l\theta^*\r + \l\theta_0\r + \sigma\Phi_{\max}\Gamma_n\right)\Phi_{\max}^2 + \sigma\Phi_{\max} = h(n),
	\end{align*}
	where $\Gamma_n: = \sum_{k=0}^n\gamma_k$.
	The proof now follows just as in Proposition \ref{prop:main}.
	\end{proof}
\paragraph{\textbf{Proof of Proposition \ref{prop:main-ls} part (2):}}
\begin{proof}
	First we extract a martingale difference from the update rule \eqref{eq:random-batch-update}. Let $f_{n}(\theta) \triangleq  (x_{i_n} - (\theta-\hat\theta_T)\tr x_{i_n})x_{i_n}$, and let $F(\theta) \triangleq \E(f_n(\theta)\mid \F_{n-1})$, where $\F_{n-1}$ is the $\sigma$-field generated by the random variables $\{i_1,\dots,i_{n-1}\}$ as before. Then
	\begin{align*}
	z_n = \theta_n - \hat\theta_T = \theta_{n-1} - \hat\theta_T - \gamma_n\left(F(\theta_{n-1})-\Delta M_n\right),
	\end{align*}
	the $\Delta  M_{n} =  F(\theta_{n-1}) - f_{n}(\theta_{n-1})$ is a martingale difference.
	
	Now since $\hat\theta_T$ is the least squares solution, $F(\hat\theta_T)=0$. Moreover $F(\cdot)$ is linear, and so we obtain a recursion:
	\begin{align*}
	z_n = z_{n-1} - \gamma_n\left(z_{n-1}\bar{A}_T-\Delta M_n\right)
	=  \tpi_{1}^{n} z_0 - \sum_{k=1}^{n}\gamma_k\tpi_{k+1}^{n}\Delta M_k,
	\end{align*}
	where 
	$\tpi_{k}^{n} \triangleq \prod_{j=k}^{n}\left(I - \gamma_j \bar A_T\right)$. 
	
	By Jensen's inequality, we have
	\begin{align}
	\E(\l z_n\r) &\le \left( \E (\langle z_n, z_n \rangle )\right)^\frac{1}{2}
	=  \left( \E \l \tpi_{1}^{n} z_0 \r^2 +  \sum_{k=1}^{n}\gamma_k^2 \E \l\tpi_{k+1}^n\Delta M_k\r^2 \right)^\frac{1}{2}\label{eq:thm7-it}
	\end{align}
	
		Notice that the largest eigenvalue of $\gamma_n \bar A_T$ is smaller than $1$, since $\gamma_n\Phi_{\max}^2 \in (0, 1)$. So,  $I - \gamma_n\bar A_T$ is positive definite, and, by (A1), has largest eigenvalue $1 - \gamma_n \mu$. Hence
		\begin{align}
		\ml \tpi_{k+1}^n \mr =& \ml\prod_{j=k+1}^{n}\left(I - \gamma_j \bar A_T\right)\mr
		\le \prod_{j=k+1}^{n} (1-\gamma_j\mu).\label{eq:thm2-pi}
		\end{align}
	Finally we need to bound the variance of the martingale difference.
	Using (A2) and (A3), a calculation shows that
	\begin{gather*}
	\E_{\xi,i_t}\langle f_{i_t}(\theta_{t-1}), f_{i_t}(\theta_{t-1})\rangle,
	\E_{\xi}\langle  F(\theta_{t-1}),  F(\theta_{t-1})\rangle
	\le h(n),
	\end{gather*}
	where we have used the bound in \eqref{eq:ls-it-bound}.
	Hence $\E [\l  \Delta M_n \r^2]\le 4h(n)^2$.
	
	The result now follows from \eqref{eq:thm7-it} and \eqref{eq:thm2-pi}.
	\end{proof}

\section{Fast LinUCB using SA and application to news-recommendation}
\label{sec:flinucb}
\subsection{Background for LinUCB}

\tikzstyle{block} = [draw, fill=white, rectangle,
minimum height=3em, minimum width=6em]
\tikzstyle{sum} = [draw, fill=white, circle, node distance=1cm]
\tikzstyle{input} = [coordinate]
\tikzstyle{output} = [coordinate]
\tikzstyle{pinstyle} = [pin edge={to-,thin,black}]

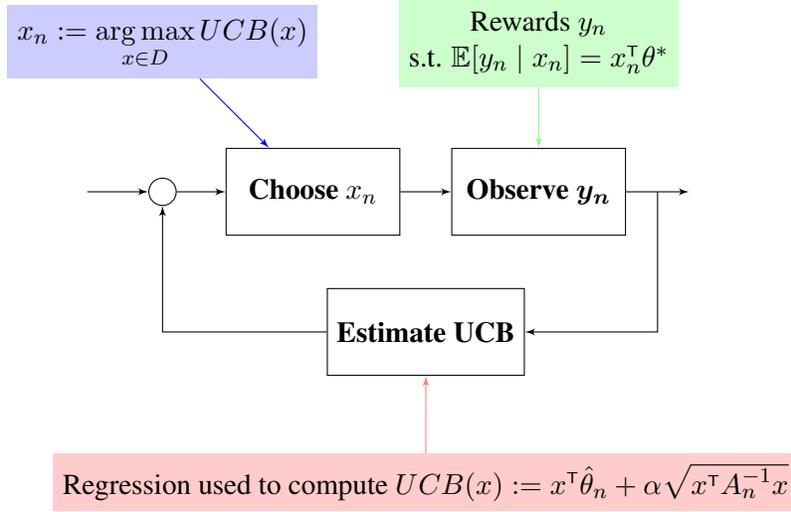
\begin{figure}
	\centering
	\scalebox{1.0}{\begin{tikzpicture}[auto, node distance=2cm,>=latex']
		\node [input, name=input] {};
		\node [sum, right of=input] (sum) {};
		\node [block, right of=sum] (arms) {\makecell{\bf Choose $x_n$}};
		\node [block, right of=arms,
		node distance=3cm] (losses) {\textbf{Observe $\boldsymbol{y_n}$}};
		\draw [->] (arms) -- node[name=u] {} (losses);
		\node [output, right of=losses] (output) {};
		\node [block, below of=u] (ols) {\textbf{Estimate UCB}  };
		%
		\draw [draw,->] (input) -- node {} (sum);
		\draw [->] (sum) -- node {} (arms);
		\draw [->] (losses) -- node [name=y] {}(output);
		\draw [->] (y) |- (ols);
		\draw [->] (ols) -| node[pos=0.99] {}
		node [near end] {} (sum);    
		\node[fill=blue!20,anchor=base,above of=sum] (t1)
		{\makecell{$x_n \triangleq \argmax\limits_{x\in D} UCB(x)$}} edge[->,
		color=blue] (arms);
		\node[fill=green!20,anchor=base,above of=losses] (t2)
		{\makecell[c]{Rewards $y_n$ \\s.t. $\E[y_n \mid x_n] = x_n\tr\theta^*$}} edge[->,
		color=green!50] (losses);
		\node[fill=red!20,anchor=base,below of=ols] (t3)
		{Regression used to compute $UCB(x)\triangleq x\tr \hat\theta_n + \alpha\sqrt{ x\tr A_n^{-1} x}$}  edge[->,
		color=red!50] (ols);
		\end{tikzpicture}}	\caption{Operational model of LinUCB}
	\label{fig:linucb}
\end{figure}

As illustrated in Fig. \ref{fig:linucb}, at each iteration $n$, 
the objective is to choose an article from a pool of $K$ articles with respective features $x_{1}(n),\ldots,x_{K}(n)$. Let $x_n$ denote the chosen article at time $n$.
LinUCB computes a regularized least squares (RLS) solution $\hat\theta_n$ based on the chosen arms $x_i$ and rewards $y_i$ seen so far, $i=1,\ldots,n-1$ as follows:
\begin{align}
\hat\theta_n = \argmin_\theta \sum_{i=1}^{n} (y_i - \theta\tr x_i)^2 + \lambda \l \theta \r^2.
\end{align}
Note that $\{x_i,y_i\}$ do not come from a distribution. Instead, at every iteration $n$, the arm $x_n$ chosen by LinUCB is based on the RLS solution $\hat\theta_n$.
The latter is used to estimate the UCB values for each of the $K$ articles as follows:
	\begin{align}
	\text{UCB}(x_k(n))\triangleq x_k(n)\tr \hat\theta_n + \kappa\sqrt{ x_k(n)\tr A_n^{-1} x_k(n)}, k=1,\ldots,K.
	\label{eq:ucb} 
	\end{align}
The algorithm then chooses the article with the largest UCB value and the cycle is repeated.

\algblock{FastLS}{EndFastLS}
\algnewcommand\algorithmicFastLS{\textbf{\em Approximate Least Squares Regression using fLS-SA}}
\algnewcommand\algorithmicendFastLS{}
\algrenewtext{FastLS}[1]{\algorithmicFastLS\ #1}
\algrenewtext{EndFastLS}{\algorithmicendFastLS}

\algblock{UCBcompute}{EndUCBcompute}
\algnewcommand\algorithmicUCBcompute{\textbf{\em UCB computation using SGD}}
\algnewcommand\algorithmicendUCBcompute{}
\algrenewtext{UCBcompute}[1]{\algorithmicUCBcompute\ #1}
\algrenewtext{EndUCBcompute}{\algorithmicendUCBcompute}

\algtext*{EndFastLS}
\algtext*{EndUCBcompute}

\begin{algorithm}[t]  
	\caption{fLinUCB-SA}
	\label{alg:LinUCBwithBGD}
	\begin{algorithmic}
		\State {\bfseries Initialization:} Set $\theta_0$, $\lambda>0$ - the regularization parameter, $\gamma_k$ - the step-size sequence.
		\For{$n = 1,2,\ldots$}
		\State Observe article features $x_{1}(n),\ldots,x_{K}(n)$
		\FastLS
		\For{$l=1 \dots \tau$}
		\State Get random sample index: $i_l\sim U(\{1,\dots,n-1\})$
		\State Update fLS-SA iterate $\theta_l(n)$ as follows:
		\State $\theta_l(n) = \theta_{l-1}(n) + \gamma_{l}  (y_{i_l} - \theta_{l-1}(n)\tr x_{i_l}) x_{i_l} - \gamma_l \frac{\lambda}{n} \theta_{l-1}(n)$ 
		\EndFor
		\EndFastLS
		\UCBcompute
		\For{$k=1 \dots K$}
		\For{$l=1 \dots \tau'$}
		\State Get random sample index: $i_l\sim U(\{1,\dots,n-1\})$
		\State Update SGD iterate $\phi_k(n)$ as follows:
		\State $\phi_k(l) = \phi_k(l-1) + \gamma_l (n^{-1}x_k(n) - (\phi_k(l-1)\tr x_{i_l})x_{i_l}),$ 
		\EndFor
		\EndFor
		\State Choose article achieving $\argmax_{k=1,\ldots,K} \theta_\tau(n)\tr x_k(n)  + \kappa\sqrt{ \phi_k(\tau')\tr x_k(n)}$
		\State Observe the reward $y_n$.
		\EndUCBcompute
		\EndFor
	\end{algorithmic}
\end{algorithm}

\subsection{Fast LinUCB using SA (fLinUCB-SA)}
We implement a fast SGD variant of LinUCB, where SGD is used for two purposes (See Algorithm \ref{alg:LinUCBwithBGD} for the pseudocode):
\begin{description}
	\item[\textbf{Least squares approximation.}] Here we use fLS-SA as a subroutine to approximate $\hat\theta_n$.  In particular, at any instant $n$ of the LinUCB algorithm, we run the update \eqref{eq:random-batch-update} for $\tau$ steps and use the resulting $\theta_{\tau}$ to derive the UCB values for each arm.
	\item[\textbf{UCB confidence term approximation.}] Here we use an SGD scheme, originally proposed in \cite{prashanth2014fastLS}, for approximating the confidence term of the UCB values \eqref{eq:ucb}.
	For a given arm $k=1,\ldots,K$, let $\hat\phi_k(n) = A_n^{-1} x_k(n)$ denote the confidence estimate in the UCB value \eqref{eq:ucb}. Recall that  $A_n = \sum\limits_{i=1}^n x_i x_i\tr$. It is easy to see that $\hat\phi_k(n)$ is the solution to the following problem:
	\begin{align}
	\label{eq:confidence-pb}
	\min_\phi \sum\limits_{i=1}^n \dfrac{(x_i\tr \phi)^2}{2} - \dfrac{x_k(n)\tr\phi}{n}.
	\end{align}
	Solving the above problem incurs a complexity of $O(d^2)$. An SGD alternative with a per-iteration complexity of $O(d)$ approximates the solution to \eqref{eq:confidence-pb} by using the following iterative scheme:
	\begin{align}
	\label{eq:confidence-gd}
	\phi_k(l) = \phi_k(l-1) + \gamma_l (n^{-1}x_k(n) - (\phi_k(l-1)\tr x_{i_l})x_{i_l}),
	\end{align}
	where $i_l$ is chosen uniformly at random in the set $\{1,\ldots,n\}$.
\end{description}

For fLinUCB-SA in both the simulation setups presented subsequently, we set $\lambda$ to $1$, $\kappa$ to $1$, $\tau,\tau'$ to $100$ and $\theta_0$ to the $d=136$ ${\mathbf 0}$ vector. Further, the step-sizes $\gamma_k$ are chosen as $c/(2(c+k))$, with $c=1.33n$ and this choice is motivated by Theorem \ref{cor:random-batch}. 
\begin{remark}
	The choice of the number of steps $\tau,\tau'$ for SGD schemes in fLinUCB-SA is an arbitrary one. Our aim is simply to show that using a stochastic approximation iterates in place of an exact solution to the least squares and confidence estimates does not significantly decrease performance of LinUCB, while it does drastically decrease the complexity.
\end{remark}

\subsection{Experiments on Yahoo! dataset} 
The motivation in this experimental setup is to establish the usefulness of fLS-SA in a higher level machine learning algorithm such as LinUCB. In other words, the objective is to test the performance of LinUCB with SGD approximating least squares and show that the resulting algorithm gains in runtime, while exhibiting comparable performance to that of regular LinUCB.

\begin{figure}[h]
	\centering
	\includegraphics[width=3in]{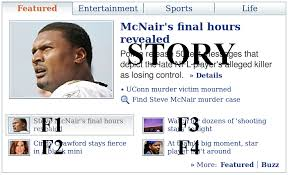}
	\caption{The {\em Featured} tab in Yahoo! Today module (src: \cite{li2010contextual})}
	\label{fig:today}
\end{figure}

For conducting the experiments, we use the framework provided by the ICML exploration and exploitation challenge \cite{li12}, based on the user click log dataset \cite{webscope} for the Yahoo! front page today module (see Fig. \ref{fig:today}). 
We run each algorithm on several data files corresponding to different days in October, 2011. 

Each data file has an average of nearly two million records of user click information. Each record in the data file contains various information obtained from a user visit. These include the displayed article, whether the user clicked on it or not, user features and a list of available articles that could be recommended. The precise format is described in \cite{li12}. The evaluation of the algorithms in this framework is done in an off-line manner using a procedure described in \cite{li2011unbiased}. 

\begin{figure}
	\centering
	\tabl{c}{\scalebox{1}{\begin{tikzpicture}
			\begin{axis}[xlabel={iteration $k$ of fLS-SA},ylabel={$\l\theta_{k}(n) - \hat\theta_n\r^2$}, smooth]
			\addplot table[x index=0,y index=1,col sep=comma] {day2_iter165.txt};
			\addlegendentry{$\l\theta_{k}(n) - \hat\theta_n\r^2$}
			\end{axis}
			\end{tikzpicture}}\\[1ex]}
	\caption{Distance between fLS-SA iterate $\theta_{k}(n)$ and $\hat\theta_n$ in iteration $n=165$ of fLinUCB-SA, with day $2$'s data file as input.}
	\label{fig:normdiff-salsa}
\end{figure}
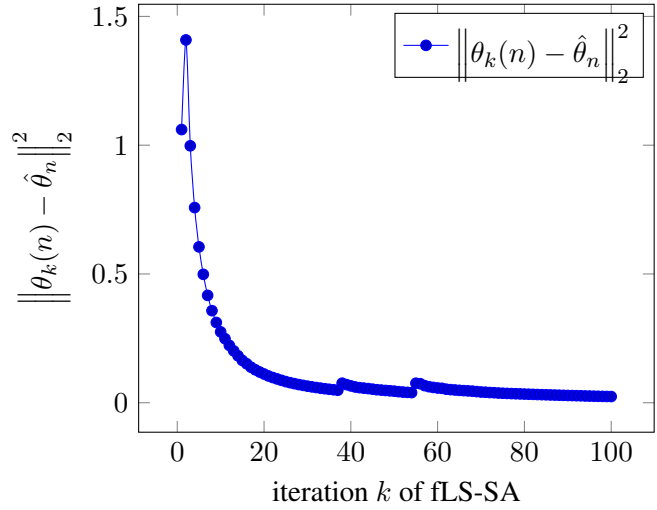

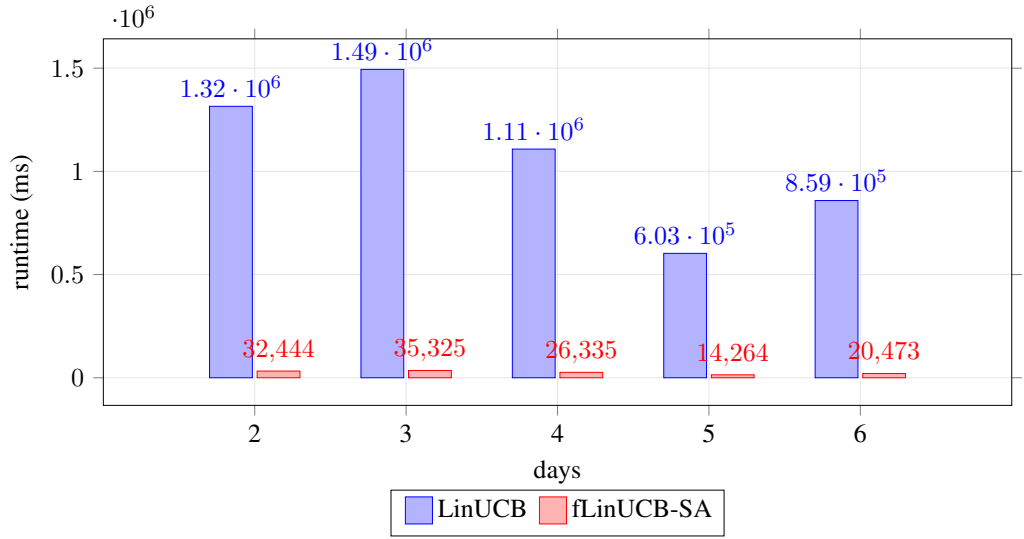
\begin{figure}
	\centering
	\tabl{c}{\scalebox{0.9}{\begin{tikzpicture}
			\begin{axis}[
			ybar={2pt},
			legend style={at={(0.5,-0.2)},anchor=north,legend columns=-1},
			legend image code/.code={\path[fill=white,white] (-2mm,-2mm) rectangle
				(-3mm,2mm); \path[fill=white,white] (-2mm,-2mm) rectangle (2mm,-3mm); \draw
				(-2mm,-2mm) rectangle (2mm,2mm);},
			ylabel={runtime (ms)},
			xlabel={days},
			symbolic x coords={0,  2, 3, 4, 5, 6, 7},
			xmin={0},
			xmax={7},
			xtick=data,
			ytick align=outside,
			bar width=15pt,
			nodes near coords,
			grid,
			grid style={gray!20},
			width=13cm,
			height=7cm,
			]
			\addplot   coordinates { (2,1315214) (3,1494127) (4,1107987) (5,602762) (6,858698)}; 
			\addplot coordinates { (2,32444) (3,35325) (4,26335) (5,14264) (6,20473) }; 
			\legend{LinUCB, fLinUCB-SA}
			\end{axis}
			\end{tikzpicture}}\\[1ex]}
	\caption{Performance comparison of the algorithms using runtimes on various days of the dataset.}
	\label{fig:runtime-compare}
\end{figure}

\paragraph{Results.}
We report the tracking error and runtimes from our experimental runs in Figs. \ref{fig:normdiff-salsa} and \ref{fig:runtime-compare}, respectively. As in the case of batchTDQ, the tracking error is the distance in ${\ell}^2$ norm between the fLS-SA iterate $\theta_n$ and the RLS solution $\hat\theta_n$ at each instant $n$ of the LinUCB algorithm. The runtimes in Fig. \ref{fig:runtime-compare} are for five different data files corresponding to five days in October, 2009 of the dataset \cite{webscope} and compare the classic RLS solver time against fLS-SA time for each day of the dataset considered. 

From Fig. \ref{fig:normdiff-salsa}, we observe that, in iteration $n=165$ of the LinUCB algorithm, fLS-SA algorithm iterate $\theta_{\tau}(n)$ converges rapidly to the corresponding RLS solution $\hat\theta_n$. The choice $165$ for the iteration is arbitrary, as we observed similar behavior across iterations of LinUCB.

The CTR score value is the ratio of the number of clicks that an algorithm gets to the total number of iterations it completes, multiplied by $10000$ for ease of visualization. We observed that the CTR score for the regular LinUCB algorithm with day $2$'s data file as input was $470$, while that of fLinUCB-SA was $390$, resulting in about $20\%$ loss in performance. Considering that the dataset contains very sparse features and also the fact that the rewards are binary, with a reward of $1$ occurring rarely, we believe LinUCB has not seen enough data to have converged UCB values and hence the observed loss in CTR may not be conclusive.

\section{Conclusions and Future Work}
\label{sec:conclusions}

We analyzed the TD algorithm with linear function approximation, under uniform sampling from a dataset. 
We provided convergence rate results for this algorithm, both in high probability and in expectation. Furthermore, we also established that using our batchTD scheme in place of LSTD does not impact the rate of convergence of the approximate value function to the true value function. These results coupled with the fact that the batchTD algorithm possesses lower computational complexity in comparison to traditional techniques makes it attractive for implementation in {\em big data} settings, where the feature dimension is large, regardless of the density of the feature vectors.
On a traffic signal control application, we demonstrated the practicality of a low-complexity alternative to LSPI that uses batchTDQ in place of LSTDQ for policy evaluation. We also extended our analysis for bounding the error of an SGD scheme for least squares regression and conducted a set of experiments that combines the SGD scheme with the LinUCB algorithm on a news-recommendation platform. 

Unlike LSTD, TD is an online algorithm and a finite-time analysis there would require notions of mixing time for Markov chains in addition to the solution scheme that we employed in this work. This is because the asymptotic limit for TD(0) is the fixed point of the Bellman operator, which assumes that the underlying MDP is begun from the stationary distribution, say $\Psi$. However, the samples provided to TD(0) come from simulations of the MDP that are not begun from $\Psi$, making the finite time analysis challenging. It would be an interesting future research direction to use the proof technique employed to analyze batchTD, and incorporate the necessary deviations to handle the more general Markov noise.  

We outline a few future research directions for improving batchTD algorithm developed here:
(i) develop extensions of batchTD to approximate LSTD($\lambda$); (ii) choose a cyclic sampling scheme instead of the uniform random sampling. Cycling through the samples is advantageous because the samples need not be stored and one can then think of batchTD with cyclic sampling as an incremental algorithm in the spirit of TD; and 
(iii) leverage recent enhancements to SGD in the context of least squares regression, cf. \cite{dieuleveut2016harder}. An orthogonal direction of future research is to develop online algorithms that track the corresponding batch solutions, efficiently and this has been partially accomplished in \cite{korda2015aaai} and \cite{tarres2011online}. 
\appendix
\section*{Appendix}

\section{Proof of Theorem \ref{cor:ls-iter-av}}
\label{sec:appendix-ls}
The proof of Theorem \ref{cor:ls-iter-av} relies on a general rate result built from Proposition \ref{prop:main-ls}
\begin{proposition}
	\label{prop:avg-hpb-IA-ls}
	Under (A1)-(A3) we have, for all $\epsilon \ge 0$ and $\forall n\geq 1$,
	\begin{align*}
	&\P(   \l z_n \r - \E \l z_n \r \ge \epsilon )
	\le \e\left(- \dfrac{\epsilon^2}
	{4h(n)^2\sum\limits_{m=1}^{n} L_m^2}
	\right),
	\end{align*}
		where
		$L_i \triangleq \frac{\gamma_i}{n}
		\left( \sum_{l=i+1}^{n-1}\prod\limits_{j=i}^{l}
		\left(1- \mu \gamma_{j+1}( 2 - \Phi_{\max}^2\gamma_{j+1}))
		\right)^{1/2}\right)$,
	and $h(n)$ is as in Proposition \ref{prop:main-ls}.
\end{proposition}
\begin{proof}
This proof follows exactly the proof of Proposition \ref{prop:flstd-avg-hpb}, except that it uses the form of $L_i$ for non-averaged iterates as derived in Proposition \ref{prop:main-ls} part (1), rather than as derived in Proposition \ref{prop:main} part (1).
\end{proof}

We specialise this result with the choice of step size $\gamma_n \triangleq (c_0 c^{\alpha})/(n+c)^{\alpha}$.
First, we prove the form of the $L_i$ constants for this choice of step size:

\begin{lemma}
	\label{lemma:avg-hpb-ls}
	Under conditions of Theorem \ref{cor:ls-iter-av}, we have 
	\begin{align*}
	\sum_{i=1}^n L_i^2
	\le
	\frac{1}{\mu^2}
	\left\{2^\alpha
	+ \left[ \left[\frac{2\alpha}{ c_0\mu c^{\alpha}}\right]^{\frac{1}{1-\alpha}}
	+ \frac{2(1 - \alpha)(c_0\mu)^{\alpha}}{\alpha} \right]
	\right\} ^2\frac{1}{n}.
	\end{align*}
\end{lemma}

Second, we bound the expected error by directly averaging the errors of the non-averaged iterates:
\begin{align}
\E\l \bar\theta_{n+1} - \hat\theta_T\r \le \frac{1}{n}\sum_{k = 1}^n\E\l\theta_k - \hat\theta_T \r,
\end{align}
and directly applying the bounds in expectation given in Proposition \ref{prop:main}.
\begin{lemma}
	\label{lemma:avg-exp-ls}
	 Under conditions of Theorem \ref{cor:ls-iter-av}, we have 
		\begin{align*}
		\E\l \bar\theta_n - \hat\theta_T\r
		\le& C_0\left(C_1\l \theta_0 - \theta_T\r
		+ 2h(n)c^{\alpha}c_0 \left(2 c_0 \mu c^{\alpha}\right)^{\frac{\alpha}{(1-\alpha)}} \sqrt{e}\left(\frac{2\alpha}{1-\alpha}\right)^{\frac{1}{2(1-\alpha)}}\right)\frac{1}{n}\\
		&+  h(n) c^\alpha c_0
		\left( 2c_0\mu c^\alpha\right)^{\frac{\alpha}{2(1-\alpha)}}
		(n+c)^{-\frac{\alpha}{2}},
		\end{align*}
		where $C_0$ and $C_1$ are as defined in Theorem \ref{cor:ls-iter-av}.
\end{lemma}

\subsection{Proof of Lemma \ref{lemma:avg-hpb-ls}}
\begin{proof}
Recall from the statement of Theorem \ref{cor:ls-iter-av} that
\begin{align}
0 < c_0 \Phi_{\max}^2 < 1.
\label{eq:largen-ls}
\end{align}
Recall also from the formula in Proposition \ref{prop:avg-hpb-IA-ls}, that
\begin{align*}
L_i
= \frac{\gamma_i}{n}
\left( \sum_{l=i+1}^{n-1}\prod\limits_{j=i}^{l}
\left(1- \mu\gamma_{j+1}( 2 - \Phi_{\max}^2\gamma_{j+1}))
\right)^{1/2}\right).
\end{align*}
Notice that
\begin{align*}
\sum_{i = 1}^n L_i^2
&= \sum_{i = 1}^n\left[\frac{\gamma_i}{n}
\left( \sum_{l=i+1}^{n-1}\prod\limits_{j=i}^{l}
\left(1- \mu\gamma_{j+1}( 2 - \Phi_{\max}^2\gamma_{j+1}))
\right)^{1/2}\right)
\right]^2\\
&\le \frac{1}{n^2}\sum_{i = 1}^n\left[ \gamma_i
\left( \sum_{l = i+1}^{n-1}
\e\left(- \sum_{j=i}^l \mu\gamma_{j+1}(2 - \Phi_{\max}^2\gamma_{j+1}))
\right)\right)
\right]^2\\
& < \frac{1}{n^2}\sum_{i = 1}^n
{\underbrace{
		\left[
		c_0\left(\frac{c}{c+i}\right)^\alpha
		\left( \sum_{l = i+1}^{n-1}
		\e\left(- c_0\mu\sum_{j=i}^l \left(\frac{c}{c+j}\right)^\alpha
		\right)\right)
		\right]}_{\triangleq(A)}}^2.
\end{align*}
To produce the final bound, we bound the summand (A) highlighted in line \eqref{eq:it-av-1} by a constant, uniformly over all values of $i$ and $n$, exactly as in the proof of Lemma \ref{lemma:avg-hpb}. Thus, we have
\begin{align*}
\sum_{i=1}^n L_i^2
\le \frac{1}{\mu^2}
\left\{2^\alpha
+ \left[ \left[\frac{2\alpha}{ c_0\mu c^{\alpha}}\right]^{\frac{1}{1-\alpha}}
+ \frac{2(1 - \alpha)(c_0\mu)^{\alpha}}{\alpha} \right]
\right\} ^2\frac{1}{n}.
\end{align*}
The rest of the proof follows that of Theorem \ref{thm:flstd-rate}.
\end{proof}
\ \\[1ex]


\subsection{Proof of Lemma \ref{lemma:avg-exp-ls}}
\label{sec:appendix-avg-exp-ls}
\begin{proof}
Recall that $\gamma_n \triangleq c_0\left(\frac{c}{(c+n)}\right)^{\alpha}$. Recall that in Theorem \ref{cor:ls-iter-av} we have assumed that
\begin{align}
0 < c_0 \Phi_{\max}^2 < 1\label{eq:ap_ia_lem2_assump-ls}.
\end{align}
	Using \eqref{eq:ls-expectation-bound} we have
	\begin{align}
	&\E\left(\l \theta_n - \hat\theta_T\r\right)^2\nonumber\\
	&\le \left[\prod_{k = 1}^n
	\left(1 - \mu\gamma_k(2 - \gamma_k\Phi_{\max}^2\right)
	\l z_0\r\right]^2
	+  4\sum_{k=1}^{n}\gamma_k^2
	\left[\prod_{j = k}^{n-1}
	(1 - \mu\gamma_j(2  - \gamma_j\Phi_{\max}^2) \right]^2
	h(k)^2\nonumber\\
	&\le \left[\prod_{k = 1}^n \left(1 - \frac{\mu c_0 c^{\alpha}}{(c+k)^{\alpha}}\right)
	\l z_0\r\right]^2
	+  4\sum_{k=1}^{n}\frac{c_0^2 c^{2\alpha}}{(c+k)^{2\alpha}}
	\left[\prod_{j = k}^{n-1}
	\left(1 - \frac{\mu c_0 c^{\alpha}}{(c+j)^{\alpha}}\right) \right]^2
	h(k)^2\label{eq:ap_ia_lem2_3-ls}\\
	&\le \left[ \e\left(-\mu c_0 \sum_{k = 1}^n \frac{c^{\alpha}}{(c+k)^{\alpha}}\right)
	\l z_0\r\right]^2
	+  4h(n)^2\sum_{k=1}^{n}\frac{c_0^2 c^{2\alpha}}{(c+k)^{2\alpha}}
	\e\left(-2\mu c_0\sum_{j = k}^{n-1}
	\frac{ c^{\alpha}}{(c+j)^{\alpha}}\right)
	\nonumber.
	\end{align}
	To obtain \eqref{eq:ap_ia_lem2_3-ls}, we have applied \eqref{eq:ap_ia_lem2_assump-ls}. For the final inequality, we have exponentiated the logarithm of the products, and used the inequality $\ln(1+x) < x$ in several places.
	
	Continuing the derivation, we have
	\begin{align}
	\E&\l \theta_n - \hat\theta_T \r\\
	&\le \e\left(-c_0\mu c^\alpha(n+c)^{1-\alpha} - c_0\mu c^\alpha(1+c)^{1-\alpha}\right)\l \theta_0 - \hat\theta_T\r \nonumber\\
	&\quad+2h(n)\left(\sum_{k = 1}^{n}c_0^2\left(\frac{c}{k+c}\right)^{2\alpha}
	\e\left(-2c_0\mu c^\alpha((n+c)^{1-\alpha} - (k+c)^{1-\alpha}\right)
	\right)^{\frac{1}{2}}\label{eq:lem2-eq-0-ls}\\
	&=  \e\left(-c_0\mu c^\alpha(n+c)^{1-\alpha}\right)\nonumber\\
	& \qquad. \Bigg[\e\left(c_0\mu c^\alpha(1+c)^{1-\alpha}\right)\l \theta_0 - \hat\theta_T\r\nonumber\\
	& \qquad \qquad+ 2h(n) \left\{\sum_{k = 1}^{n}c_0^2\left(\frac{c}{k+c}\right)^{2\alpha}
	\e\left(2c_0\mu c^\alpha((k+c)^{1-\alpha}\right)
	\right\}^{\frac{1}{2}}
	\Bigg]\nonumber\\	
	\le&\e\left(-c_0\mu c^\alpha(n+c)^{1-\alpha}\right)\nonumber\\
	& \qquad. \Bigg[\e\left(c_0\mu c^\alpha(1+c)^{1-\alpha}\right)\l \theta_0 - \hat\theta_T\r\nonumber\\
	& \qquad \qquad+ 2 h(n) \left\{ c^{2\alpha} c_0^2\int_{1}^{n+c}x^{-2\alpha}\e \left(2c_0\mu c^\alpha x^{1-\alpha}\right)dx
	\right\}^{\frac{1}{2}}
	\Bigg]\label{eq:lem2-eq-2-ls}\\	
	\le&\e\left(-c_0\mu c^\alpha(n+c)^{1-\alpha}\right)\nonumber\\
	& \qquad. \Bigg[\e\left(c_0\mu c^\alpha(1+c)^{1-\alpha}\right)\l \theta_0 - \hat\theta_T\r\nonumber\\
	& \qquad \qquad+ 2 h(n) \left\{ c^{2\alpha} c_0^2\left(2c_0\mu c^\alpha \right)^{\frac{2\alpha}{1-\alpha}}
	.\int_{\left(2c_0\mu c^\alpha\right)^{1/(1-\alpha)}}^{(n+c)\left(2c_0\mu c^\alpha\right)^{1/(1-\alpha)}}
	y^{-2\alpha}\e (y^{1-\alpha})dy
	\right\}^{\frac{1}{2}}
	\Bigg]\label{eq:lem2-eq-3-ls}
	\end{align}
	As in the proof of Theorem \ref{thm:flstd-avg-rate}, for arriving at \eqref{eq:lem2-eq-0-ls}, we have used Jensen's Inequality, and that $\sum_{j=k}^{n-1}(c+j)^{-\alpha}\ge \int_{j=k}^n(c+j)^{1-\alpha}dj=(c+n)^{1-\alpha} - (c+k)^{1-\alpha}$. To obtain \eqref{eq:lem2-eq-2-ls}, we have upper bounded the sum with an integral, the validity of which follows from the observation that $x\mapsto x^{-2\alpha}e^{x^{1-\alpha}}$ is convex for $x\ge 1$. Finally, for \eqref{eq:lem2-eq-3-ls}, we have applied the change of variables $y = (2c_0\mu c^\alpha)^{1/(1-\alpha)}x$.
	
	Now, since $y^{-2\alpha} \le \frac{2}{1-\alpha} ((1-\alpha)y^{-2\alpha} - \alpha y^{-(1+\alpha)})$ when $y\ge\left( \frac{2\alpha}{1-\alpha}\right)^{\frac{1}{1-\alpha}}$, we have
	\begin{align*}
	&\int_{\left( \frac{2\alpha}{1-\alpha}\right)^{\frac{1}{1-\alpha}}}
	^{(n+c)\left(2c_0\mu c^\alpha\right)^{1/(1-\alpha)}}
	y^{-2\alpha}\e (y^{1-\alpha})dy\\
	&\quad
	\le \frac{2}{1-\alpha} \int_{\left( \frac{2\alpha}{1-\alpha}\right)^{\frac{1}{1-\alpha}}}
	^{(n+c)\left(2c_0\mu c^\alpha\right)^{1/(1-\alpha)}}
	((1-\alpha)y^{-2\alpha} - \alpha y^{-(1+\alpha)})
	\e (y^{1-\alpha})dy\\
	&\quad
	\le \frac{2}{1-\alpha} \e \left(2c_0\mu c^\alpha (n+c)^{1-\alpha}\right)
	(n+c)^{-\alpha}\left(2c_0\mu c^\alpha\right)^{-\alpha/(1-\alpha)}
	\end{align*}
	and furthermore, since $y\mapsto y^{-2\alpha}\e(y^{1-\alpha})$ is decreasing for $y\le\left( \frac{2\alpha}{1-\alpha}\right)^{\frac{1}{1-\alpha}}$, we have
	\begin{align*}
	\int_{1}^{\left( \frac{2\alpha}{1-\alpha}\right)^{\frac{1}{1-\alpha}}}
	y^{-2\alpha}\e (y^{1-\alpha})dy
	\le e \left( \frac{2\alpha}{1-\alpha}\right)^{\frac{1}{1-\alpha}}.
	\end{align*}
	Plugging these into \eqref{eq:lem2-eq-3-ls}, we obtain
	\begin{align*}
	&\E\l \theta_n - \hat\theta_T \r
	\le \e\left(-c_0\mu c^\alpha(n+c)^{1-\alpha}\right)\nonumber\\
	&.\left(\e\left(c_0\mu c^\alpha(1+c)^{1-\alpha}\right)\l \theta_0 - \theta_T\r
	+ 2h(n)c^{\alpha}c_0 \left(2 c_0 \mu c^{\alpha}\right)^{\frac{\alpha}{(1-\alpha)}} \sqrt{e}\left(\frac{2\alpha}{1-\alpha}\right)^{\frac{1}{2(1-\alpha)}}\right)\\
	&+ 2 h(n)c^\alpha c_0
	\left( 2c_0\mu c^\alpha\right)^{\frac{\alpha}{2(1-\alpha)}} 
	(n+c)^{-\frac{\alpha}{2}}.
	\end{align*}
	Hence, we obtain
	\begin{align*}
	&\E\l \bar\theta_n - \hat\theta_T\r
	\le \left(\sum_{n=1}^{\infty}
	\e\left(-c_0\mu c^\alpha(n+c)^{1-\alpha}\right) \right)\\
	&\quad
	.\left(\e\left(c_0\mu c^\alpha(1+c)^{1-\alpha}\right)\l \theta_0 - \theta_T\r
	+ 2h(n)c^{\alpha}c_0 \left(2 c_0 \mu c^{\alpha}\right)^{\frac{\alpha}{(1-\alpha)}} \sqrt{e}\left(\frac{2\alpha}{1-\alpha}\right)^{\frac{1}{2(1-\alpha)}}\right)\frac{1}{n}\\
	&+ 2 h(n) c^\alpha c_0
	\left( 2c_0\mu c^\alpha\right)^{\frac{\alpha}{2(1-\alpha)}}
	(n+c)^{-\frac{\alpha}{2}}.
	\end{align*}
\end{proof}

\bibliographystyle{amsplain}
\bibliography{references}


\end{document}